\documentclass[a4paper,10pt,onecolumn]{article}
\usepackage[dvips]{graphicx} 
\usepackage{epsfig}  
\usepackage{subfigure}
\usepackage{algorithmic}
\usepackage{algorithm}
\usepackage{textcomp}
\usepackage{multirow}
\usepackage{verbatim}
\usepackage{amsmath}
\usepackage{mathrsfs}
\usepackage{amsfonts}
\usepackage{amssymb}
\usepackage{amsthm}
\usepackage{epsfig}  
\usepackage{subfigure}
\usepackage{sectsty}
\sectionfont{\large}
\subsectionfont{\normalsize}
\usepackage[top=4cm, bottom=3cm, left=4cm, right=4cm]{geometry}

\newtheorem{theorem}{Theorem}

\title{\textbf{\Large Converged Algorithms for \\ Orthogonal Nonnegative Matrix Factorizations}}
\author{\large Andri Mirzal\\ \normalsize Graduate School of Information Science and Technology,\\ \normalsize Hokkaido University, Kita 14 Nishi 9, Kita-Ku,\\
\normalsize Sapporo 060-0814, Japan}
\date{}
\begin{document}

\setcounter{page}{1}
\maketitle

\noindent\textbf{Abstract}: This paper proposes uni-orthogonal and bi-orthogonal nonnegative matrix factorization algorithms with robust convergence proofs. We design the algorithms based on the work of Lee and Seung \cite{Lee2}, and derive the converged versions by utilizing ideas from the work of Lin \cite{CJLin2}. The experimental results confirm the theoretical guarantees of the convergences. 

\section{Introduction} \label{introduction}
The nonnegative matrix factorization (NMF) is a technique that decomposes a nonnegative data matrix into a pair of other nonnegative matrices of lower rank:
\begin{equation}
\mathbf{A} \approx \mathbf{B}\mathbf{C},
\label{eq1}
\end{equation}
where $\mathbf{A}\in\mathbb{R}_{+}^{M\times N}=\left[\mathbf{a}_1,\ldots,\mathbf{a}_N\right]$ denotes the data matrix, $\mathbf{B}\in\mathbb{R}_{+}^{M\times K}=\left[\mathbf{b}_1,\ldots,\mathbf{b}_K\right]$ denotes the basis matrix, $\mathbf{C}\in\mathbb{R}_{+}^{K\times N}=\left[\mathbf{c}_1,\ldots,\mathbf{c}_N\right]$ denotes the coef\mbox{}ficient matrix, and $K$ denotes the number of factors which usually is chosen so that $K\ll\min(M,N)$. To compute $\mathbf{B}$ and $\mathbf{C}$, usually eq.~\ref{eq1} is rewritten into a minimization problem in Frobenius norm criterion.
\begin{equation}
\min_{\mathbf{B},\mathbf{C}}J(\mathbf{B},\mathbf{C})=\frac{1}{2}\|\mathbf{A}-\mathbf{B}\mathbf{C}\|_{F}^{2}\;\,\mathrm{s.t.}\;\, \mathbf{B}\ge\mathbf{0},\mathbf{C}\ge\mathbf{0}.
\label{ch3:eq2}
\end{equation}

Orthogonal NMFs are introduced by Ding et al.~\cite{Ding1} to enforce orthogonality constraints on columns of $\mathbf{B}$ and/or rows of $\mathbf{C}$ in order to improve clustering capability of the standard NMF (we will refer NMF objective in eq.~\ref{ch3:eq2} as the standard NMF for the rest of this paper). Because clustering indicator matrices are orthogonal (hard clustering cases), imposing orthogonality on columns of $\mathbf{B}$ (rows of $\mathbf{C}$) will potentially produce a sharper row clustering indicator matrix (column clustering indicator matrix), and therefore it is expected that this mechanism will lead to better clustering methods. 

However, as the original orthogonal NMF algorithms \cite{Ding1} and the variants \cite{Yoo1,Yoo2, Choi} are all based on the multiplicative update (MU) rules, there is no convergence guarantee for these algorithms (in section \ref{MUR} we will explain why MU based algorithms do not have convergence guarantee). And because the orthogonality constraints cannot be recast into alternating nonnegativity least square (ANLS) framework (see \cite{HKim2, HKim} for discussion on ANLS), converged algorithms for the standard NMF, e.g., \cite{DKim, DKim2, CJLin2, CJLin, HKim, JKim2}, cannot be utilized for solving orthogonal NMF problems. Thus, there is still no converged algorithm for orthogonal NMFs.

The proposed algorithms are designed by generalizing the work of Lin \cite{CJLin2} in which he provides a converged algorithm for the standard NMF based on the additive update (AU) rules. The generalization presented in this chapter is not trivial since the proofs are developed in matrix form, thus providing a framework for developing converged algorithms for other NMF objectives that have matrix based auxiliary constraints with mutually dependency between columns and/or rows (Lin uses vector form for developing the proofs, so the interdependency between columns and/or rows cannot be captured). 

Also, in the process of developing the proofs, the objectives need to be decomposed into the Taylor series. When the objectives have only up to second order derivatives, then the nonincreasing properties can be proven by showing the positive-definiteness of the Hessians of the objectives \cite{Lee2, CJLin2}. But in general cases, the objectives can have more than second order derivatives. And in particular, the orthogonality constraints make the objectives have more than second order derivatives. Thus, the same strategy cannot be used for the general cases. Accordingly, we introduce a strategy to deal with this kind of objectives. Note that the proofs presented here are sufficiently general to be a framework for developing converged algorithms for other NMF objectives with well-defined partial derivatives up to second order.

\section{Multiplicative update algorithm} \label{MUR}

In \cite{Lee2}, Lee and Seung introduce two MU rules based algorithms for the standard NMF using the Frobenius norm and the Kullback-Leibler divergence respectively as the distance measure. In addition, they also show how to modify the Frobenius norm based MU algorithm into AU version. However, due to numerical difficulties of the Kullback-Leibler divergence, and computational requirements of the AU algorithm, only the Frobenius norm based MU algorithm is being extensively studied. In this section, we will review the Frobenius norm based MU algorithm and discuss the reason why this algorithm do not have convergence guarantee. Note that only the Frobenius norm will be considered for the rest of this chapter.

First let us rewrite the standard NMF objective in eq.~\ref{ch3:eq2}.
\begin{equation}
\min_{\mathbf{B},\mathbf{C}}J(\mathbf{B},\mathbf{C})=\frac{1}{2}\|\mathbf{A}-\mathbf{BC}\|_{F}^{2}\;\,\mathrm{s.t.}\;\,\mathbf{B}\ge\mathbf{0},\mathbf{C}\ge\mathbf{0}. \label{eq4}
\end{equation} 
The KKT function of the objective is:
\begin{equation*}
L(\mathbf{B},\mathbf{C})=\;J(\mathbf{B},\mathbf{C})-\mathrm{tr}\;(\mathbf{\Gamma}_{\mathbf{B}}\mathbf{B}^T)-\mathrm{tr}\;(\mathbf{\Gamma}_{\mathbf{C}}\mathbf{C}), 
\end{equation*}
where $\mathbf{\Gamma}_{\mathbf{B}}\in\mathbb{R}_{+}^{M\times R}$ and $\mathbf{\Gamma}_{\mathbf{C}}\in\mathbb{R}_{+}^{N\times R}$ are the KKT multipliers. Partial derivatives of $L$ with respect to $\mathbf{B}$ and $\mathbf{C}$ can be written as:
\begin{align*}
\nabla_{\mathbf{B}}L(\mathbf{B})=\;&\nabla_{\mathbf{B}}J(\mathbf{B})-\mathbf{\Gamma}_{\mathbf{B}},\;\;\;\mathrm{and} 
\\
\nabla_{\mathbf{C}}L(\mathbf{C})=\;&\nabla_{\mathbf{C}}J(\mathbf{C})-\mathbf{\Gamma}_{\mathbf{C}}^T, 
\end{align*}
with
\begin{align*}
\nabla_{\mathbf{B}}J(\mathbf{B})=\;&\mathbf{BCC}^T-\mathbf{AC}^T,\;\;\;\mathrm{and} 
\\
\nabla_{\mathbf{C}}J(\mathbf{C})=\;&\mathbf{B}^T\mathbf{BC}-\mathbf{B}^T\mathbf{A}. 
\end{align*}
By results from optimization studies, ($\mathbf{B}^*,\mathbf{C}^*$) is a stationary point of eq.~\ref{eq4} if it satisfies the KKT optimality conditions \cite{Bertsekas}, i.e.,
\begin{align}
\mathbf{B}^*&\ge\mathbf{0},&\mathbf{C}^*\ge\mathbf{0},\nonumber \\ 
\nabla_{\mathbf{B}}J(\mathbf{B}^*)=\mathbf{\Gamma}_{\mathbf{B}}&\ge\mathbf{0},&\nabla_{\mathbf{C}}J(\mathbf{C}^*)=\mathbf{\Gamma}_{\mathbf{C}}^T\ge\mathbf{0},\nonumber \\
\nabla_{\mathbf{B}}J(\mathbf{B}^*)\odot\mathbf{B}^*&=\mathbf{0},&\nabla_{\mathbf{C}}J(\mathbf{C}^*)\odot\mathbf{C}^*=\mathbf{0}, \label{eq12}
\end{align}
where $\odot$ denotes component-wise multiplications, and eq.~\ref{eq12} is known as the complementary slackness.

The MU algorithm is derived by utilizing the complementary slackness:
\begin{align*}
\big(\mathbf{BCC}^T-\mathbf{AC}^T\big)\odot\mathbf{B}&=\mathbf{0},\\
\big(\mathbf{B}^T\mathbf{BC}-\mathbf{B}^T\mathbf{A}\big)\odot\mathbf{C}&=\mathbf{0}.
\end{align*}
These equations lead to the following update rules \cite{Lee2}:
\begin{align}
b_{mr}^{k+1} &\longleftarrow b_{mr}^{k}\frac{\big(\mathbf{AC}^T\big)_{mr}}{\big(\mathbf{BCC}^T\big)_{mr}}\;\;\forall m,r, \label{eq15}\\
c_{rn}^{k+1} &\longleftarrow c_{rn}^{k}\frac{\big(\mathbf{B}^T\mathbf{A}\big)_{rn}}{\big(\mathbf{B}^T\mathbf{BC}\big)_{rn}}\;\;\forall r,n, \label{eq16}
\end{align}
where $k=0,\ldots,K$ denotes the iteration, $K$ denotes the maximum iteration, $b_{mr}^k$ and $c_{rn}^k$ denote ($m,r$) entry of $\mathbf{B}$ and ($r,n$) entry of $\mathbf{C}$ at $k$-th iteration respectively. These equations are the MU algorithm for the standard NMF problem in eq.~\ref{eq4}. 

\begin{theorem}[Lee and Seung \cite{Lee2}]\label{theorem1}
Objective in eq.~\ref{eq4} is nonincreasing under the update rules eq.~\ref{eq15} and \ref{eq16}, i.e., $J\big(\mathbf{B}^{k+1},\mathbf{C}^{k+1}\big)\le J\big(\mathbf{B}^{k+1},\mathbf{C}^{k}\big)\le J\big(\mathbf{B}^{k},\mathbf{C}^{k}\big)\;\;\forall k\ge 0$.
\end{theorem}
\begin{theorem}[Lin \cite{CJLin}] \label{theorem2}
If $\mathbf{A}$ has neither zero column nor row, and $\mathbf{B}^0>\mathbf{0}$ and $\mathbf{C}^0>\mathbf{0}$, then $\mathbf{B}^k>\mathbf{0}$ and $\mathbf{C}^k>\mathbf{0}\;\;\forall k\ge 0$ under the update rules eq.~\ref{eq15} and \ref{eq16}.
\end{theorem}
\begin{theorem}\label{theorem3}
Given $\mathbf{A}$, $\mathbf{B}^0$, and $\mathbf{C}^0$ satisfy the conditions in theorem \ref{theorem2}, if ($\mathbf{B}^*,\mathbf{C}^*$) is a stationary point on the feasible region, then the update rules eq.~\ref{eq15} and \ref{eq16} will stop updating $\mathbf{B}^*$ and $\mathbf{C}^*$.
\end{theorem}
\begin{proof}
Because any stationary point satisfies the KKT conditions and $\mathbf{B}^k>\mathbf{0}$ and $\mathbf{C}^k>\mathbf{0}\;\;\forall k\ge 0$, then by using the complementary slackness it can be shown that $\nabla_{\mathbf{B}}J(\mathbf{B}^*)=\mathbf{0}$ and $\nabla_{\mathbf{C}}J(\mathbf{C}^*)=\mathbf{0}$. Accordingly, $\mathbf{AC}^{*T}=\mathbf{B}^*\mathbf{C}^*\mathbf{C}^{*T}$ and $\mathbf{B}^{*T}\mathbf{A}=\mathbf{B}^{*T}\mathbf{B}^*\mathbf{C}^*$, therefore $\mathbf{B}^k=\mathbf{B}^*$ and $\mathbf{C}^k=\mathbf{C}^*\;\;\forall k>*$.
\end{proof}
\begin{theorem}\label{theorem4}
If there exists ($m,r$) or ($r,n$) so that $b_{mr}^l=0$ or $c_{rn}^l=0$ for some $l\ge 0$, then when the eq.~\ref{eq15} and \ref{eq16} stop updating, there is no guarantee that this point is a stationary point.
\end{theorem}
\begin{proof}
If $b_{mr}^l=0$ ($c_{rn}^l=0$), then $b_{mr}^k=0$ ($c_{rn}^k=0$) $\forall k\ge l$. Consequently, we must make sure that $\nabla_{\mathbf{B}}J(\mathbf{B})_{mr}^k\ge 0$ ($\nabla_{\mathbf{C}}J(\mathbf{C})_{rn}^k\ge 0$) $\forall k\ge l$ for this point to satisfy the KKT conditions. When there exists $k$ such that this requirement is not satisfied, then there is no stationarity guarantee.
\end{proof}

So, while theorem \ref{theorem3} states that the MU algorithm can reach stationary points, theorem \ref{theorem4} gives the reason why the MU algorithm cannot guarantee to converge to the stationary points.

To avoid division by zero, the MU algorithm usually is modified into:
\begin{align*}
b_{mr}^{(k+1)} &\longleftarrow b_{mr}^{k}\frac{\big(\mathbf{AC}^T\big)_{mr}}{\big(\mathbf{BCC}^T\big)_{mr}+\delta}\;\;\forall m,r, \\
c_{rn}^{(k+1)} &\longleftarrow c_{rn}^{k}\frac{\big(\mathbf{B}^T\mathbf{A}\big)_{rn}}{\big(\mathbf{B}^T\mathbf{BC}\big)_{rn}+\delta}\;\;\forall r,n,
\end{align*}
where $\delta$ is a small positive number. The complete MU algorithm for the standard NMF is given in algorithm \ref{algorithm1}.

\begin{algorithm}
\caption{The MU algorithm for the standard NMF (Lee \& Seung algorithm \cite{Lee2}).}
\label{algorithm1}
\begin{algorithmic}
\STATE Initialization, $\mathbf{B}^0>\mathbf{0}$ and $\mathbf{C}^0>\mathbf{0}$.
\FOR {$k=0,\ldots,K$}
\STATE \begin{align*} b_{mr}^{(k+1)} &\longleftarrow b_{mr}^{k}\frac{\big(\mathbf{AC}^{kT}\big)_{mr}}{\big(\mathbf{B}^{k}\mathbf{C}^{k}\mathbf{C}^{kT}\big)_{mr}+\delta}\;\;\forall m,r \\
c_{rn}^{(k+1)} &\longleftarrow c_{rn}^{k}\frac{\big(\mathbf{B}^{(k+1)T}\mathbf{A}\big)_{rn}}{\big(\mathbf{B}^{(k+1)T}\mathbf{B}^{(k+1)}\mathbf{C}^{k}\big)_{rn}+\delta}\;\;\forall r,n \end{align*}
\ENDFOR
\end{algorithmic}
\end{algorithm}

As stated in theorem \ref{theorem4}, the initial values of $\mathbf{B}$ and $\mathbf{C}$ in algorithm \ref{algorithm1} have to be all positive to avoid zero locking from the start (see, e.g., \cite{CJLin2, CJLin} for detailed discussion on zero locking phenomenon). But, as shown in theorem \ref{theorem2}, assigning positive initialization will lead to solutions that lie on positive orthant of the feasible region, i.e., $\mathbf{B}^k>0$ and $\mathbf{C}^k>0\;\;\forall k\ge 0$ (at least theoretically). And consequently, the algorithm cannot find stationary points that lie on the boundary of the feasible region.

Note that some literatures, e.g.~\cite{CJLin2,Xu} recommend to normalize $\mathbf{B}$ for each iteration so that the Euclidian length of each its columns is one to guarantee the uniqueness of the solution (and consequently, each row of $\mathbf{C}$ has to be adjusted accordingly to preserve the objective value).

\begin{figure}
 \begin{center}
  \includegraphics[width=0.6\textwidth]{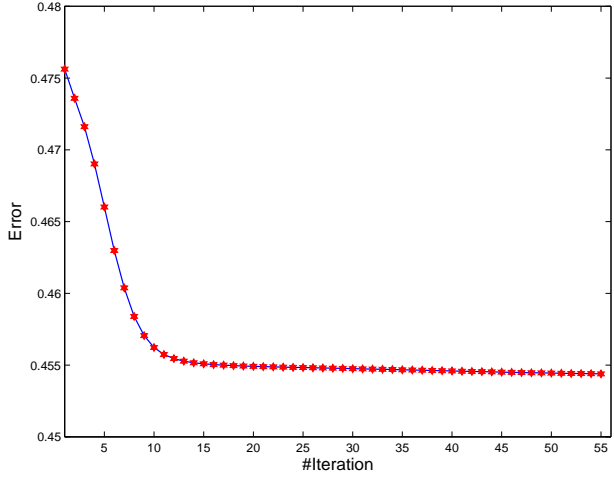}
  \caption{Error per iteration (Reuters4 dataset) of algorithm \ref{algorithm1}.}
  \label{fig0}
 \end{center}
\end{figure}

Fig.~\ref{fig0} shows the nonincreasing property of the algorithm \ref{algorithm1} which is guaranteed by theorem \ref{theorem1} for Reuters4 dataset (see section \ref{datasets} for discussion on the datasets). As the error, objective of the algorithm \ref{algorithm1} (eq.~\ref{eq4}) is used.

\section{Original Orthogonal NMF algorithms} \label{DingOrtho}

In \cite{Ding1}, Ding et al.~propose two MU rules based orthogonal NMF algorithms: uni-orthogonal NMF and bi-orthogonal NMF.

\subsection{Uni-orthogonal NMF}\label{uniortho}

Uni-orthogonal NMF (UNMF) imposes orthogonality constraint on either columns of $\mathbf{B}$ or rows of $\mathbf{C}$. We will discuss the orthogonality constraint on rows of $\mathbf{C}$ here. Similar result for $\mathbf{B}$ can be derived equivalently.

Objective for UNMF with orthogonality constraint on rows of $\mathbf{C}$ can be written as:
\begin{align}
&\min_{\mathbf{B},\mathbf{C}}J(\mathbf{B},\mathbf{C})=\frac{1}{2}\|\mathbf{A}-\mathbf{BC}\|_{F}^{2} \label{eq21}\\
&\mathrm{s.t.}\;\,\mathbf{B}\ge\mathbf{0},\;\,\mathbf{C}\ge\mathbf{0},\;\,\frac{1}{2}\big(\mathbf{CC}^T-\mathbf{I}\big)=\mathbf{0}. \nonumber
\end{align}
The KKT function of this objective is:
\begin{align}
L(\mathbf{B},\mathbf{C})=\;&J(\mathbf{B},\mathbf{C})-\mathrm{tr}\;\left(\mathbf{\Gamma}_{\mathbf{B}}\mathbf{B}^T\right)-\mathrm{tr}\;\left(\mathbf{\Gamma}_{\mathbf{C}}\mathbf{C}\right) + \frac{1}{2}\mathrm{tr}\;\left(\mathbf{\Lambda}_{\mathbf{C}}\left(\mathbf{C}\mathbf{C}^T-\mathbf{I}\right)\right), \label{eq22}
\end{align}
where $\mathbf{\Gamma}_{\mathbf{B}}\in\mathbb{R}_{+}^{M\times R}$, $\mathbf{\Gamma}_{\mathbf{C}}\in\mathbb{R}_{+}^{N\times R}$, and $\mathbf{\Lambda}_{\mathbf{C}}\in\mathbb{R}_{+}^{R\times R}$ are the KKT multipliers. Instead of solving the three-constraint objective in eq.~\ref{eq21}, Ding et al.~\cite{Ding1} propose the following objective:
\begin{align}
&\min_{\mathbf{B},\mathbf{C}}J(\mathbf{B},\mathbf{C})=\frac{1}{2}\|\mathbf{A}-\mathbf{BC}\|_{F}^{2} + \frac{1}{2}\mathrm{tr}\;\left(\mathbf{\Lambda}_{\mathbf{C}}\left(\mathbf{C}\mathbf{C}^T-\mathbf{I}\right)\right)\label{eq23}\\
&\mathrm{s.t.}\;\,\mathbf{B}\ge\mathbf{0},\;\,\mathbf{C}\ge\mathbf{0}.\nonumber
\end{align}
Note that, even though both objectives (eq.~\ref{eq21} and \ref{eq23}) have the same KKT function, i.e.,~eq.~\ref{eq22}, they are not exactly the same, as the orthogonality constraint is absorbed into the minimization problem. 

The KKT conditions for objective in eq.~\ref{eq23} are:
\begin{equation}
\begin{array}{rr}
\mathbf{B}^*\ge\mathbf{0}, &\mathbf{C}^*\ge\mathbf{0},\\
\nabla_{\mathbf{B}}J(\mathbf{B}^*)=\mathbf{\Gamma}_{\mathbf{B}}\ge\mathbf{0}, &\nabla_{\mathbf{C}}J(\mathbf{C}^*)=\mathbf{\Gamma}_{\mathbf{C}}^T\ge\mathbf{0},\\
\nabla_{\mathbf{B}}J(\mathbf{B}^*)\odot\mathbf{B}^*=\mathbf{0}, &\nabla_{\mathbf{C}}J(\mathbf{C}^*)\odot\mathbf{C}^*=\mathbf{0}, \nonumber
\end{array}
\end{equation}
with
\begin{align*}
\nabla_{\mathbf{B}}J(\mathbf{B})&=\mathbf{BCC}^T-\mathbf{AC}^T \\
\nabla_{\mathbf{C}}J(\mathbf{C})&=\mathbf{B}^T\mathbf{BC}-\mathbf{B}^T\mathbf{A}+\mathbf{\Lambda}_{\mathbf{C}}\mathbf{C}
\end{align*}

By using the same strategy as in section \ref{MUR}, MU rules based UNMF algorithm can be written as:
\begin{align}
b_{mr} &\longleftarrow b_{mr}\frac{(\mathbf{A}\mathbf{C}^T)_{mr}}{(\mathbf{BCC}^T)_{mr}} \label{eq27}\\
c_{rn} &\longleftarrow c_{rn}\frac{(\mathbf{B}^T\mathbf{A})_{rn}}{\left[(\mathbf{B}^T\mathbf{B} + \mathbf{\Lambda}_{\mathbf{C}})\mathbf{C}\right]_{rn}}. \label{eq28}
\end{align}

The problem with this algorithm is how to determine $\mathbf{\Lambda}_{\mathbf{C}}$. By summing over index $r$, Ding et al.~find an exact formulation for the diagonal entries:
\begin{equation}
\big(\mathbf{\Lambda}_{\mathbf{C}}\big)_{rr} = \big(\mathbf{B}^T\mathbf{AC}^T - \mathbf{B}^T\mathbf{B}\big)_{rr}. \label{eq29}
\end{equation}
The off-diagonal entries are obtained by ignoring the nonnegativity constraint on $\mathbf{C}$ and by setting $\nabla_{\mathbf{C}}J(\mathbf{C})$ ($J$ in eq.~\ref{eq23}) to zero matrix:
\begin{align}
\nabla_{\mathbf{C}}J(\mathbf{C}) &= -\mathbf{B}^T\mathbf{A} + \mathbf{B}^T\mathbf{BC} + \mathbf{\Lambda}_{\mathbf{C}}\mathbf{C} = \mathbf{0}, \label{eq30}\\
\big(\mathbf{\Lambda}_{\mathbf{C}}\big)_{rs} &= \big(\mathbf{B}^T\mathbf{AC}^T - \mathbf{B}^T\mathbf{B}\big)_{rs}.\;\;\forall r\ne s.\label{eq31} 
\end{align}
Eq.~\ref{eq30} is derived from eq.~\ref{eq23} by using the fact $\|\mathbf{X}\|_F^2 = \mathrm{tr}\big(\mathbf{A}^T\mathbf{A}\big)$, and eq.~\ref{eq31} is derived from eq.~\ref{eq30} by using the orthogonality constraint $\mathbf{CC}^T = \mathbf{I}$. By combining eq.~\ref{eq29} and eq.~\ref{eq31}, $\mathbf{\Lambda}_{\mathbf{C}}$ can be defined as:
\begin{equation}
\mathbf{\Lambda}_{\mathbf{C}} = \mathbf{B}^T\mathbf{AC}^T - \mathbf{B}^T\mathbf{B}.
\label{eq32} 
\end{equation}
Accordingly, the UNMF algorithm can be rewritten as:
\begin{align}
b_{mr} &\longleftarrow b_{mr}\frac{(\mathbf{A}\mathbf{C}^T)_{mr}}{(\mathbf{BCC}^T)_{mr}} \label{eq33}\\
c_{rn} &\longleftarrow c_{rn}\frac{(\mathbf{B}^T\mathbf{A})_{rn}}{\big(\mathbf{B}^T\mathbf{AC}^T\mathbf{C}\big)_{rn}}. \label{eq34}
\end{align}
The complete UNMF algorithm for eq.~\ref{eq33} and \ref{eq34} is given in algorithm \ref{algorithm2}. Unlike in algorithm \ref{algorithm1}, normalization will change the objective value in eq.~\ref{eq23} as there is $\mathrm{tr}\;\big(\mathbf{\Lambda}_{\mathbf{C}}\big(\mathbf{C}\mathbf{C}^T-\mathbf{I}\big)\big)$ component, thus it is not recommended.
\begin{algorithm}
\caption{UNMF algorithm due to the work of Ding et al.~\cite{Ding1}.}
\label{algorithm2}
\begin{algorithmic}
\STATE Initialization, $\mathbf{B}^0>\mathbf{0}$ and $\mathbf{C}^0>\mathbf{0}$.
\FOR {$k=0,\ldots,K$}
\STATE \begin{align*} b_{mr}^{(k+1)} &\longleftarrow b_{mr}^{k}\frac{\big(\mathbf{AC}^{kT}\big)_{mr}}{\big(\mathbf{B}^{k}\mathbf{C}^{k}\mathbf{C}^{kT}\big)_{mr}+\delta}\;\;\forall m,r \\
c_{rn}^{(k+1)} &\longleftarrow c_{rn}^{k}\frac{\big(\mathbf{B}^{(k+1)T}\mathbf{A}\big)_{rn}}{\big(\mathbf{B}^{(k+1)T}\mathbf{A}\mathbf{C}^{kT}\mathbf{C}^{k}\big)_{rn}+\delta}\;\;\forall r,n \end{align*}
\ENDFOR
\end{algorithmic}
\end{algorithm}

\begin{figure}
 \begin{center}
  \includegraphics[width=0.6\textwidth]{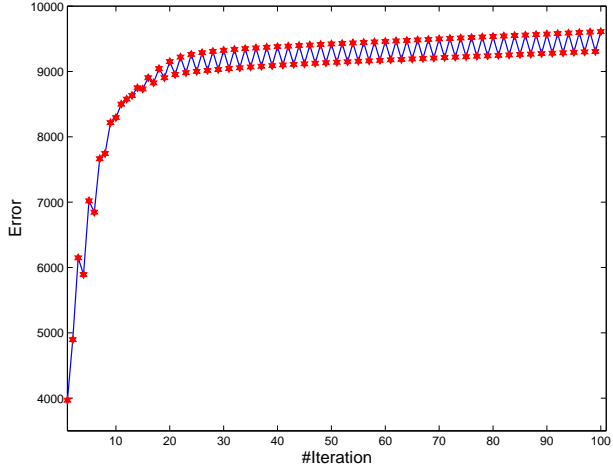}
  \caption{Error per iteration (Reuters4 dataset) of algorithm \ref{algorithm2}.}
  \label{fig1}
 \end{center}
\end{figure}

Note that as there is an assumption in deriving $\mathbf{\Lambda}_{\mathbf{C}}$, algorithm \ref{algorithm2} may or may not be minimizing the objective eq.~\ref{eq23}. Further, the auxiliary function used by the authors to prove the nonincreasing property is for UNMF algorithm in eq.~\ref{eq27} and \ref{eq28}, not for algorithm \ref{algorithm2}. So there is no guarantee that algorithm \ref{algorithm2} has the nonincreasing property. Figure \ref{fig1} gives a numerical example on how algorithm \ref{algorithm2} not only does not have the nonincreasing property but also fails to minimize the objective. As the error, the objective of UNMF (eq.~\ref{eq23}) is used with $\mathbf{\Lambda}_{\mathbf{C}}$ defined in eq.~\ref{eq32}.

\subsection{Bi-orthogonal NMF}\label{biortho}

Bi-orthogonal NMF (BNMF) puts orthogonality constraints on both columns of $\mathbf{B}$ and rows of $\mathbf{C}$. Therefore it is expected that this technique can be used to simultaneously cluster columns and rows of $\mathbf{A}$. The following objective is the BNMF objective proposed by Ding et al.~\cite{Ding1}.
\begin{align}
&\min_{\mathbf{B},\mathbf{C},\mathbf{S}}J(\mathbf{B},\mathbf{C},\mathbf{S})=\frac{1}{2}\|\mathbf{A}-\mathbf{BSC}\|_{F}^{2} \label{eq37}\\
&\mathrm{s.t.}\;\,\mathbf{B}\ge\mathbf{0},\;\,\mathbf{S}\ge\mathbf{0},\;\,\mathbf{C}\ge\mathbf{0},\;\,\frac{1}{2}\big(\mathbf{CC}^T-\mathbf{I}\big)=\mathbf{0},\;\,\frac{1}{2}\big(\mathbf{B}^T\mathbf{B}-\mathbf{I}\big)=\mathbf{0}, \nonumber
\end{align}
where $\mathbf{B}\in\mathbb{R}_{+}^{M\times P}$ and $\mathbf{C}\in\mathbb{R}_{+}^{Q\times N}$ are defined similarly as before, and $\mathbf{S}\in\mathbb{R}_{+}^{P\times Q}$ is a matrix that introduced to absorb the different scales of $\mathbf{A}$, $\mathbf{B}$, and $\mathbf{C}$ due to the strict orthogonality constraints on $\mathbf{B}$ and $\mathbf{C}$. We will set $P=Q$ for the rest of this chapter.

The KKT function can be defined as:
\begin{align*}
L(\mathbf{B},\mathbf{C},\mathbf{S})=\;&J(\mathbf{B},\mathbf{C},\mathbf{S})-\mathrm{tr}\;\big(\mathbf{\Gamma}_{\mathbf{B}}\mathbf{B}^T\big)-\mathrm{tr}\;\big(\mathbf{\Gamma}_{\mathbf{S}}\mathbf{S}^T\big)-\mathrm{tr}\;\big(\mathbf{\Gamma}_{\mathbf{C}}\mathbf{C}\big) \\
&+ \frac{1}{2}\mathrm{tr}\;\big(\mathbf{\Lambda}_{\mathbf{C}}\big(\mathbf{CC}^T-\mathbf{I}\big)\big) + \frac{1}{2}\mathrm{tr}\;\big(\mathbf{\Lambda}_{\mathbf{B}}\big(\mathbf{B}^T\mathbf{B}-\mathbf{I}\big)\big),
\end{align*}
where $\mathbf{\Gamma}_{\mathbf{B}}$, $\mathbf{\Gamma}_{\mathbf{C}}$, $\mathbf{\Lambda}_{\mathbf{C}}$, $\mathbf{\Gamma}_{\mathbf{S}}\in\mathbb{R}_{+}^{P\times Q}$, and $\mathbf{\Lambda}_{\mathbf{B}}\in\mathbb{R}_{+}^{P\times P}$ are the KKT multipliers. 

An equivalent objective to eq.~\ref{eq37} is proposed by Ding et al.~\cite{Ding1} to absorb the orthogonality constraints into the objective:
\begin{align}
\min_{\mathbf{B},\mathbf{C},\mathbf{S}}J(\mathbf{B},\mathbf{C},\mathbf{S})=\;&\frac{1}{2}\|\mathbf{A}-\mathbf{BSC}\|_{F}^{2} + \frac{1}{2}\mathrm{tr}\;\big(\mathbf{\Lambda}_{\mathbf{C}}\big(\mathbf{C}\mathbf{C}^T-\mathbf{I}\big)\big) \nonumber \\
\;&+ \frac{1}{2}\mathrm{tr}\;\big(\mathbf{\Lambda}_{\mathbf{B}}\big(\mathbf{B}^T\mathbf{B}-\mathbf{I}\big)\big)\label{eq39}\\
\mathrm{s.t.}\;\,\mathbf{B}\ge\mathbf{0},\;\,\mathbf{C}\ge\mathbf{0}.\nonumber
\end{align}
The KKT conditions for objective in eq.~\ref{eq39} are:
\begin{equation}
\begin{array}{rrr}
\mathbf{B}^*\ge\mathbf{0}, & \mathbf{S}^*\ge\mathbf{0}, & \mathbf{C}^*\ge\mathbf{0}, \\
\nabla_{\mathbf{B}}J(\mathbf{B}^*)=\mathbf{\Gamma}_{\mathbf{B}}\ge\mathbf{0}, & \nabla_{\mathbf{S}}J(\mathbf{S}^*)=\mathbf{\Gamma}_{\mathbf{S}}\ge\mathbf{0}, & \nabla_{\mathbf{C}}J(\mathbf{C}^*)=\mathbf{\Gamma}_{\mathbf{C}}^T\ge\mathbf{0},\\
\nabla_{\mathbf{B}}J(\mathbf{B}^*)\odot\mathbf{B}^*=\mathbf{0}, & \nabla_{\mathbf{S}}J(\mathbf{S}^*)\odot\mathbf{S}^*=\mathbf{0}, & \nabla_{\mathbf{C}}J(\mathbf{C}^*)\odot\mathbf{C}^*=\mathbf{0}, \nonumber
\end{array}
\end{equation}
with
\begin{align*}
\nabla_{\mathbf{B}}J(\mathbf{B})&=\mathbf{BSCC}^T\mathbf{S}^T-\mathbf{AC}^T\mathbf{S}^T+\mathbf{B\Lambda}_{\mathbf{B}}, \\
\nabla_{\mathbf{C}}J(\mathbf{C})&=\mathbf{S}^T\mathbf{B}^T\mathbf{BSC}-\mathbf{S}^T\mathbf{B}^T\mathbf{A}+\mathbf{\Lambda}_{\mathbf{C}}\mathbf{C}, \\
\nabla_{\mathbf{S}}J(\mathbf{S})&=\mathbf{B}^T\mathbf{BSCC}^T-\mathbf{B}^T\mathbf{AC}^T.
\end{align*}

Then, by using the same strategy as in section \ref{MUR}, BNMF algorithm can be written as:
\begin{align*}
b_{mp} &\longleftarrow b_{mp}\frac{\big(\mathbf{A}\mathbf{C}^T\mathbf{S}^T\big)_{mp}}{\left[\mathbf{B}\big(\mathbf{SCC}^T\mathbf{S}^T + \mathbf{\Lambda}_{\mathbf{B}}\big)\right]_{mp}}, 
\\
c_{qn} &\longleftarrow c_{qn}\frac{\big(\mathbf{S}^T\mathbf{B}^T\mathbf{A}\big)_{qn}}{\left[\big(\mathbf{S}^T\mathbf{B}^T\mathbf{B}\mathbf{S} + \mathbf{\Lambda}_{\mathbf{C}}\big)\mathbf{C}\right]_{qn}}, 
\\
s_{pq} &\longleftarrow s_{pq}\frac{(\mathbf{B}^T\mathbf{A}\mathbf{C}^T)_{pq}}{(\mathbf{B}^T\mathbf{BSCC}^T)_{pq}}, 
\end{align*}
with
\begin{align*}
\mathbf{\Lambda}_{\mathbf{B}} &= \mathbf{B}^T\mathbf{AC}^T\mathbf{S}^T - \mathbf{SCC}^T\mathbf{S}^T\;\;\;\; \mathrm{and}\\
\mathbf{\Lambda}_{\mathbf{C}} &= \mathbf{S}^T\mathbf{B}^T\mathbf{AC}^T - \mathbf{S}^T\mathbf{B}^T\mathbf{B}\mathbf{S}
\end{align*}
are derived exactly for the diagonal entries, and approximately for off-diagonal entries by relaxing the nonnegativity constraints as in section \ref{uniortho}. 

The complete BNMF algorithm is shown in algorithm \ref{algorithm3}. And as in algorithm \ref{algorithm2}, the normalization step is not recommended as it will change the objective value.
\begin{algorithm}
\caption{BNMF algorithm due to the work of Ding et al.~\cite{Ding1}.}
\label{algorithm3}
\begin{algorithmic}
\STATE Initialization, $\mathbf{B}^0>\mathbf{0}$, $\mathbf{C}^0>\mathbf{0}$, and $\mathbf{S}^0>\mathbf{0}$.
\FOR {$k=0,\ldots,K$}
\STATE \begin{align*} b_{mp}^{(k+1)} &\longleftarrow b_{mp}^{k}\frac{\big(\mathbf{AC}^{kT}\mathbf{S}^{kT}\big)_{mp}}{\big(\mathbf{B}^{k}\mathbf{B}^{kT}\mathbf{A}\mathbf{C}^{kT}\mathbf{S}^{kT}\big)_{mp}+\delta}\;\;\forall m,p \\
c_{qn}^{(k+1)} &\longleftarrow c_{qn}^{k}\frac{\big(\mathbf{S}^{kT}\mathbf{B}^{(k+1)T}\mathbf{A}\big)_{qn}}{\big(\mathbf{S}^{kT}\mathbf{B}^{(k+1)T}\mathbf{A}\mathbf{C}^{kT}\mathbf{C}^{k}\big)_{qn}+\delta}\;\;\forall q,n \\
s_{pq}^{(k+1)} &\longleftarrow s_{pq}^{k}\frac{\big(\mathbf{B}^{(k+1)T}\mathbf{A}\mathbf{C}^{(k+1)T}\big)_{pq}}{\big(\mathbf{B}^{(k+1)T}\mathbf{B}^{(k+1)}\mathbf{S}^{k}\mathbf{C}^{(k+1)}\mathbf{C}^{(k+1)T}\big)_{pq}+\delta}\;\;\forall p,q \end{align*}
\ENDFOR
\end{algorithmic}
\end{algorithm}

\begin{figure}
\begin{center}
\includegraphics[width=0.6\textwidth]{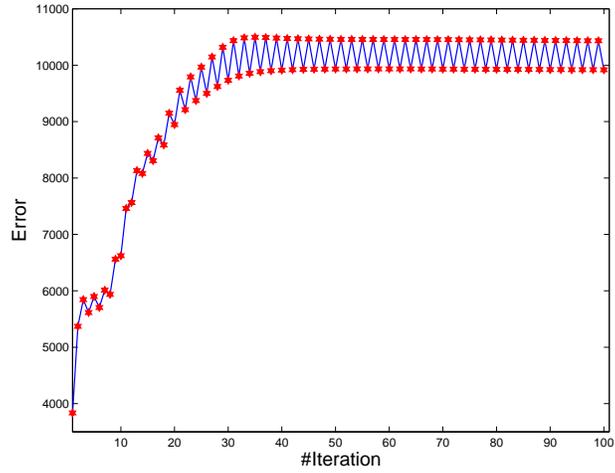}
\caption{Error per iteration of algorithm \ref{algorithm3} for Reuters4 dataset}
\label{fig2}
\end{center}
\end{figure}

Figure \ref{fig2} shows error per iteration of algorithm \ref{algorithm3}, with error is the objective value in eq.~\ref{eq39}. As in the UNMF case, the assumptions taken for obtaining $\mathbf{\Lambda}_{\mathbf{B}}$ and $\mathbf{\Lambda}_{\mathbf{C}}$ seem to be unreasonable since algorithm \ref{algorithm3} not only does not have the nonincreasing property but also fails to minimize the objective value.

\section{Converged orthogonal NMF algorithms}\label{converged}

In this section, we will present converged algorithms for UNMF and BNMF based on the AU rules which have been previously shown by Lin \cite{CJLin2} to have convergence guarantee. We will recast the orthogonality constraints directly into the objectives, and thus avoiding the necessity of absorbing them. We will show that this strategy allows us to design converged algorithms for UNMF and BNMF as easy as in the standard NMF case.

\subsection{Converged uni-orthogonal NMF} \label{myuniortho}

We define UNMF objective in following formulation:
\begin{align}
&\min_{\mathbf{B},\mathbf{C}}J(\mathbf{B},\mathbf{C})=\frac{1}{2}\|\mathbf{A}-\mathbf{B}\mathbf{C}\|_{F}^{2} + \frac{\alpha}{2}\|\mathbf{CC}^T-\mathbf{I}\|_{F}^{2} \label{eq52}\\ 
&\mathrm{s.t.}\;\, \mathbf{B}\ge\mathbf{0},\mathbf{C}\ge\mathbf{0}, \nonumber
\end{align}
with $\alpha$ is a constant to adjust the degree of orthogonality of $\mathbf{C}$. As shown, the orthogonality constraint is recast directly into the objective, and the constraints are now similar to the standard NMF.

The KKT function can be defined as:
\begin{align*}
L(\mathbf{B},\mathbf{C})=\;&J(\mathbf{B},\mathbf{C})-\mathrm{tr}\;\big(\mathbf{\Gamma}_{\mathbf{B}}\mathbf{B}^T\big)-\mathrm{tr}\;\big(\mathbf{\Gamma}_{\mathbf{C}}\mathbf{C}\big). 
\end{align*}
And the KKT conditions are:
\begin{equation}
\begin{array}{rr}
\mathbf{B}^*\ge\mathbf{0}, & \mathbf{C}^*\ge\mathbf{0}, \\
\nabla_{\mathbf{B}}J(\mathbf{B}^*)=\mathbf{\Gamma}_{\mathbf{B}}\ge\mathbf{0}, & \nabla_{\mathbf{C}}J(\mathbf{C}^*)=\mathbf{\Gamma}_{\mathbf{C}}^T\ge\mathbf{0},\\
\nabla_{\mathbf{B}}J(\mathbf{B}^*)\odot\mathbf{B}^*=\mathbf{0}, & \nabla_{\mathbf{C}}J(\mathbf{C}^*)\odot\mathbf{C}^*=\mathbf{0}, \label{eq54}
\end{array}
\end{equation}
with
\begin{align*}
\nabla_{\mathbf{B}}J(\mathbf{B})&=\mathbf{BCC}^T-\mathbf{AC}^T, \\
\nabla_{\mathbf{C}}J(\mathbf{C})&=\mathbf{B}^T\mathbf{BC}-\mathbf{B}^T\mathbf{A}+\alpha\mathbf{CC}^T\mathbf{C}-\alpha\mathbf{C}.
\end{align*}
Then, MU algorithm for objective in eq.~\ref{eq52} can be written as:
\begin{align} 
b_{mr} &\longleftarrow b_{mr}\frac{\big(\mathbf{AC}^T\big)_{mr}}{\big(\mathbf{BCC}^T\big)_{mr}}, \label{eq57}\\
c_{rn} &\longleftarrow c_{rn}\frac{\big(\mathbf{B}^T\mathbf{A}+\alpha\mathbf{C}\big)_{rn}}{\big(\mathbf{B}^T\mathbf{BC}+\alpha\mathbf{CC}^T\mathbf{C}\big)_{rn}}. \label{eq58}
\end{align}
The complete algorithm is given in algorithm \ref{algorithm4}. 
\begin{algorithm}
\caption{The MU algorithm for UNMF problem in eq.~\ref{eq52}.}
\label{algorithm4}
\begin{algorithmic}
\STATE Initialization, $\mathbf{B}^0>\mathbf{0}$ and $\mathbf{C}^0>\mathbf{0}$.
\FOR {$k=0,\ldots,K$}
\STATE \begin{align*} b_{mr}^{(k+1)} &\longleftarrow b_{mr}^{k}\frac{\big(\mathbf{AC}^{kT}\big)_{mr}}{\big(\mathbf{B}^{k}\mathbf{C}^{k}\mathbf{C}^{kT}\big)_{mr}+\delta}\;\;\forall m,r \\
c_{rn}^{(k+1)} &\longleftarrow c_{rn}^{k}\frac{\big(\mathbf{B}^{(k+1)T}\mathbf{A}+\alpha\mathbf{C}^{k}\big)_{rn}}{\big(\mathbf{B}^{(k+1)T}\mathbf{B}^{(k+1)}\mathbf{C}^{k}+\alpha\mathbf{C}^{k}\mathbf{C}^{kT}\mathbf{C}^{k}\big)_{rn}+\delta}\;\;\forall r,n \end{align*}
\ENDFOR
\end{algorithmic}
\end{algorithm}

As shown in \cite{CJLin2}, MU algorithm can be modified into an equivalent algorithm with robust convergence guarantee by: 1) transforming MU rules into AU rules, and 2) replacing zero entries that do not satisfy the KKT conditions with small positive number to escape the zero locking. We will employ this strategy to derive converged algorithms for UNMF.

AU version of the algorithm in eq.~\ref{eq57} and \ref{eq58} can be defined as:
\begin{align*} 
b_{mr} &\longleftarrow b_{mr} - \frac{b_{mr}}{\big(\mathbf{BCC}^T\big)_{mr}}\nabla_{\mathbf{B}}J(\mathbf{B})_{mr}, 
\\
c_{rn} &\longleftarrow c_{rn} - \frac{c_{rn}}{\big(\mathbf{B}^T\mathbf{BC}+\alpha\mathbf{CC}^T\mathbf{C}\big)_{rn}}\nabla_{\mathbf{C}}J(\mathbf{C})_{rn}. 
\end{align*}
As shown, this algorithm is equivalent to the algorithm in eq.~\ref{eq57} and \ref{eq58}. By inspection, it is clear that this algorithm inherits the zero locking phenomenon (when $\nabla_{\mathbf{B}}J(\mathbf{B})_{mr}<0$ \& $b_{mr}=0$; or when $\nabla_{\mathbf{C}}J(\mathbf{C})_{rn}<0$ \& $c_{rn}=0$) from its MU version. Therefore a strategy to escape it must be introduced. Algorithm \ref{algorithm5} gives the necessary modifications to avoid the zero locking, where
\begin{align}
\bar{b}_{mr}^k &\equiv \left\{
  \begin{array}{rl}
    b_{mr}^k\hspace{13 mm} & \text{if  } \nabla_{\mathbf{B}}J\big(\mathbf{B}^k,\mathbf{C}^k\big)_{mr} \ge 0 \\
    \max(b_{mr}^k, \sigma) & \text{if  } \nabla_{\mathbf{B}}J\big(\mathbf{B}^k,\mathbf{C}^k\big)_{mr} < 0
  \end{array}, \right. \label{eq65} \\
\bar{c}_{rn}^k &\equiv \left\{
  \begin{array}{rl}
    c_{rn}^k\hspace{13 mm} & \text{if  } \nabla_{\mathbf{C}}J\big(\mathbf{B}^{(k+1)}, \mathbf{C}^k\big)_{rn} \ge 0 \\
    \max(c_{rn}^k, \sigma) & \text{if  } \nabla_{\mathbf{C}}J\big(\mathbf{B}^{(k+1)}, \mathbf{C}^k\big)_{rn} < 0
  \end{array}, \right. \label{eq66}
\end{align}
are the modifications to avoid the zero locking with $\sigma$ is a small positive number, $\mathbf{\bar{B}}$ and $\mathbf{\bar{C}}$ are matrices that contain $\bar{b}_{mr}$ and $\bar{c}_{rn}$ respectively, and
\begin{align*}
\nabla_{\mathbf{B}}J(\mathbf{B}^k,\mathbf{C}^k)&=\mathbf{B}^k\mathbf{C}^k\mathbf{C}^{kT}-\mathbf{AC}^{kT},\\ 
\nabla_{\mathbf{C}}J(\mathbf{B}^{k+1},\mathbf{C}^k)&=\mathbf{B}^{(k+1)T}\mathbf{B}^{(k+1)}\mathbf{C}^k-\mathbf{B}^{(k+1)T}\mathbf{A}+\alpha\mathbf{C}^k\mathbf{C}^{kT}\mathbf{C}^k-\alpha\mathbf{C}^k.
\end{align*}

\begin{algorithm}
\caption{The AU algorithm for UNMF problem in eq.~\ref{eq52}.}
\label{algorithm5}
\begin{algorithmic}
\STATE Initialization, $\mathbf{B}^0\ge\mathbf{0}$ and $\mathbf{C}^0\ge\mathbf{0}$.
\FOR {$k=0,\ldots,K$}
\STATE \begin{align} b_{mr}^{(k+1)} \longleftarrow & \;b_{mr}^{k} - \frac{\bar{b}_{mr}^{k}\times\nabla_{\mathbf{B}}J(\mathbf{B}^k,\mathbf{C}^k)_{mr}}{\big(\mathbf{\bar{B}}^{k}\mathbf{C}^{k}\mathbf{C}^{kT}\big)_{mr}+\delta}\;\;\forall m,r \label{eq69}\\
c_{rn}^{(k+1)} \longleftarrow & \;c_{rn}^{k} - \frac{\bar{c}_{rn}^{k}\times\nabla_{\mathbf{C}}J(\mathbf{B}^{k+1},\mathbf{C}^k)_{rn}}{\big(\mathbf{B}^{(k+1)T}\mathbf{B}^{(k+1)}\mathbf{\bar{C}}^{k}+\alpha\mathbf{\bar{C}}^{k}\mathbf{\bar{C}}^{kT}\mathbf{\bar{C}}^{k}\big)_{rn}+\delta_{\mathbf{C}}^k} \;\;\forall r,n, \label{eq70}
\end{align}
\ENDFOR
\end{algorithmic}
\end{algorithm}

Note that as algorithm \ref{algorithm5} is free from the zero locking, $\mathbf{B}^0$ and $\mathbf{C}^0$ can be initialized with nonnegative matrices. Theorem \ref{theorem5} explains this formally. Also, we have $\delta_{\mathbf{C}}^k$ in eq.~\ref{eq70}. So it is no longer a constant, but a variable that may be different in each iteration. As will be explained later, $\delta_{\mathbf{C}}^k$ plays a crucial role in guaranteeing convergence of the algorithm.

\begin{theorem}\label{theorem5}
If $\mathbf{B}^0>0$ and $\mathbf{C}^0>0$, then $\mathbf{B}^k>0$ and $\mathbf{C}^k>0$, $\forall k\ge 0$. And if $\mathbf{B}^0\ge 0$ and $\mathbf{C}^0 \ge0$, then $\mathbf{B}^k\ge 0$ and $\mathbf{C}^k\ge 0$, $\forall k\ge 0$
\end{theorem}
\begin{proof}
This statement is clear for $k=0$, so we need only to prove for $k>0$.

\noindent \emph{Case 1}: $\nabla_{\mathbf{B}}J_{mr}\ge 0 \Rightarrow \bar{b}_{mr} = b_{mr}$.
\begin{align*}
b_{mr}^{(k+1)} &= \frac{\big(\mathbf{B}^k\mathbf{C}^k\mathbf{C}^{kT}\big)_{mr}b_{mr}^k+\delta b_{mr}^k}{\big(\mathbf{B}^k\mathbf{C}^k\mathbf{C}^{kT}\big)_{mr}+\delta}-\frac{\big(\mathbf{B}^k\mathbf{C}^k\mathbf{C}^{kT}\big)_{mr}b_{mr}^k-\big(\mathbf{AC}^{kT}\big)_{mr}b_{mr}^k}{\big(\mathbf{B}^k\mathbf{C}^k\mathbf{C}^{kT}\big)_{mr}+\delta} \\
&=\frac{\big[\big(\mathbf{AC}^{kT}\big)_{mr}+\delta\big]b_{mr}^k}{\big(\mathbf{B}^k\mathbf{C}^k\mathbf{C}^{kT}\big)_{mr}+\delta}.
\end{align*}
Thus, if $b_{mr}^k>0$ then $b_{mr}^{(k+1)}>0\;\forall m,r$, and if $b_{mr}^k\ge 0$ then $b_{mr}^{(k+1)}\ge 0\;\forall m,r,\;\;\forall k>0$.

\noindent \emph{Case 2}: $\nabla_{\mathbf{B}}J_{mr}<0 \Rightarrow \bar{b}_{mr} \ne b_{mr}$.
\begin{align*}
b_{mr}^{(k+1)} = b_{mr}^k-\frac{\max\big(b_{mr}^k,\sigma\big)\nabla_{\mathbf{B}}J\big(\mathbf{B}^k,\mathbf{C}^k\big)_{mr}}{\big(\mathbf{\bar{B}}^k\mathbf{C}^k\mathbf{C}^{kT}\big)_{mr}+\delta}.
\end{align*}
Note that $\max\big(b_{mr}^k,\sigma\big)>0$ and $\nabla_{\mathbf{B}}J\big(\mathbf{B}^k,\mathbf{C}^k\big)_{mr}<0$. Thus if $b_{mr}^k>0$ then $b_{mr}^{(k+1)}>0\;\forall m,r$, and if $b_{mr}^k\ge 0$ then $b_{mr}^{(k+1)}>0\;\forall m,r,\;\;\forall k>0$.

\noindent \emph{Case 3}: $\nabla_{\mathbf{C}}J_{rn}\ge 0 \Rightarrow \bar{c}_{rn} = c_{rn}$.
\begin{align*}
c_{rn}^{(k+1)} = &\frac{\big(\mathbf{B}^{(k+1)T}\mathbf{B}^{(k+1)}\mathbf{C}^{k}+\alpha\mathbf{C}^{k}\mathbf{C}^{kT}\mathbf{C}^{k}\big)_{rn}c_{rn}^k + \delta_{\mathbf{C}}^k c_{rn}^k}{\big(\mathbf{B}^{(k+1)T}\mathbf{B}^{(k+1)}\mathbf{C}^{k}+\alpha\mathbf{C}^{k}\mathbf{C}^{kT}\mathbf{C}^{k}\big)_{rn} + \delta_{\mathbf{C}}^k} - \\
&\frac{\big(\mathbf{B}^{(k+1)T}\mathbf{B}^{(k+1)}\mathbf{C}^{k}+\alpha\mathbf{C}^{k}\mathbf{C}^{kT}\mathbf{C}^{k}\big)_{rn}c_{rn}^k}{\big(\mathbf{B}^{(k+1)T}\mathbf{B}^{(k+1)}\mathbf{C}^{k}+\alpha\mathbf{C}^{k}\mathbf{C}^{kT}\mathbf{C}^{k}\big)_{rn} + \delta_{\mathbf{C}}^k} + \\
&\frac{\big(\mathbf{B}^{(k+1)T}\mathbf{A}+\alpha\mathbf{C}^k\big)_{rn} c_{rn}^k}{\big(\mathbf{B}^{(k+1)T}\mathbf{B}^{(k+1)}\mathbf{C}^{k}+\alpha\mathbf{C}^{k}\mathbf{C}^{kT}\mathbf{C}^{k}\big)_{rn} + \delta_{\mathbf{C}}^k} \\
=&\frac{\big[\big(\mathbf{B}^{(k+1)T}\mathbf{A}+\alpha\mathbf{C}^k\big)_{rn} + \delta_{\mathbf{C}}^k\big]c_{rn}^k}{\big(\mathbf{B}^{(k+1)T}\mathbf{B}^{(k+1)}\mathbf{C}^{k}+\alpha\mathbf{C}^{k}\mathbf{C}^{kT}\mathbf{C}^{k}\big)_{rn} + \delta_{\mathbf{C}}^k},
\end{align*}
Thus if $c_{rn}^k>0$ then $c_{rn}^{(k+1)}>0$, and if $c_{rn}^k\ge 0\;\forall r,n$ then $c_{rn}^{(k+1)}\ge 0\;\forall r,n,\;\;\forall k>0$.

\noindent \emph{Case 4}: $\nabla_{\mathbf{C}}J_{rn}<0 \Rightarrow \bar{c}_{rn} \ne c_{rn}$.
\begin{align*}
c_{rn}^{(k+1)} = c_{rn}^k-\frac{\max\big(c_{rn}^k,\sigma\big)\nabla_{\mathbf{C}}J\big(\mathbf{B}^{(k+1)},\mathbf{C}^k\big)_{rn}}{\big(\mathbf{B}^{(k+1)T}\mathbf{B}^{(k+1)}\mathbf{\bar{C}}^{k}+\alpha\mathbf{\bar{C}}^{k}\mathbf{\bar{C}}^{kT}\mathbf{\bar{C}}^{k}\big)_{rn} + \delta_{\mathbf{C}}^k}.
\end{align*}
Note that $\max\big(c_{rn}^k,\sigma\big)>0$ and $\nabla_{\mathbf{C}}J\big(\mathbf{B}^{(k+1)},\mathbf{C}^k\big)_{rn}<0$. Thus if $c_{rn}^k>0$ then $c_{rn}^{(k+1)}>0\;\forall r,n$, and if $c_{rn}^k\ge 0$ then $c_{rn}^{(k+1)}>0\;\forall r,n,\;\;\forall k>0$.

By combining results for $k=0$ and $k>0$ in case 1-4, the proof is completed.
\end{proof}

\subsubsection{Convergence analysis}

To analyze convergence property of algorithm \ref{algorithm5}, the nonincreasing property will be shown first as it is the necessary condition for the convergence. Because the algorithm solves the problem in alternating fashion, i.e., fixing one variable while solving the other, sequence $J\big(\mathbf{B}^k\big)$ and $J\big(\mathbf{C}^k\big)$ can be analyzed separately. Thus, by showing that:
\begin{align}
J\big(\mathbf{B}^{(k+1)}\big) &\le J\big(\mathbf{B}^k\big) \;\;\mathrm{and} \label{eq71}\\
J\big(\mathbf{C}^{(k+1)}\big) &\le J\big(\mathbf{C}^k\big),\;\forall k\ge 0, \label{eq72}
\end{align}
the nonincreasing property of algorithm \ref{algorithm5}, i.e., $J\big(\mathbf{B}^{(k+1)}$,$\mathbf{C}^{(k+1)}\big)$ $\le$ $J\big(\mathbf{B}^{(k+1)}$,$\mathbf{C}^k\big)$ $\le$ $J\big(\mathbf{B}^k$,$\mathbf{C}^k\big)$, will be proven.

\paragraph{A.~The nonincreasing property of $J\big(\mathbf{B}^k\big)$}

The nonincreasing property of sequence $J\big(\mathbf{B}^k\big)$ of algorithm \ref{algorithm5} (eq.~\ref{eq71}) has been proven by Lin \cite{CJLin2}. Here we will describe his proof in accord to our more general approach.

So far, there is no method to directly prove $J\big(\mathbf{B}^{(k+1)}\big) \le J\big(\mathbf{B}^k\big)$. Fortunately, the auxiliary function approach \cite{Lee2} can be utilized as an intermediate function: 
\begin{equation*}
J\big(\mathbf{B}^{(k+1)}\big)= G\big(\mathbf{B}^{(k+1)},\mathbf{B}^{(k+1)}\big) \le G\big(\mathbf{B}^{(k+1)},\mathbf{B}^k\big) \le G\big(\mathbf{B}^k,\mathbf{B}^k\big) = J\big(\mathbf{B}^k\big).
\end{equation*}
To define $G$, let's rearrange $\mathbf{B}$ into:
\begin{equation*}
\mathfrak{B}^T \equiv
\begin{bmatrix}
\mathfrak{b}_1^T & & & \\
& \mathfrak{b}_2^T & & \\
& & \ddots & \\
& & & \mathfrak{b}_M^T
\end{bmatrix}\in\mathbb{R}_+^{MR\times M},
\end{equation*}
where $\mathfrak{b}_m$ is the $m$-th row of $\mathbf{B}$. And also let's define:
\begin{equation*}
\nabla_{\mathfrak{B}^T}\mathfrak{J}\big(\mathfrak{B}^{kT}\big) \equiv
\begin{bmatrix}
\nabla_{\mathbf{B}}\mathfrak{J}\big(\mathbf{B}^k\big)_1^T & & & \\
& \nabla_{\mathbf{B}}\mathfrak{J}\big(\mathbf{B}^k\big)_2^T & & \\
& & \ddots & \\
& & & \nabla_{\mathbf{B}}\mathfrak{J}\big(\mathbf{B}^k\big)_M^T
\end{bmatrix}\in\mathbb{R}_+^{MR\times M},
\end{equation*}
where $\nabla_{\mathbf{B}}\mathfrak{J}\big(\mathbf{B}^k\big)_m$ is the $m$-th row of $\nabla_{\mathbf{B}}J(\mathbf{B}^k)= \mathbf{B}^k\mathbf{C}^k\mathbf{C}^{kT} - \mathbf{AC}^{kT}$. Then define:
\begin{equation*}
\mathbf{D} \equiv \mathrm{diag}\;\big(\mathbf{D}^1,\ldots,\mathbf{D}^M\big)\in\mathbb{R}_+^{MR\times MR},
\end{equation*}
where $\mathbf{D}^m$ is a diagonal matrix with its diagonal entries defined as:
\begin{equation*}
d_{rr}^m \equiv \left\{
 \begin{array}{ll}
   \frac{\big(\mathbf{\bar{B}}^k\mathbf{C}^k\mathbf{C}^{kT}\big)_{mr}+\delta}{\bar{b}_{mr}^k} & \mathrm{if}\;\; r\in \mathcal{I}_m \\
   \star & \mathrm{if}\;\; r\notin \mathcal{I}_m
 \end{array} \right.
\end{equation*}
with
\begin{align*}
\mathcal{I}_m \equiv \big\{r|&b_{mr}^k>0,\;\nabla_{\mathbf{B}}J\big(\mathbf{B}^k\big)_{mr}\ne 0,\;\mathrm{or} \\
&b_{mr}^k=0,\;\nabla_{\mathbf{B}}J\big(\mathbf{B}^k\big)_{mr} < 0\big\}
\end{align*}
is the set of non-KKT indices in $m$-th row of $\mathbf{B}^k$, and $\star$ is defined so that $\star \equiv 0$ and $\star^{-1} \equiv 0$.

Then, the auxiliary function $\mathfrak{G}$ can be defined as:
\begin{align}
\mathfrak{G}\big(\mathfrak{B}^T,\mathfrak{B}^{kT}\big) \equiv \;&\mathfrak{J}\big(\mathfrak{B}^{kT}\big) + \mathrm{tr}\;\big\{\big(\mathfrak{B}-\mathfrak{B}^k\big)\nabla_{\mathfrak{B}^T}\mathfrak{J}\big(\mathfrak{B}^{kT}\big)\big\} \nonumber \\
&+ \frac{1}{2}\mathrm{tr}\;\big\{\big(\mathfrak{B}-\mathfrak{B}^k\big)\mathbf{D}\big(\mathfrak{B}-\mathfrak{B}^k\big)^T\big\}. \label{eq75}
\end{align}
Note that $\mathfrak{J}$ and $\mathfrak{G}$ are equivalent to $J$ and $G$ with $\mathbf{B}$ is rearranged into $\mathfrak{B}^T$, and other parameters are reordered accordingly (one can still use $J$ and $G$, but it won't be as compact as our approach), and also whenever $\mathbf{X}^{(k+1)}$ is a variable, we remove $(k+1)$ sign. And:
\begin{equation*}
\nabla_{\mathfrak{B}^T}\mathfrak{G}\big(\mathfrak{B}^T,\mathfrak{B}^{kT}\big)=\mathbf{D}\big(\mathfrak{B}-\mathfrak{B}^k\big)^T + \nabla_{\mathfrak{B}^T}\mathfrak{J}\big(\mathfrak{B}^{kT}\big).
\end{equation*}
By definition, $\mathbf{D}$ is positive definite for all $\mathbf{B}^k$ not satisfy the KKT conditions and positive semidefinite if and only if $\mathbf{B}^k$ satisfies the KKT conditions. Thus $\mathfrak{G}\big(\mathfrak{B}^T,\mathfrak{B}^{kT}\big)$ is a strict convex function, and consequently has a unique minimum, so that:
\begin{align}
\mathbf{D}\big(\mathfrak{B}-\mathfrak{B}^k\big)^T + \nabla_{\mathfrak{B}^T}\mathfrak{J}\big(\mathfrak{B}^{kT}\big)=0, \label{eq76}\\
\mathfrak{B}^T = \mathfrak{B}^{kT} - \mathbf{D}^{-1}\nabla_{\mathfrak{B}^T}\mathfrak{J}\big(\mathfrak{B}^{kT}\big), \nonumber
\end{align}
which is exactly the update rule for $\mathbf{B}$ in eq.~\ref{eq69}. 

To obtain an alternative formulation for $\mathfrak{J}\big(\mathfrak{B}^T\big)$ that in the same fashion with $\mathfrak{G}$ formulation, the Taylor series expansion is used.
\begin{align}
\mathfrak{J}\big(\mathfrak{B}^T\big) = \;&\mathfrak{J}\big(\mathfrak{B}^{kT}\big) + \mathrm{tr}\;\big\{\big(\mathfrak{B}-\mathfrak{B}^k\big)\nabla_{\mathfrak{B}^T}\mathfrak{J}\big(\mathfrak{B}^{kT}\big)\big\} \nonumber \\
&+ \frac{1}{2}\mathrm{tr}\;\big\{\big(\mathfrak{B}-\mathfrak{B}^k\big)\nabla_{\mathbf{B}}^2\mathbf{J}\big(\mathbf{B}^k\big)\big(\mathfrak{B}-\mathfrak{B}^k\big)^T\big\}. \label{eq77}
\end{align}
where
\begin{equation*}
\nabla_{\mathbf{B}}^2\mathbf{J}\big(\mathbf{B}^k\big) \equiv
\begin{bmatrix}
\nabla_{\mathbf{B}}^2 J \big(\mathbf{B}^k\big) & & \\
& \ddots & \\
& & \nabla_{\mathbf{B}}^2 J \big(\mathbf{B}^k\big)
\end{bmatrix}\in\mathbb{R}_+^{MR\times MR}
\end{equation*}
with $\nabla_{\mathbf{B}}^2 J \big(\mathbf{B}^k\big)=\mathbf{C}^k\mathbf{C}^{kT}$ components are arranged along its diagonal area (there are $M$ components).

Then, for $\mathfrak{G}$ to be the auxiliary function, we must prove:
\begin{enumerate}
\item $\mathfrak{G}\big(\mathfrak{B}^T,\mathfrak{B}^T\big)=\mathfrak{J}\big(\mathfrak{B}^T\big)$,
\item $\mathfrak{G}\big(\mathfrak{B}^{kT},\mathfrak{B}^{kT}\big)=\mathfrak{J}\big(\mathfrak{B}^{kT}\big)$,
\item $\mathfrak{G}\big(\mathfrak{B}^T,\mathfrak{B}^T\big) \le \mathfrak{G}\big(\mathfrak{B}^T,\mathfrak{B}^{kT}\big)$, and
\item $\mathfrak{G}\big(\mathfrak{B}^T,\mathfrak{B}^{kT}\big) \le \mathfrak{G}\big(\mathfrak{B}^{kT},\mathfrak{B}^{kT}\big)$,
\end{enumerate}
so that $\mathfrak{J}\big(\mathfrak{B}^T\big) \le \mathfrak{J}\big(\mathfrak{B}^{kT}\big)$. Because $\mathfrak{B}$ is equivalent to $\mathbf{B}$ with reordered rows, this implies $J\big(\mathbf{B}^{(k+1)}\big) \le J\big(\mathbf{B}^k\big)$, which is the nonincreasing property of the sequence $J\big(\mathbf{B}^k\big)$. The first and second will be proven in theorem \ref{theorem6}, the third in theorem \ref{theorem7}, and the fourth in theorem \ref{theorem8}.
\begin{theorem} \label{theorem6}
$\mathfrak{G}\big(\mathfrak{B}^T,\mathfrak{B}^T\big)=\mathfrak{J}\big(\mathfrak{B}^T\big)$ and $\mathfrak{G}\big(\mathfrak{B}^{kT},\mathfrak{B}^{kT}\big)=\mathfrak{J}\big(\mathfrak{B}^{kT}\big)$.
\end{theorem}
\begin{proof}
These are obvious from the definition of $\mathfrak{G}$ in eq.~\ref{eq75}.
\end{proof}
\begin{theorem} \label{theorem7}
$\mathfrak{G}\big(\mathfrak{B}^T,\mathfrak{B}^T\big) \le \mathfrak{G}\big(\mathfrak{B}^T,\mathfrak{B}^{kT}\big)$. Moreover if and only if $\mathbf{B}^k$ satisfies the KKT conditions in eq.~\ref{eq54}, then $\mathfrak{G}\big(\mathfrak{B}^T,\mathfrak{B}^T\big) = \mathfrak{G}\big(\mathfrak{B}^T,\mathfrak{B}^{kT}\big)$.
\end{theorem}
\begin{proof}
By substracting eq.~\ref{eq75} from eq.~\ref{eq77}, we get:
\begin{align*}
\mathfrak{G}\big(\mathfrak{B}^T,\mathfrak{B}^{kT}\big)-\mathfrak{G}\big(\mathfrak{B}^T,\mathfrak{B}^T\big)&=\frac{1}{2}\,\mathrm{tr}\,\big\{\big(\mathfrak{B}-\mathfrak{B}^k\big)\big(\mathbf{D}-\nabla_{\mathbf{B}}^2\mathbf{J}\big(\mathbf{B}^k\big)\big)\big(\mathfrak{B}-\mathfrak{B}^k\big)^T\big\} 
\\
&=\frac{1}{2}\sum_{m=1}^M\left[\big(\mathfrak{b}_m-\mathfrak{b}_m^k\big)\big(\mathbf{D}^m-\nabla_{\mathbf{B}}^2J\big(\mathbf{B}^k\big)\big)\big(\mathfrak{b}_m-\mathfrak{b}_m^k\big)^T\right] 
\end{align*}
If $\mathbf{D}^m-\nabla_{\mathbf{B}}^2J\big(\mathbf{B}^k\big)\;\,\forall m$ are all positive definite, then the inequality always holds except when $\mathfrak{b}_m=\mathfrak{b}_m^k\;\,\forall m$. Thus, it is sufficient to prove the positive definiteness of $\mathbf{D}^m-\nabla_{\mathbf{B}}^2J\big(\mathbf{B}^k\big)\;\,\forall m$.

Let $\mathbf{v}_m^T=\mathfrak{b}_m-\mathfrak{b}_m^k\ne \mathbf{0}$, then we must prove:
\begin{equation*}
\mathbf{v}_m^T\big(\mathbf{D}^m-\nabla_{\mathbf{B}}^2J\big(\mathbf{B}^k\big)\big)\mathbf{v}_m > 0.
\end{equation*}
Note that
\begin{equation*}
d_{rr}^m \equiv \left\{
 \begin{array}{ll}
   \frac{\big(\mathbf{\bar{b}}_m^k\mathbf{X}^k\big)_{mr}+\delta}{\bar{b}_{mr}^k} & \mathrm{if}\;\; r\in \mathcal{I}_m \\
   \star & \mathrm{if}\;\; r\notin \mathcal{I}_m
 \end{array} \right.
\end{equation*}
with $\mathbf{X}^k=\mathbf{C}^k\mathbf{C}^{kT}=\nabla_{\mathbf{B}}^2 J \big(\mathbf{B}^k\big)$ and $\mathbf{D}^m$ are symmetric. Thus,
\begin{align*}
\mathbf{v}_m^T\big(\mathbf{D}^m-\nabla_{\mathbf{B}}^2J\big(\mathbf{B}^k\big)\big)\mathbf{v}_m =& \sum_{r=1}^R v_r^2 \frac{\delta}{\bar{b}_{mr}^k} + \sum_{r=1}^R v_r^2 \frac{\big(\mathbf{X}^k \mathbf{\bar{b}}_m^{kT}\big)_{mr}}{\bar{b}_{mr}^k} - \sum_{r,s=1}^R v_r v_s x_{rs}. \\
>& \sum_{r=1}^R v_r^2 \frac{\sum_{s=1}^R x_{rs} \big(\mathbf{\bar{b}}_m^k\big)_s}{\bar{b}_{mr}^k} - \sum_{r=1}^R \sum_{s=1}^R v_r v_s x_{rs} \\
=& \frac{1}{2} \sum_{r=1}^R \sum_{s=1}^R v_r^2 \frac{ x_{rs}\big(\mathbf{\bar{b}}_m^k\big)_s } {\bar{b}_{mr}^k} + \frac{1}{2} \sum_{r=1}^R \sum_{s=1}^R v_s^2 \frac{ x_{sr}\big(\mathbf{\bar{b}}_m^k\big)_r } {\bar{b}_{ms}^k} \\
&- \sum_{r=1}^R \sum_{s=1}^R v_r v_s x_{rs} \\
=& \frac{1}{2} \sum_{r=1}^R \sum_{s=1}^R x_{rs} \left(\sqrt{ \frac{ \bar{b}_{ms}^k }{ \bar{b}_{mr}^k } }v_r - \sqrt{ \frac{ \bar{b}_{mr}^k }{ \bar{b}_{ms}^k } }v_s \right)^2 \ge 0
\end{align*}
where $v_r$ is the $r$-th entry of $\mathbf{v}_m$ and $x_{rs}$ is the $(r,s)$ entry of $\mathbf{X}$. Therefore, $\mathbf{D}^m-\nabla_{\mathbf{B}}^2J\big(\mathbf{B}^k\big)\;\,\forall m$ are positive definite, and consequently the equality happens if and only if $\mathbf{B}=\mathbf{B}^k$ which by the update rule in eq.~\ref{eq69} and the boundedness theorem \ref{theorem16} happens if and only if $\mathbf{B}^k$ satisfies the KKT conditions.
\end{proof}

\begin{theorem} \label{theorem8}
$\mathfrak{G}\big(\mathfrak{B}^T,\mathfrak{B}^{kT}\big) \le \mathfrak{G}\big(\mathfrak{B}^{kT},\mathfrak{B}^{kT}\big)$. Moreover, if and only if $\mathbf{B}$ satisfies the KKT conditions in eq.~\ref{eq54}, then $\mathfrak{G}\big(\mathfrak{B}^T,\mathfrak{B}^{kT}\big) = \mathfrak{G}\big(\mathfrak{B}^{kT},\mathfrak{B}^{kT}\big)$.
\end{theorem}
\begin{proof}
\begin{align*}
\mathfrak{G}\big(\mathfrak{B}^{kT},\mathfrak{B}^{kT}\big)-\mathfrak{G}\big(\mathfrak{B}^T,\mathfrak{B}^{kT}\big) = &-\mathrm{tr}\;\big\{\big(\mathfrak{B}-\mathfrak{B}^k\big)\nabla_{\mathfrak{B}^T}\mathfrak{J}\big(\mathfrak{B}^{kT}\big)\big\} \\
&- \frac{1}{2}\mathrm{tr}\;\big\{\big(\mathfrak{B}-\mathfrak{B}^k\big)\mathbf{D}\big(\mathfrak{B}-\mathfrak{B}^k\big)^T\big\}.
\end{align*}
By using eq.~\ref{eq76}, and the fact that $\mathbf{D}$ is positive semi-definite:
\begin{equation*}
\mathfrak{G}\big(\mathfrak{B}^{kT},\mathfrak{B}^{kT}\big)-\mathfrak{G}\big(\mathfrak{B}^T,\mathfrak{B}^{kT}\big) = \frac{1}{2}\mathrm{tr}\;\big\{\big(\mathfrak{B}-\mathfrak{B}^k\big)\mathbf{D}\big(\mathfrak{B}-\mathfrak{B}^k\big)^T\big\} \ge 0, 
\end{equation*}
we proved that $\mathfrak{G}\big(\mathfrak{B}^T,\mathfrak{B}^{kT}\big) \le \mathfrak{G}\big(\mathfrak{B}^{kT},\mathfrak{B}^{kT}\big)$. Now, let's prove the second part of the theorem. By the update rule eq.~\ref{eq69}, if $\mathbf{B}^k$ satisfies the KKT conditions, then $\mathbf{B}$ will be equal to $\mathbf{B}^k$, and thus the equality holds. Now we need to prove that if the equality holds, then $\mathbf{B}^k$ satisfies the KKT conditions.

To prove this, let consider a contradiction situation where the equality holds but $\mathbf{B}^k$ does not satisfy the KKT conditions. In this case, there exists at least an index $(m,r)$ such that:
\begin{equation*}
b_{mr}\ne b_{mr}^k\;\;\mathrm{and}\;\; d_{rr}^m = \frac{\big(\mathbf{\bar{b}}_m^k\mathbf{X}^k\big)_{mr}+\delta}{\bar{b}_{mr}^k}\ge\frac{\delta}{\bar{b}_{mr}^k}. 
\end{equation*}
Note that by the definition in eq.~\ref{eq65}, if $\bar{b}_{mr}^k$ is equal to zero, then it satisfies the KKT conditions. Accordingly, $b_{mr}= b_{mr}^k$ which violates the condition for the contradiction. So, $\bar{b}_{mr}^k$ cannot be equal to zero, and thus $d_{rr}^m$ is well defined. Consequently,
\begin{equation*}
\mathfrak{G}\big(\mathfrak{B}^{kT},\mathfrak{B}^{kT}\big)-\mathfrak{G}\big(\mathfrak{B}^T,\mathfrak{B}^{kT}\big) \ge \frac{\big(b_{mr}-b_{mr}^k\big)^2\delta}{\bar{b}_{mr}^k} > 0,
\end{equation*}
which violates the equality. Thus, it is proven that if the equality holds, then $\mathbf{B}^k$ satisfies the KKT conditions.
\end{proof}

The following theorem summarizes the nonincreasing property of $J\big(\mathbf{B}^k\big)$.
\begin{theorem}\label{theorem9}
$J\big(\mathbf{B}^{k+1}\big) \le J\big(\mathbf{B}^k\big)\;\,\forall k\ge 0$ under update rule eq.~\ref{eq69} with the equality happens if and only if $\mathbf{B}^k$ satisfies the KKT conditions in eq.~\ref{eq54}. 
\end{theorem}
\begin{proof}
This theorem is the corollary of theorem \ref{theorem6}, \ref{theorem7}, and \ref{theorem8}
\end{proof}

\paragraph{B.~The nonincreasing property of $J\big(\mathbf{C}^k\big)$}

Now we prove the nonincreasing property of $J\big(\mathbf{C}^k\big)$, i.e., eq.~\ref{eq72}: $J\big(\mathbf{C}^{(k+1)}\big)\le J\big(\mathbf{C}^k\big)\;\forall k\ge 0$. Note that to prove this, $\mathbf{B}^k$ and $\mathbf{C}^k$ must be bounded. The boundedness of $\mathbf{B}^k$ and $\mathbf{C}^k$ will be proven in theorem \ref{theorem16}.

By using the auxiliary function approach, the nonincreasing property of $J\big(\mathbf{C}^k\big)$ can be proven by showing that:
\begin{equation*}
J\big(\mathbf{C}^{(k+1)}\big)= G\big(\mathbf{C}^{(k+1)},\mathbf{C}^{(k+1)}\big) \le G\big(\mathbf{C}^{(k+1)},\mathbf{C}^k\big) \le G\big(\mathbf{C}^k,\mathbf{C}^k\big) = J\big(\mathbf{C}^k\big). 
\end{equation*}
To define auxiliary function $G$, $\mathbf{C}$ is rearranged into:
\begin{equation*}
\mathfrak{C} \equiv
\begin{bmatrix}
\mathbf{c}_1 & & & \\
& \mathbf{c}_2 & & \\
& & \ddots & \\
& & & \mathbf{c}_N
\end{bmatrix}\in\mathbb{R}_+^{NR\times N},
\end{equation*}
where $\mathbf{c}_n$ is the $n$-th column of $\mathbf{C}$. And also let's define:
\begin{equation*}
\nabla_{\mathfrak{C}}\mathfrak{J}\big(\mathfrak{C}^k\big) \equiv
\begin{bmatrix}
\nabla_{\mathbf{C}}\mathfrak{J}\big(\mathbf{C}^k\big)_1 & & & \\
& \nabla_{\mathbf{C}}\mathfrak{J}\big(\mathbf{C}^k\big)_2 & & \\
& & \ddots & \\
& & & \nabla_{\mathbf{C}}\mathfrak{J}\big(\mathbf{C}^k\big)_N
\end{bmatrix}\in\mathbb{R}_+^{NR\times N},
\end{equation*}
where $\nabla_{\mathbf{C}}\mathfrak{J}\big(\mathbf{C}^k\big)_n$ is the $n$-th column of $\nabla_{\mathbf{C}}J(\mathbf{C}^k)=\mathbf{B}^{(k+1)T}\mathbf{B}^{(k+1)}\mathbf{C}^k-\mathbf{B}^{(k+1)T}\mathbf{A}+\alpha \mathbf{C}^k\mathbf{C}^{kT}\mathbf{C}^k-\alpha \mathbf{C}^k$. And:
\begin{equation*}
\mathbf{D} \equiv \mathrm{diag}\;\big(\mathbf{D}^1,\ldots,\mathbf{D}^N\big)\in\mathbb{R}_+^{NR\times NR}, 
\end{equation*}
where $\mathbf{D}^n$ is a diagonal matrix with its diagonal entries defined as:
\begin{equation*}
d_{rr}^n \equiv \left\{
 \begin{array}{ll}
   \frac{\big(\mathbf{B}^{(k+1)T}\mathbf{B}^{(k+1)}\mathbf{\bar{C}}^k + \alpha \mathbf{\bar{C}}^k\mathbf{\bar{C}}^{kT}\mathbf{\bar{C}}^k\big)_{rn}+\delta_{\mathbf{C}}^k}{\bar{c}_{rn}^k} & \mathrm{if}\;\; r\in \mathcal{I}_n \\
   \star & \mathrm{if}\;\; r\notin \mathcal{I}_n
 \end{array} \right.
\end{equation*}
with
\begin{align*}
\mathcal{I}_n \equiv \big\{r|&c_{rn}^k>0,\;\nabla_{\mathbf{C}}J\big(\mathbf{C}^k\big)_{rn}\ne 0,\;\mathrm{or} \\
&c_{rn}^k=0,\;\nabla_{\mathbf{C}}J\big(\mathbf{C}^k\big)_{rn} < 0\big\}
\end{align*}
is the set of non-KKT indices in $n$-th column of $\mathbf{C}^k$, and $\star$ is defined as before.

Then, the auxiliary function $\mathfrak{G}$ can be written as:
\begin{equation}
\mathfrak{G}\big(\mathfrak{C},\mathfrak{C}^k\big) \equiv \;\mathfrak{J}\big(\mathfrak{C}^k\big) + \mathrm{tr}\;\big\{\big(\mathfrak{C}-\mathfrak{C}^k\big)^T\nabla_{\mathfrak{C}}\mathfrak{J}\big(\mathfrak{C}^k\big)\big\} + \frac{1}{2}\mathrm{tr}\;\big\{\big(\mathfrak{C}-\mathfrak{C}^k\big)^T\mathbf{D}\big(\mathfrak{C}-\mathfrak{C}^k\big)\big\}. \label{eq80}
\end{equation}
Also:
\begin{equation*}
\nabla_{\mathfrak{C}}\mathfrak{G}\big(\mathfrak{C},\mathfrak{C}^k\big)=\mathbf{D}\big(\mathfrak{C}-\mathfrak{C}^k\big) + \nabla_{\mathfrak{C}}\mathfrak{J}\big(\mathfrak{C}^k\big).
\end{equation*}
Since $\mathbf{D}$ here is equivalent to $\mathbf{D}$ in $J\big(\mathbf{B}^k\big)$ case, $\mathfrak{G}\big(\mathfrak{C},\mathfrak{C}^k\big)$ is a strict convex function, and consequently has a unique minimum, so that:
\begin{align}
\mathbf{D}\big(\mathfrak{C}-\mathfrak{C}^k\big) + \nabla_{\mathfrak{C}}\mathfrak{J}\big(\mathfrak{C}^k\big)=0, \label{eq81}\\
\mathfrak{C} = \mathfrak{C}^k - \mathbf{D}^{-1}\nabla_{\mathfrak{C}}\mathfrak{J}\big(\mathfrak{C}^k\big), \nonumber
\end{align}
which is exactly the update rule for $\mathbf{C}$ in eq.~\ref{eq70}.

By using the Taylor series, alternative formulation for $\mathfrak{J}\big(\mathfrak{C}\big)$ can be written as:
\begin{align}
\mathfrak{J}\big(\mathfrak{C}\big) = &\;\mathfrak{J}\big(\mathfrak{C}^k\big) + \mathrm{tr}\;\big\{\big(\mathfrak{C}-\mathfrak{C}^k\big)^T\nabla_{\mathfrak{C}}\mathfrak{J}\big(\mathfrak{C}^k\big)\big\} + \nonumber \\
&\;\frac{1}{2}\mathrm{tr}\;\big\{\big(\mathfrak{C}-\mathfrak{C}^k\big)^T\nabla_{\mathbf{C}}^2\mathbf{J}\big(\mathbf{C}^k\big)\big(\mathfrak{C}-\mathfrak{C}^k\big)\big\} + \mathbf{\varepsilon}_{\mathbf{C}}^k \label{eq82}
\end{align}
where $\mathbf{\varepsilon}_{\mathbf{C}}^k$ is the higher components of the Taylor series:
\begin{align*}
\mathbf{\varepsilon}_{\mathbf{C}}^k = &\;\frac{1}{6}\mathrm{tr}\;\big\{\big(\mathfrak{C}-\mathfrak{C}^k\big)^T(6\alpha\mathfrak{C}^k\big)\big(\mathfrak{C}-\mathfrak{C}^k\big)^T\big(\mathfrak{C}-\mathfrak{C}^k\big)\big\} + \\
&\;\frac{1}{24}\mathrm{tr}\;\big\{\big(\mathfrak{C}-\mathfrak{C}^k\big)^T\big(\mathfrak{C}-\mathfrak{C}^k\big)(6\alpha\mathbf{I})\big(\mathfrak{C}-\mathfrak{C}^k\big)^T\big(\mathfrak{C}-\mathfrak{C}^k\big)\big\},
\end{align*}
and
\begin{equation*}
\nabla_{\mathbf{C}}^2\mathbf{J}\big(\mathbf{C}^k\big) \equiv
\begin{bmatrix}
\nabla_{\mathbf{C}}^2 J \big(\mathbf{C}^k\big) & & \\
& \ddots & \\
& & \nabla_{\mathbf{C}}^2 J \big(\mathbf{C}^k\big)
\end{bmatrix}\in\mathbb{R}_+^{NR\times NR}
\end{equation*}
with $\nabla_{\mathbf{C}}^2 J \big(\mathbf{C}^k\big)=\mathbf{B}^{(k+1)T}\mathbf{B}^{(k+1)} + 3\alpha\mathbf{C}^k\mathbf{C}^{kT}-\alpha\mathbf{I}$ components are arranged along its diagonal area (there are $N$ components).

As before, for $\mathfrak{G}$ to be the auxiliary function, we must prove:
\begin{enumerate}
\item $\mathfrak{G}\big(\mathfrak{C},\mathfrak{C}\big)=\mathfrak{J}\big(\mathfrak{C}\big)$,
\item $\mathfrak{G}\big(\mathfrak{C}^k,\mathfrak{C}^k\big)=\mathfrak{J}\big(\mathfrak{C}^k\big)$,
\item $\mathfrak{G}\big(\mathfrak{C},\mathfrak{C}\big) \le \mathfrak{G}\big(\mathfrak{C},\mathfrak{C}^k\big)$, and
\item $\mathfrak{G}\big(\mathfrak{C},\mathfrak{C}^k\big) \le \mathfrak{G}\big(\mathfrak{C}^k,\mathfrak{C}^k\big)$,
\end{enumerate}
The first and second will be proven in theorem \ref{theorem10}, the third in theorem \ref{theorem11}, and the fourth in theorem \ref{theorem12}.
\begin{theorem} \label{theorem10}
$\mathfrak{G}\big(\mathfrak{C},\mathfrak{C}\big)=\mathfrak{J}\big(\mathfrak{C}\big)$, and $\mathfrak{G}\big(\mathfrak{C}^k,\mathfrak{C}^k\big)=\mathfrak{J}\big(\mathfrak{C}^k\big)$,
\end{theorem}
\begin{proof}
These are obvious from the definition of $\mathfrak{G}$ in eq.~\ref{eq80}.
\end{proof}
\begin{theorem} \label{theorem11}
Given sufficiently large $\delta_{\mathbf{C}}^k$ and the boundedness of $\mathbf{B}^k$ and $\mathbf{C}^k$, then it can be shown that $\mathfrak{G}\big(\mathfrak{C},\mathfrak{C}\big) \le \mathfrak{G}\big(\mathfrak{C},\mathfrak{C}^k\big)$. Moreover, if and only if $\mathbf{C}^k$ satisfies the KKT conditions, then the equality holds.
\end{theorem}
\begin{proof}
As $\mathfrak{G}\big(\mathfrak{C},\mathfrak{C}\big)=\mathfrak{J}\big(\mathfrak{C}\big)$, we need to show that $\mathfrak{G}\big(\mathfrak{C},\mathfrak{C}^k\big)-\mathfrak{J}\big(\mathfrak{C}\big) \ge 0$ for sufficiently large $\delta_{\mathbf{C}}^k$. By substracting eq.~\ref{eq80} from eq.~\ref{eq82}, we get:
\begin{align}
\mathfrak{G}\big(\mathfrak{C},\mathfrak{C}^k\big)-\mathfrak{J}\big(\mathfrak{C}\big)&=\frac{1}{2}\,\mathrm{tr}\,\big\{\big(\mathfrak{C}-\mathfrak{C}^k\big)^T\big(\mathbf{D}-\nabla_{\mathbf{C}}^2\mathbf{J}\big(\mathbf{C}^k\big)\big)\big(\mathfrak{C}-\mathfrak{C}^k\big)\big\} - \mathbf{\varepsilon}_{\mathbf{C}}^k \nonumber\\
&=\frac{1}{2}\sum_{n=1}^N\left[\big(\mathbf{c}_n-\mathbf{c}_n^k\big)^T\big(\mathbf{D}^n-\nabla_{\mathbf{C}}^2 J \big(\mathbf{C}^k\big)\big)\big(\mathbf{c}_n-\mathbf{c}_n^k\big)\right] - \mathbf{\varepsilon}_{\mathbf{C}}^k. \label{eqC}
\end{align}
Let $\mathbf{v}_n = \mathbf{c}_n - \mathbf{c}_n^k$, then:
\begin{align*}
\mathbf{v}_n^T\big(\mathbf{D}^n-\nabla_{\mathbf{C}}^2J\big(\mathbf{C}^k\big)\big)\mathbf{v}_n &= \mathbf{v}_n^T\big(\mathbf{D}^n + \alpha\mathbf{I} - \big(\mathbf{B}^{(k+1)T}\mathbf{B}^{(k+1)} + 3\alpha\mathbf{C}^k\mathbf{C}^{kT}\big)\big)\mathbf{v}_n \\
&= \mathbf{v}_n^T\big(\mathbf{\bar{D}}^n + \delta_{\mathbf{C}}^k\mathbf{\hat{D}}^n + \alpha\mathbf{I} - \big(\mathbf{B}^{(k+1)T}\mathbf{B}^{(k+1)} + 3\alpha\mathbf{C}^k\mathbf{C}^{kT}\big)\big)\mathbf{v}_n,
\end{align*}
where $\mathbf{\bar{D}}^n$ and $\delta_{\mathbf{C}}^k\mathbf{\hat{D}}^n$ are diagonal matrices that summed up to $\mathbf{D}^n$, with
\begin{align*}
\bar{d}_{rr}^n &\equiv \left\{
 \begin{array}{ll}
   \frac{\big( \mathbf{B}^{(k+1)T}\mathbf{B}^{(k+1)}\mathbf{\bar{C}}^k + \alpha\mathbf{\bar{C}}^k\mathbf{\bar{C}}^{kT}\mathbf{\bar{C}}^k \big)_{rn}}{\bar{c}_{rn}^k} & \mathrm{if}\;\; r\in \mathcal{I}_n \\
   \star & \mathrm{if}\;\; r\notin \mathcal{I}_n,
 \end{array} \right.
\text{and}\;
\hat{d}_{rr}^n &\equiv \left\{
 \begin{array}{ll}
   \frac{1}{\bar{c}_{rn}^k} & \mathrm{if}\;\; r\in \mathcal{I}_n \\
   \star & \mathrm{if}\;\; r\notin \mathcal{I}_n.
 \end{array} \right.
\end{align*}
Accordingly,
\begin{align}
\mathfrak{G}\big(\mathfrak{C},\mathfrak{C}^k\big)-\mathfrak{J}\big(\mathfrak{C}\big) = &\frac{1}{2} \sum_{n=1}^N \left\{\sum_{r=1}^R v_{rn}^2 \bar{d}_{rr}^n + \delta_{\mathbf{C}}^k \sum_{r=1}^R v_{rn}^2 \hat{d}_{rr}^n + \alpha \sum_{r=1}^R v_{rn}^2 \right\} \nonumber \\
&- \frac{1}{2} \sum_{n=1}^N \mathbf{v}_n^T \big( \mathbf{B}^{(k+1)T}\mathbf{B}^{(k+1)} + 3\alpha\mathbf{C}^k\mathbf{C}^{kT}\big)\mathbf{v}_n - \varepsilon_{\mathbf{C}}^k. \label{eqE}
\end{align}
As shown, with the boundedness of $\mathbf{B}^k$ and $\mathbf{C}^k$ and by sufficiently large $\delta_{\mathbf{C}}^k$, $\mathfrak{G}\big(\mathfrak{C},\mathfrak{C}\big) \le \mathfrak{G}\big(\mathfrak{C},\mathfrak{C}^k\big)$ can be guaranteed. Next we prove that if and only if $\mathbf{C}^k$ satisfies the KKT conditions, then the equality holds. 

If $\mathbf{C}^k$ satisfies the KKT conditions, then this is obvious by eq.~\ref{eqC} regardless of $\delta_{\mathbf{C}}^k$. And by eq.~\ref{eqE}, since $\delta_{\mathbf{C}}^k$ is a variable, the equality happens if and only if $\mathbf{C} = \mathbf{C}^k$ which by the update rule in eq.~\ref{eq70} and the boundedness of $\mathbf{B}^k$ and $\mathbf{C}^k$ happens if and only if $\mathbf{C}^k$ satisfies the KKT conditions. This completes the proof.
\end{proof}
Note that $\alpha$ should not be adjusted to ensure $\mathfrak{G}\big(\mathfrak{C},\mathfrak{C}\big) \le \mathfrak{G}\big(\mathfrak{C},\mathfrak{C}^k\big)$, since not only $\varepsilon_{\mathbf{C}}^k$ contains $\alpha$, but also $\alpha$ has a role in determining the orthogonality degree of $\mathbf{C}$ which should be determined from the start as a constant.

\begin{theorem} \label{theorem12}
$\mathfrak{G}\big(\mathfrak{C},\mathfrak{C}^k\big) \le \mathfrak{G}\big(\mathfrak{C}^k,\mathfrak{C}^k\big)$. Moreover if and only if $\mathbf{C}^k$ satisfies the KKT conditions in eq.~\ref{eq54}, then $\mathfrak{G}\big(\mathfrak{C},\mathfrak{C}^k\big) = \mathfrak{G}\big(\mathfrak{C}^k,\mathfrak{C}^k\big)$.
\end{theorem}
\begin{proof}
\begin{equation*}
\mathfrak{G}\big(\mathfrak{C}^k,\mathfrak{C}^k\big) - \mathfrak{G}\big(\mathfrak{C},\mathfrak{C}^k\big) = -\mathrm{tr}\;\big\{\big(\mathfrak{C}-\mathfrak{C}^k\big)^T\nabla_{\mathfrak{C}}\mathfrak{J}\big(\mathfrak{C}^{kT}\big)\big\} - \frac{1}{2}\mathrm{tr}\;\big\{\big(\mathfrak{C}-\mathfrak{C}^k\big)^T\mathbf{D}\big(\mathfrak{C}-\mathfrak{C}^k\big)\big\}.
\end{equation*}
By using eq.~\ref{eq81} and the fact that $\mathbf{D}$ is positive semi-definite:
\begin{equation*}
\mathfrak{G}\big(\mathfrak{C}^k,\mathfrak{C}^k\big) - \mathfrak{G}\big(\mathfrak{C},\mathfrak{C}^k\big) = \frac{1}{2}\mathrm{tr}\;\big\{\big(\mathfrak{C}-\mathfrak{C}^k\big)^T\mathbf{D}\big(\mathfrak{C}-\mathfrak{C}^k\big)\big\} \ge 0, 
\end{equation*}
By the update rule eq.~\ref{eq70}, if $\mathbf{C}^k$ satisfies the KKT conditions, then $\mathbf{C}=\mathbf{C}^k$, and therefore the equality holds. Now we need to prove that if the equality holds, then $\mathbf{C}^k$ satisfies the KKT conditions.

To prove this, let consider a contradiction situation where the equality holds but $\mathbf{C}^k$ does not satisfy the KKT conditions. In this case, there exists at least an index $(r,n)$ such that:
\begin{equation*}
c_{rn}\ne c_{rn}^k\;\;\mathrm{and}\;\; d_{rr}^n = \frac{\big(\mathbf{B}^{(k+1)T}\mathbf{B}^{(k+1)}\mathbf{\bar{C}}^k + \alpha \mathbf{\bar{C}}^k\mathbf{\bar{C}}^{kT}\mathbf{\bar{C}}^k\big)_{rn}+\delta_{\mathbf{C}}^k}{\bar{c}_{rn}^k} \ge \frac{\delta_{\mathbf{C}}^k}{\bar{c}_{rn}^k}.
\end{equation*}
Note that by the definition in eq.~\ref{eq66}, if $\bar{c}_{rn}^k$ is equal to zero, then $c_{rn}= c_{rn}^k$ which violates the condition for the contradiction, so $\bar{c}_{rn}^k$ cannot be equal to zero. Consequently,
\begin{equation*}
\mathfrak{G}\big(\mathfrak{C}^k,\mathfrak{C}^k\big)-\mathfrak{G}\big(\mathfrak{C},\mathfrak{C}^k\big) \ge \frac{\big(c_{rn}-c_{rn}^k\big)^2\delta_{\mathbf{C}}^k}{\bar{c}_{rn}^k} > 0,
\end{equation*}
which violates the equality. Thus, it is proven that if the equality holds, then $\mathbf{C}^k$ satisfies the KKT conditions.
\end{proof}

\begin{theorem}\label{theorem13}
Given sufficiently large $\delta_{\mathbf{C}}^k$ and the boundedness of $\mathbf{B}^k$ and $\mathbf{C}^k$, $J\big(\mathbf{C}^{k+1}\big)$ $\le$ $J\big(\mathbf{C}^k\big)\;\,\forall k\ge 0$ under update rule eq.~\ref{eq70} with the equality happens if and only if $\mathbf{C}^k$ satisfies the KKT conditions in eq.~\ref{eq54}. 
\end{theorem}
\begin{proof}
This theorem is the corollary of theorem \ref{theorem10}, \ref{theorem11}, and \ref{theorem12}.
\end{proof}

\paragraph{C.~Convergence guarantee of algorithm \ref{algorithm5}}

To show the convergence of algorithm \ref{algorithm5}, the following statements must be proven \cite{CJLin2}: 
\begin{enumerate}
\item the nonincreasing property of sequence $J\big(\mathbf{B}^k,\mathbf{C}^k\big)$, i.e., $J\big(\mathbf{B}^{(k+1)}$,$\mathbf{C}^{(k+1)}\big)$ $\le$ $J\big(\mathbf{B}^{(k+1)}$,$\mathbf{C}^k\big)$ $\le$ $J\big(\mathbf{B}^k$,$\mathbf{C}^k\big)$,
\item any limit point of sequence $\big\{\mathbf{B}^k,\mathbf{C}^k\big\}$ generated by algorithm \ref{algorithm5} is a stationary point, and
\item sequence $\big\{\mathbf{B}^k,\mathbf{C}^k\big\}$ has at least one limit point.
\end{enumerate}
The first will be proven in theorem \ref{theorem14}, the second in theorem \ref{theorem15}, and the third in theorem \ref{theorem16}. Note that satisfying the KKT conditions is sufficient for stationarity. 

\begin{theorem} \label{theorem14}
Given sufficiently large $\delta_{\mathbf{C}}^k$ and the boundedness of $\mathbf{B}^k$ and $\mathbf{C}^k$, $J\big(\mathbf{B}^{(k+1)}$, $\mathbf{C}^{(k+1)}\big)$ $\le$ $J\big(\mathbf{B}^{(k+1)}$, $\mathbf{C}^k\big)$ $\le$ $J\big(\mathbf{B}^k$, $\mathbf{C}^k\big)$ under update rules in algorithm \ref{algorithm5} with the equalities happen if and only if $\big(\mathbf{B}^k$, $\mathbf{C}^k\big)$ is a stationary point.
\end{theorem}
\begin{proof}
$J\big(\mathbf{B}^{(k+1)},\mathbf{C}^k\big)$ $\le$ $J\big(\mathbf{B}^k,\mathbf{C}^k\big)$ is due to theorem \ref{theorem9} with the equality happens if and only if $\mathbf{B}^k$ satisfies the KKT conditions. And for sufficiently large $\delta_{\mathbf{C}}^k$ and the boundedness of $\mathbf{B}^k$ and $\mathbf{C}^k$, $J\big(\mathbf{B}^{(k+1)},\mathbf{C}^{(k+1)}\big)$ $\le$ $J\big(\mathbf{B}^{(k+1)},\mathbf{C}^k\big)$ is due to theorem \ref{theorem13} with the equality happens if and only if $\mathbf{C}^k$ satisfies the KKT conditions. And by combining theorem \ref{theorem9} and \ref{theorem13}, algorithm \ref{algorithm5} will stop updating sequence $J\big(\mathbf{B}^{k},\mathbf{C}^k\big)$ if and only if both $\mathbf{B}^k$ and $\mathbf{C}^k$ satisfy the KKT conditions., i.e., $\big(\mathbf{B}^k$, $\mathbf{C}^k\big)$ is a stationary point. 
\end{proof}
\begin{theorem}\label{theorem15}
Given sufficiently large $\delta_{\mathbf{C}}^k$ and with the boundedness of $\mathbf{B}^k$ and $\mathbf{C}^k$, it can be shown that any limit point of sequence $\big\{\mathbf{B}^k,\mathbf{C}^k\big\}$ generated by algorithm \ref{algorithm5} is a stationary point.
\end{theorem}
\begin{proof}
By theorem \ref{theorem14}, algorithm \ref{algorithm5} produces strictly decreasing sequence $J\big(\mathbf{B}^k$, $\mathbf{C}^k\big)$ until reaching a point that satisfies the KKT conditions. Because $J\big(\mathbf{B}^k$, $\mathbf{C}^k\big)$ $\ge$ $0$, this sequence is bounded and thus converges. And by combining results of theorem \ref{theorem9} and \ref{theorem13}, algorithm \ref{algorithm5} stop updating $J\big(\mathbf{B}^k$, $\mathbf{C}^k\big)$ if and only if $\big(\mathbf{B}^k$, $\mathbf{C}^k\big)$ satisfies the KKT conditions. And by update rules in algorithm \ref{algorithm5}, after a point satisfies the KKT conditions, the algorithm will stop updating $\big(\mathbf{B}^k$, $\mathbf{C}^k\big)$, i.e., $\mathbf{B}^{(k+1)}$ $=$ $\mathbf{B}^k$ and $\mathbf{C}^{(k+1)}$ $=$ $\mathbf{C}^k\;\,\forall k\ge *$ ($*$ is the first iteration where the stationarity is reached). This completes the proof.
\end{proof}
\begin{theorem}\label{theorem16}
Sequence $\big\{\mathbf{B}^k,\mathbf{C}^k\big\}$ has at least one limit point.
\end{theorem}
\begin{proof}
As stated by Lin \cite{CJLin2}, it suffices to prove that sequence $\big\{\mathbf{B}^k,\mathbf{C}^k\big\}$ is in a closed and bounded set. The boundedness of $\big\{\mathbf{C}^k\big\}$ is clear by the objective in eq.~\ref{eq52}; if there exists $l$ such that $\lim c_{rn}^l\to \infty$, then $\lim J(\mathbf{B}^l,\mathbf{C}^l) \to \infty > J(\mathbf{B}^0,\mathbf{C}^0)$ which violates theorem \ref{theorem14}. And if $\big\{\mathbf{B}^k\big\}$ is not bounded, then there exists $l$ such that $\lim b_{mr}^l\to \infty$, $b_{mr}^l < b_{mr}^{(l+1)}$. Because due to theorem \ref{theorem14}, $J(\mathbf{B}^k,\mathbf{C}^k)$ is bounded, then $c_{rn}^l$ $\forall n$ must be equal to zero. And if $c_{rn}^l=0\;\forall n$, then $\nabla_{\mathbf{B}}J\big(\mathbf{B}^l,\mathbf{C}^l\big)_{mr}=0\;\,\forall m$, so that $b_{mr}^{(l+1)}=b_{mr}^l\;\,\forall m$ which conflicting the condition for unboundedness of $\mathbf{B}^l$. Thus, $\mathbf{B}^l$ is also bounded. With nonnegativity guarantee from theorem \ref{theorem5}, it is proven that $\big\{\mathbf{B}^k,\mathbf{C}^k\big\}$ is in a closed and bounded set.
\end{proof}

Algorithm \ref{algorithm6} shows some modifications to algorithm \ref{algorithm5} in order to guarantee the convergence as suggested by theorem \ref{theorem14}, \ref{theorem15}, and \ref{theorem16}, with step is a constant that determines how fast $\delta_{\mathbf{C}}^k$ grows in order to satisfy the nonincreasing property.

\begin{algorithm}
\caption{Converged algorithm for UNMF.}
\label{algorithm6}
\begin{algorithmic}
\STATE Initialization, $\mathbf{B}^0\ge\mathbf{0}$, $\mathbf{C}^0\ge\mathbf{0}$.
\FOR {$k=0,\ldots,K$}
\STATE \begin{equation*} b_{mr}^{(k+1)} \longleftarrow b_{mr}^{k} - \frac{\bar{b}_{mr}^{k}\times\nabla_{\mathbf{B}}J(\mathbf{B}^k,\mathbf{C}^k)_{mr}}{\big(\mathbf{\bar{B}}^{k}\mathbf{C}^{k}\mathbf{C}^{kT}\big)_{mr}+\delta}\;\;\forall m,r 
\end{equation*}
\STATE $\delta_{\mathbf{C}}^k \longleftarrow \delta$
\REPEAT 
\STATE \begin{align*} c_{rn}^{(k+1)} \longleftarrow & \;c_{rn}^{k} - \frac{\bar{c}_{rn}^{k}\times\nabla_{\mathbf{C}}J(\mathbf{B}^{k+1},\mathbf{C}^k)_{rn}}{\big(\mathbf{B}^{(k+1)T}\mathbf{B}^{(k+1)}\mathbf{\bar{C}}^{k}+\alpha\mathbf{\bar{C}}^{k}\mathbf{\bar{C}}^{kT}\mathbf{\bar{C}}^{k}\big)_{rn}+\delta_{\mathbf{C}}^k} \;\;\forall r,n \\
\delta_{\mathbf{C}}^k \longleftarrow & \;\delta_{\mathbf{C}}^k\times \mathrm{step}  \end{align*}
\UNTIL {$J\big(\mathbf{B}^{(k+1)}, \mathbf{C}^{(k+1)}\big) \le J\big(\mathbf{B}^{(k+1)},\mathbf{C}^k\big)$}
\ENDFOR
\end{algorithmic}
\end{algorithm}

\subsection{Converged bi-orthogonal NMF} \label{mybiortho}

Converged algorithm for BNMF will be derived equivalently as in UNMF case. However, we will not cut the steps in deriving the algorithm. The readers can refer to algorithm \ref{algorithm9} for the final form.

First, let's define BNMF objective with following:
\begin{align}
&\min_{\mathbf{B},\mathbf{C}}J(\mathbf{B},\mathbf{C},\mathbf{S})=\frac{1}{2}\|\mathbf{A}-\mathbf{B}\mathbf{S}\mathbf{C}\|_{F}^{2} + \frac{\alpha}{2}\|\mathbf{CC}^T-\mathbf{I}\|_{F}^{2} + \frac{\beta}{2}\|\mathbf{B}^T\mathbf{B}-\mathbf{I}\|_{F}^{2} \label{eq90}\\
&\mathrm{s.t.}\;\, \mathbf{B}\ge\mathbf{0},\mathbf{C}\ge\mathbf{0},\mathbf{S}\ge\mathbf{0}, \nonumber
\end{align}
with $\alpha$ and $\beta$ are constants to adjust the degree of orthogonality of $\mathbf{C}$ and $\mathbf{B}$ respectively.
The KKT function of the objective can be written as:
\begin{align*}
L(\mathbf{B},\mathbf{C})=\;&J(\mathbf{B},\mathbf{C})-\mathrm{tr}\;\big(\mathbf{\Gamma}_{\mathbf{B}}\mathbf{B}^T\big)-\mathrm{tr}\;\big(\mathbf{\Gamma}_{\mathbf{S}}\mathbf{S}^T\big)-\mathrm{tr}\;\big(\mathbf{\Gamma}_{\mathbf{C}}\mathbf{C}\big). 
\end{align*}
And the KKT conditions are:
\begin{equation}
\begin{array}{rrr}
\mathbf{B}^*\ge\mathbf{0}, & \mathbf{S}^*\ge\mathbf{0}, & \mathbf{C}^*\ge\mathbf{0}, \\
\nabla_{\mathbf{B}}J(\mathbf{B}^*)=\mathbf{\Gamma}_{\mathbf{B}}\ge\mathbf{0}, & \nabla_{\mathbf{S}}J(\mathbf{S}^*)=\mathbf{\Gamma}_{\mathbf{S}}\ge\mathbf{0}, & \nabla_{\mathbf{C}}J(\mathbf{C}^*)=\mathbf{\Gamma}_{\mathbf{C}}^T\ge\mathbf{0},\\
\nabla_{\mathbf{B}}J(\mathbf{B}^*)\odot\mathbf{B}^*=\mathbf{0}, & \nabla_{\mathbf{S}}J(\mathbf{S}^*)\odot\mathbf{S}^*=\mathbf{0}, & \nabla_{\mathbf{C}}J(\mathbf{C}^*)\odot\mathbf{C}^*=\mathbf{0}, \label{eq91}
\end{array}
\end{equation}
where
\begin{align*}
\nabla_{\mathbf{B}}J(\mathbf{B})&=\mathbf{BSCC}^T\mathbf{S}^T-\mathbf{AC}^T\mathbf{S}^T+\beta\mathbf{BB}^T\mathbf{B}-\beta\mathbf{B}, \\
\nabla_{\mathbf{C}}J(\mathbf{C})&=\mathbf{S}^T\mathbf{B}^T\mathbf{BSC}-\mathbf{S}^T\mathbf{B}^T\mathbf{A}+\alpha\mathbf{CC}^T\mathbf{C}-\alpha\mathbf{C}, \\
\nabla_{\mathbf{S}}J(\mathbf{S})&=\mathbf{B}^T\mathbf{BSCC}^T-\mathbf{B}^T\mathbf{AC}^T.
\end{align*}
Then, the MU algorithm for objective in eq.~\ref{eq90} can be written as:
\begin{align*} 
b_{mp} &\longleftarrow b_{mp}\frac{\big(\mathbf{AC}^T\mathbf{S}^T+\beta\mathbf{B}\big)_{mp}}{\big(\mathbf{BSCC}^T\mathbf{S}^T+\beta\mathbf{BB}^T\mathbf{B}\big)_{mp}}, \\
c_{qn} &\longleftarrow c_{qn}\frac{\big(\mathbf{S}^T\mathbf{B}^T\mathbf{A}+\alpha\mathbf{C}\big)_{qn}}{\big(\mathbf{S}^T\mathbf{B}^T\mathbf{BSC}+\alpha\mathbf{CC}^T\mathbf{C}\big)_{qn}}, \\
s_{pq} &\longleftarrow s_{pq}\frac{\big(\mathbf{B}^T\mathbf{AC}^T\big)_{pq}}{\big(\mathbf{B}^T\mathbf{BSCC}^T\big)_{pq}}.
\end{align*}

The complete MU algorithm is given in algorithm \ref{algorithm7}, and the AU version is in algorithm \ref{algorithm8}.

\begin{algorithm}
\caption{The MU algorithm for BNMF problem in eq.~\ref{eq90}.}
\label{algorithm7}
\begin{algorithmic}
\STATE Initialization, $\mathbf{B}^0>\mathbf{0}$, $\mathbf{C}^0>\mathbf{0}$, and $\mathbf{S}^0>\mathbf{0}$.
\FOR {$k=0,\ldots,K$}
\STATE \begin{align*} b_{mp}^{(k+1)} &\longleftarrow b_{mp}^{k}\frac{\big(\mathbf{AC}^{kT}\mathbf{S}^{kT} + \beta\mathbf{B}^k\big)_{mp}}{\big(\mathbf{B}^{k}\mathbf{S}^{k}\mathbf{C}^k\mathbf{C}^{kT}\mathbf{S}^{kT}+\beta\mathbf{B}^k\mathbf{B}^{kT}\mathbf{B}^k\big)_{mp}+\delta}\;\;\forall m,p 
\\
c_{qn}^{(k+1)} &\longleftarrow c_{qn}^{k}\frac{\big(\mathbf{S}^{kT}\mathbf{B}^{(k+1)T}\mathbf{A}+\alpha\mathbf{C}^k\big)_{qn}}{\big(\mathbf{S}^{kT}\mathbf{B}^{(k+1)T}\mathbf{B}^{(k+1)}\mathbf{S}^k\mathbf{C}^k+\alpha\mathbf{C}^k\mathbf{C}^{kT}\mathbf{C}^k\big)_{qn}+\delta}\;\;\forall q,n 
\\
s_{pq}^{(k+1)} &\longleftarrow s_{pq}^{k}\frac{\big(\mathbf{B}^{(k+1)T}\mathbf{A}\mathbf{C}^{(k+1)T}\big)_{pq}}{\big(\mathbf{B}^{(k+1)T}\mathbf{B}^{(k+1)}\mathbf{S}^{k}\mathbf{C}^{(k+1)}\mathbf{C}^{(k+1)T}\big)_{pq}+\delta}\;\;\forall p,q 
\end{align*}
\ENDFOR
\end{algorithmic}
\end{algorithm}

\begin{algorithm}
\caption{The AU algorithm for BNMF problem in eq.~\ref{eq90}.}
\label{algorithm8}
\begin{algorithmic}
\STATE Initialization, $\mathbf{B}^0\ge\mathbf{0}$, $\mathbf{C}^0\ge\mathbf{0}$, and $\mathbf{S}^0\ge\mathbf{0}$.
\FOR {$k=0,\ldots,K$}
\STATE \begin{align} b_{mp}^{(k+1)} \longleftarrow & \;b_{mp}^{k} - \frac{\bar{b}_{mp}^{k}\times\nabla_{\mathbf{B}}J(\mathbf{B}^k,\mathbf{S}^k,\mathbf{C}^k)_{mp}}{\big(\mathbf{\bar{B}}^{k}\mathbf{S}^{k}\mathbf{C}^k\mathbf{C}^{kT}\mathbf{S}^{kT}+\beta\mathbf{\bar{B}}^k\mathbf{\bar{B}}^{kT}\mathbf{\bar{B}}^k\big)_{mp}+\delta_{\mathbf{B}}^k}\;\;\forall m,p \label{eq98}\\
c_{qn}^{(k+1)} \longleftarrow & \;c_{qn}^{k} - \frac{\bar{c}_{qn}^{k}\times\nabla_{\mathbf{C}}J(\mathbf{B}^{k+1},\mathbf{S}^k,\mathbf{C}^k)_{qn}}{\big(\mathbf{S}^{kT}\mathbf{B}^{(k+1)T}\mathbf{B}^{(k+1)}\mathbf{S}^k\mathbf{\bar{C}}^k+\alpha\mathbf{\bar{C}}^k\mathbf{\bar{C}}^{kT}\mathbf{\bar{C}}^k\big)_{qn}+\delta_{\mathbf{C}}^k} \;\;\forall q,n \label{eq99} \\
s_{pq}^{(k+1)} \longleftarrow & \;s_{pq}^{k} - \frac{\bar{s}_{pq}^{k}\times\nabla_{\mathbf{S}}J(\mathbf{B}^{k+1},\mathbf{S}^k,\mathbf{C}^{(k+1)})_{pq}}{\big(\mathbf{B}^{(k+1)T}\mathbf{B}^{(k+1)}\mathbf{\bar{S}}^{k}\mathbf{C}^{(k+1)}\mathbf{C}^{(k+1)T}\big)_{pq}+\delta_{\mathbf{S}}^k} \;\;\forall p,q \label{eq100}
\end{align}
\ENDFOR
\end{algorithmic}
\end{algorithm}

There are $\bar{b}_{mp}^k$, $\bar{c}_{qn}^k$, and $\bar{s}_{pq}^k$ in algorithm \ref{algorithm8} which are the modifications to $b_{mp}^k$, $c_{qn}^k$, and $s_{pq}^k$ to avoid the zero locking. The following gives their definitions.
\begin{align}
\bar{b}_{mp}^k &\equiv \left\{
  \begin{array}{rl}
    b_{mp}^k\hspace{13 mm} & \text{if  } \nabla_{\mathbf{B}}J\big(\mathbf{B}^k,\mathbf{S}^k,\mathbf{C}^k\big)_{mp} \ge 0 \\
    \max(b_{mp}^k, \sigma) & \text{if  } \nabla_{\mathbf{B}}J\big(\mathbf{B}^k,\mathbf{S}^k,\mathbf{C}^k\big)_{mp} < 0
  \end{array}, \right. \label{eq101} \\
\bar{c}_{qn}^k &\equiv \left\{
  \begin{array}{rl}
    c_{qn}^k\hspace{13 mm} & \text{if  } \nabla_{\mathbf{C}}J\big(\mathbf{B}^{(k+1)},\mathbf{S}^k,\mathbf{C}^k\big)_{qn} \ge 0 \\
    \max(c_{qn}^k, \sigma) & \text{if  } \nabla_{\mathbf{C}}J\big(\mathbf{B}^{(k+1)},\mathbf{S}^k,\mathbf{C}^k\big)_{qn} < 0
  \end{array}, \right. \label{eq102} \\
\bar{s}_{pq}^k &\equiv \left\{
  \begin{array}{rl}
    s_{pq}^k\hspace{13 mm} & \text{if  } \nabla_{\mathbf{S}}J\big(\mathbf{B}^{(k+1)},\mathbf{S}^k,\mathbf{C}^{(k+1)}\big)_{pq} \ge 0 \\
    \max(s_{pq}^k, \sigma) & \text{if  } \nabla_{\mathbf{S}}J\big(\mathbf{B}^{(k+1)},\mathbf{S}^k,\mathbf{C}^{(k+1)}\big)_{pq} < 0
  \end{array}, \right. \label{eq103}
\end{align}
with $\sigma$ is a small positive number, $\mathbf{\bar{B}}$, $\mathbf{\bar{C}}$, and $\mathbf{\bar{S}}$ are matrices that contain $\bar{b}_{mp}$, $\bar{c}_{qn}$, and $\bar{s}_{pq}$ respectively. And:
\begin{align*}
\nabla_{\mathbf{B}}J(\mathbf{B}^k,\mathbf{S}^k,\mathbf{C}^k)=&\;\mathbf{B}^{k}\mathbf{S}^{k}\mathbf{C}^k\mathbf{C}^{kT}\mathbf{S}^{kT} - \mathbf{AC}^{kT}\mathbf{S}^{kT} + \beta\mathbf{B}^k\mathbf{B}^{kT}\mathbf{B}^k - \beta\mathbf{B}^k,
\\ 
\nabla_{\mathbf{C}}J(\mathbf{B}^{k+1},\mathbf{S}^k,\mathbf{C}^k)=&\;\mathbf{S}^{kT}\mathbf{B}^{(k+1)T}\mathbf{B}^{(k+1)}\mathbf{S}^k\mathbf{C}^k - \mathbf{S}^{kT}\mathbf{B}^{(k+1)T}\mathbf{A} + \nonumber\\ &\;\alpha\mathbf{C}^k\mathbf{C}^{kT}\mathbf{C}^k - \alpha\mathbf{C}^k, 
\\
\nabla_{\mathbf{S}}J(\mathbf{B}^{k+1},\mathbf{S}^k,\mathbf{C}^{k+1})=&\;\mathbf{B}^{(k+1)T}\mathbf{B}^{(k+1)}\mathbf{S}^{k}\mathbf{C}^{(k+1)}\mathbf{C}^{(k+1)T} - \mathbf{B}^{(k+1)T}\mathbf{A}\mathbf{C}^{(k+1)T}. 
\end{align*}

As in subsection \ref{myuniortho}, due to the zero locking, there is no convergence guarantee for algorithm \ref{algorithm7}. And also as in subsection \ref{myuniortho}, $\delta_{\mathbf{B}}^k$, $\delta_{\mathbf{C}}^k$, and $\delta_{\mathbf{S}}^k$ in algorithm \ref{algorithm8} are variables that play crucial roles in guaranteeing the convergence of the algorithm. Note that, algorithm \ref{algorithm7} must be initialized with positive matrices to avoid the zero locking from the start, and nonnegative matrices can be used to initialize algorithm \ref{algorithm8}. The following theorem explains this formally.

\begin{theorem}\label{theorem17}
If $\mathbf{B}^0>0$, $\mathbf{C}^0>0$, and $\mathbf{S}^0>0$, then $\mathbf{B}^k>0$, $\mathbf{C}^k>0$, and $\mathbf{S}^k>0$ $\forall k\ge 0$. And if $\mathbf{B}^0\ge 0$, $\mathbf{C}^0 \ge0$, and $\mathbf{S}^0 \ge0$, then $\mathbf{B}^k\ge 0$, $\mathbf{C}^k\ge 0$, and $\mathbf{S}^k\ge 0$ $\forall k\ge 0$
\end{theorem}
\begin{proof}
This statement is clear for $k=0$, so we need only to prove for $k>0$.\\
\\
\emph{Case 1}: $\nabla_{\mathbf{B}}J_{mp}\ge 0 \Rightarrow \bar{b}_{mp} = b_{mp}$.
\begin{align*}
b_{mp}^{(k+1)} = & \frac{\big(\mathbf{B}^{k}\mathbf{S}^{k}\mathbf{C}^k\mathbf{C}^{kT}\mathbf{S}^{kT} + \beta\mathbf{B}^k\mathbf{B}^{kT}\mathbf{B}^k\big)_{mp}b_{mp}^k+\delta_{\mathbf{B}}^k b_{mp}^k}{\big(\mathbf{B}^{k}\mathbf{S}^{k}\mathbf{C}^k\mathbf{C}^{kT}\mathbf{S}^{kT} + \beta\mathbf{B}^k\mathbf{B}^{kT}\mathbf{B}^k\big)_{mp}+\delta_{\mathbf{B}}^k}- \\
&\frac{\big( \mathbf{B}^{k}\mathbf{S}^{k}\mathbf{C}^k\mathbf{C}^{kT}\mathbf{S}^{kT} + \beta\mathbf{B}^k\mathbf{B}^{kT}\mathbf{B}^k - \mathbf{AC}^{kT}\mathbf{S}^{kT} - \beta\mathbf{B}^k \big)_{mp}b_{mp}^k}{\big( \mathbf{B}^{k}\mathbf{S}^{k}\mathbf{C}^k\mathbf{C}^{kT}\mathbf{S}^{kT} + \beta\mathbf{B}^k\mathbf{B}^{kT}\mathbf{B}^k \big)_{mp}+\delta_{\mathbf{B}}^k} \\
= & \frac{ \big[ \big( \mathbf{AC}^{kT}\mathbf{S}^{kT} + \beta\mathbf{B}^k \big)_{mp} + \delta_{\mathbf{B}}^k \big] b_{mp}^k}{\big( \mathbf{B}^{k}\mathbf{S}^{k}\mathbf{C}^k\mathbf{C}^{kT}\mathbf{S}^{kT} + \beta\mathbf{B}^k\mathbf{B}^{kT}\mathbf{B}^k \big)_{mp} + \delta_{\mathbf{B}}^k}.
\end{align*}
Thus, if $b_{mp}^k>0$ then $b_{mp}^{(k+1)}>0\;\forall m,p$, and if $b_{mp}^k\ge 0$ then $b_{mp}^{(k+1)}\ge 0\;\forall m,p,\;\;\forall k>0$.\\
\\
\emph{Case 2}: $\nabla_{\mathbf{B}}J_{mp}<0 \Rightarrow \bar{b}_{mp} \ne b_{mp}$.
\begin{align*}
b_{mp}^{(k+1)} = b_{mp}^k-\frac{\max\big(b_{mp}^k,\sigma\big)\times\nabla_{\mathbf{B}}J\big(\mathbf{B}^k,\mathbf{S}^k,\mathbf{C}^k\big)_{mp}}{\big( \mathbf{\bar{B}}^{k}\mathbf{S}^{k}\mathbf{C}^k\mathbf{C}^{kT}\mathbf{S}^{kT} + \beta\mathbf{\bar{B}}^k\mathbf{\bar{B}}^{kT}\mathbf{\bar{B}}^k \big)_{mp}+\delta_{\mathbf{B}}^k}.
\end{align*}
Note that $\max\big(b_{mp}^k,\sigma\big)>0$ and $\nabla_{\mathbf{B}}J\big(\mathbf{B}^k,\mathbf{S}^k,\mathbf{C}^k\big)_{mp}<0$. Thus if $b_{mp}^k>0$ then $b_{mp}^{(k+1)}>0\;\forall m,p$, and if $b_{mp}^k\ge 0$ then $b_{mp}^{(k+1)}>0\;\forall m,p,\;\;\forall k>0$.\\
\\
\emph{Case 3}: $\nabla_{\mathbf{C}}J_{qn}\ge 0 \Rightarrow \bar{c}_{qn} = c_{qn}$.
\begin{align*}
c_{qn}^{(k+1)} = &\frac{\big( \mathbf{S}^{kT}\mathbf{B}^{(k+1)T}\mathbf{B}^{(k+1)}\mathbf{S}^k\mathbf{C}^k + \alpha\mathbf{C}^k\mathbf{C}^{kT}\mathbf{C}^k \big)_{qn} c_{qn}^k + \delta_{\mathbf{C}}^k c_{qn}^k} {\big( \mathbf{S}^{kT}\mathbf{B}^{(k+1)T}\mathbf{B}^{(k+1)}\mathbf{S}^k\mathbf{C}^k + \alpha\mathbf{C}^k\mathbf{C}^{kT}\mathbf{C}^k \big)_{qn} + \delta_{\mathbf{C}}^k} - \\
&\frac{\big( \mathbf{S}^{kT}\mathbf{B}^{(k+1)T}\mathbf{B}^{(k+1)}\mathbf{S}^k\mathbf{C}^k + \alpha\mathbf{C}^k\mathbf{C}^{kT}\mathbf{C}^k - \mathbf{S}^{kT}\mathbf{B}^{(k+1)T}\mathbf{A} - \alpha\mathbf{C}^k \big)_{qn} c_{qn}^k} {\big( \mathbf{S}^{kT}\mathbf{B}^{(k+1)T}\mathbf{B}^{(k+1)}\mathbf{S}^k\mathbf{C}^k + \alpha\mathbf{C}^k\mathbf{C}^{kT}\mathbf{C}^k \big)_{qn} + \delta_{\mathbf{C}}^k} \\
=&\frac{\big[\big( \mathbf{S}^{kT}\mathbf{B}^{(k+1)T}\mathbf{A} + \alpha\mathbf{C}^k \big)_{qn} + \delta_{\mathbf{C}}^k\big] c_{qn}^k} {\big( \mathbf{S}^{kT}\mathbf{B}^{(k+1)T}\mathbf{B}^{(k+1)}\mathbf{S}^k\mathbf{C}^k + \alpha\mathbf{C}^k\mathbf{C}^{kT}\mathbf{C}^k \big)_{qn} + \delta_{\mathbf{C}}^k},
\end{align*}
Thus if $c_{qn}^k>0$ then $c_{qn}^{(k+1)}>0\;\forall q,n$, and if $c_{qn}^k\ge 0$ then $c_{qn}^{(k+1)}\ge 0\;\forall q,n,\;\;\forall k>0$.\\
\\
\emph{Case 4}: $\nabla_{\mathbf{C}}J_{qn}<0 \Rightarrow \bar{c}_{qn} \ne c_{qn}$.
\begin{align*}
c_{qn}^{(k+1)} = c_{qn}^k-\frac{\max\big(c_{qn}^k,\sigma\big)\times\nabla_{\mathbf{C}}J\big(\mathbf{B}^{(k+1)},\mathbf{S}^k,\mathbf{C}^k\big)_{qn}} {\big( \mathbf{S}^{kT}\mathbf{B}^{(k+1)T}\mathbf{B}^{(k+1)}\mathbf{S}^k\mathbf{\bar{C}}^k + \alpha\mathbf{\bar{C}}^k\mathbf{\bar{C}}^{kT}\mathbf{\bar{C}}^k \big)_{qn} + \delta_{\mathbf{C}}^k}.
\end{align*}
Note that $\max\big(c_{qn}^k,\sigma\big)>0$ and $\nabla_{\mathbf{C}}J\big(\mathbf{B}^{(k+1)},\mathbf{S}^k,\mathbf{C}^k\big)_{qn}<0$. Thus if $c_{qn}^k>0$ then $c_{qn}^{(k+1)}>0\;\forall q,n$, and if $c_{rn}^k\ge 0$ then $c_{rn}^{(k+1)}>0\;\forall q,n,\;\;\forall k>0$.\\
\\
\emph{Case 5}: $\nabla_{\mathbf{S}}J_{pq}\ge 0 \Rightarrow \bar{s}_{pq} = s_{pq}$.
\begin{align*}
s_{pq}^{(k+1)} = &\frac{\big( \mathbf{B}^{(k+1)T}\mathbf{B}^{(k+1)}\mathbf{S}^{k}\mathbf{C}^{(k+1)}\mathbf{C}^{(k+1)T} \big)_{pq} s_{pq}^k + \delta_{\mathbf{S}}^k s_{pq}^k} {\big( \mathbf{B}^{(k+1)T}\mathbf{B}^{(k+1)}\mathbf{S}^{k}\mathbf{C}^{(k+1)}\mathbf{C}^{(k+1)T} \big)_{pq} + \delta_{\mathbf{S}}^k} - \\
&\frac{\big( \mathbf{B}^{(k+1)T}\mathbf{B}^{(k+1)}\mathbf{S}^{k}\mathbf{C}^{(k+1)}\mathbf{C}^{(k+1)T} - \mathbf{B}^{(k+1)T}\mathbf{A}\mathbf{C}^{(k+1)T} \big)_{pq} s_{pq}^k} {\big( \mathbf{B}^{(k+1)T}\mathbf{B}^{(k+1)}\mathbf{S}^{k}\mathbf{C}^{(k+1)}\mathbf{C}^{(k+1)T}  \big)_{pq} + \delta_{\mathbf{S}}^k} \\
=&\frac{\big[\big( \mathbf{B}^{(k+1)T}\mathbf{A}\mathbf{C}^{(k+1)T} \big)_{qn} + \delta_{\mathbf{S}}^k\big] s_{pq}^k} {\big( \mathbf{B}^{(k+1)T}\mathbf{B}^{(k+1)}\mathbf{S}^{k}\mathbf{C}^{(k+1)}\mathbf{C}^{(k+1)T} \big)_{pq} + \delta_{\mathbf{S}}^k},
\end{align*}
Thus if $s_{pq}^k>0$ then $s_{pq}^{(k+1)}>0\;\forall p,q$, and if $s_{pq}^k\ge 0$ then $s_{pq}^{(k+1)}\ge 0\;\forall p,q,\;\;\forall k>0$.\\
\\
\emph{Case 6}: $\nabla_{\mathbf{S}}J_{pq}<0 \Rightarrow \bar{s}_{pq} \ne c_{pq}$.
\begin{align*}
s_{pq}^{(k+1)} = s_{pq}^k-\frac{\max\big(s_{pq}^k,\sigma\big)\times\nabla_{\mathbf{S}}J\big(\mathbf{B}^{(k+1)},\mathbf{S}^k,\mathbf{C}^{(k+1)}\big)_{pq}} {\big( \mathbf{B}^{(k+1)T}\mathbf{B}^{(k+1)}\mathbf{\bar{S}}^{k}\mathbf{C}^{(k+1)}\mathbf{C}^{(k+1)T} \big)_{pq} + \delta_{\mathbf{S}}^k}.
\end{align*}
Note that $\max\big(s_{pq}^k,\sigma\big)>0$ and $\nabla_{\mathbf{S}}J\big(\mathbf{B}^{(k+1)},\mathbf{S}^k,\mathbf{C}^{(k+1)}\big)_{pq}<0$. Thus if $s_{pq}^k>0$ then $s_{pq}^{(k+1)}>0\;\forall p,q$, and if $s_{pq}^k\ge 0$ then $s_{pq}^{(k+1)}>0\;\forall p,q,\;\;\forall k>0$.

By combining the results for $k=0$ and for $k>0$ in case 1-6, the proof is completed.
\end{proof}

\subsubsection{Convergence analysis}

We will now analyze convergence property of algorithm \ref{algorithm8}. As stated previously, the nonincreasing property of sequence $J\big(\mathbf{B}^k,\mathbf{S}^k,\mathbf{C}^k\big)$ need to be proven first as it is the necessary condition for the convergence. And because algorithm \ref{algorithm8} uses alternating strategy, the nonincreasing property can be analyzed separately.

\paragraph{A.~The nonincreasing property of $J\big(\mathbf{B}^k\big)$}

By using the auxiliary function approach, the nonincreasing property of $J\big(\mathbf{B}^k\big)$ can be proven through:
\begin{equation*}
J\big(\mathbf{B}^{(k+1)}\big) = \; G\big(\mathbf{B}^{(k+1)},\mathbf{B}^{(k+1)}\big) \le G\big(\mathbf{B}^{(k+1)},\mathbf{B}^k\big) \le G\big(\mathbf{B}^k,\mathbf{B}^k\big) = J\big(\mathbf{B}^k\big). 
\end{equation*}

To define $G$, let's rearrange $\mathbf{B}$ into:
\begin{equation*}
\mathfrak{B}^T \equiv
\begin{bmatrix}
\mathfrak{b}_1^T & & & \\
& \mathfrak{b}_2^T & & \\
& & \ddots & \\
& & & \mathfrak{b}_M^T
\end{bmatrix}\in\mathbb{R}_+^{MP\times M},
\end{equation*}
where $\mathfrak{b}_m$ is the $m$-th row of $\mathbf{B}$. And also let's define:
\begin{equation*}
\nabla_{\mathfrak{B}^T}\mathfrak{J}\big(\mathfrak{B}^{kT}\big) \equiv
\begin{bmatrix}
\nabla_{\mathbf{B}}\mathfrak{J}\big(\mathbf{B}^k\big)_1^T & & & \\
& \nabla_{\mathbf{B}}\mathfrak{J}\big(\mathbf{B}^k\big)_2^T & & \\
& & \ddots & \\
& & & \nabla_{\mathbf{B}}\mathfrak{J}\big(\mathbf{B}^k\big)_M^T
\end{bmatrix}\in\mathbb{R}_+^{MP\times M},
\end{equation*}
where $\nabla_{\mathbf{B}}\mathfrak{J}\big(\mathbf{B}^k\big)_m$ is the $m$-th row of $\nabla_{\mathbf{B}}J(\mathbf{B}^k)= \mathbf{B}^k\mathbf{S}^k\mathbf{C}^k\mathbf{C}^{kT}\mathbf{S}^{kT} - \mathbf{AC}^{kT}\mathbf{S}^k + \beta\mathbf{B}^k\mathbf{B}^{kT}\mathbf{B}^k-\beta\mathbf{B}^k$. Then define:
\begin{equation*}
\mathbf{D} \equiv \mathrm{diag}\;\big(\mathbf{D}^1,\ldots,\mathbf{D}^M\big)\in\mathbb{R}_+^{MP\times MP},
\end{equation*}
where $\mathbf{D}^m$ is a diagonal matrix with its diagonal entries defined as:
\begin{equation*}
d_{pp}^m \equiv \left\{
 \begin{array}{ll}
   \frac{\big( \mathbf{\bar{B}}^k\mathbf{S}^k\mathbf{C}^k\mathbf{C}^{kT}\mathbf{S}^{kT} + \beta\mathbf{\bar{B}}^k\mathbf{\bar{B}}^{kT}\mathbf{\bar{B}}^k \big)_{mp}+\delta_{\mathbf{B}}^k}{\bar{b}_{mp}^k} & \mathrm{if}\;\; p\in \mathcal{I}_m \\
   \star & \mathrm{if}\;\; p\notin \mathcal{I}_m
 \end{array} \right.
\end{equation*}
with
\begin{align*}
\mathcal{I}_m \equiv \big\{p|&b_{mp}^k>0,\;\nabla_{\mathbf{B}}J\big(\mathbf{B}^k\big)_{mp}\ne 0,\;\mathrm{or} \\
&b_{mp}^k=0,\;\nabla_{\mathbf{B}}J\big(\mathbf{B}^k\big)_{mp} < 0\big\}
\end{align*}
is the set of non-KKT indices in $m$-th row of $\mathbf{B}^k$, and $\star$ is defined as before.

Then, the auxiliary function $\mathfrak{G}$ can be defined as:
\begin{align}
\mathfrak{G}\big(\mathfrak{B}^T,\mathfrak{B}^{kT}\big) \equiv \;&\mathfrak{J}\big(\mathfrak{B}^{kT}\big) + \mathrm{tr}\;\big\{\big(\mathfrak{B}-\mathfrak{B}^k\big)\nabla_{\mathfrak{B}^T}\mathfrak{J}\big(\mathfrak{B}^{kT}\big)\big\} \nonumber \\
&+ \frac{1}{2}\mathrm{tr}\;\big\{\big(\mathfrak{B}-\mathfrak{B}^k\big)\mathbf{D}\big(\mathfrak{B}-\mathfrak{B}^k\big)^T\big\}. \label{eq109}
\end{align}
Note that whenever $\mathbf{X}^{(k+1)}$ is a variable, we remove the $(k+1)$ sign, and
\begin{equation*}
\nabla_{\mathfrak{B}^T}\mathfrak{G}\big(\mathfrak{B}^T,\mathfrak{B}^{kT}\big)=\mathbf{D}\big(\mathfrak{B}-\mathfrak{B}^k\big)^T + \nabla_{\mathfrak{B}^T}\mathfrak{J}\big(\mathfrak{B}^{kT}\big).
\end{equation*}
By definition, $\mathbf{D}$ is positive definite for all $\mathbf{B}^k$ not satisfy the KKT conditions, so $\mathfrak{G}\big(\mathfrak{B}^T,\mathfrak{B}^{kT}\big)$ is a strict convex function, and consequently has a unique minimum.
\begin{align}
\mathbf{D}\big(\mathfrak{B}-\mathfrak{B}^k\big)^T + \nabla_{\mathfrak{B}^T}\mathfrak{J}\big(\mathfrak{B}^{kT}\big)=0, \label{eq110}\\
\mathfrak{B}^T = \mathfrak{B}^{kT} - \mathbf{D}^{-1}\nabla_{\mathfrak{B}^T}\mathfrak{J}\big(\mathfrak{B}^{kT}\big), \nonumber
\end{align}
which is exactly the update rule for $\mathbf{B}^k$ in eq.~\ref{eq98}. 

By using the Taylor series expansion, $\mathfrak{J}\big(\mathfrak{B}^T\big)$ can also be written as:
\begin{align}
\mathfrak{J}\big(\mathfrak{B}^T\big) = &\;\mathfrak{J}\big(\mathfrak{B}^{kT}\big) + \mathrm{tr}\;\big\{\big(\mathfrak{B}-\mathfrak{B}^k\big)\nabla_{\mathfrak{B}^T}\mathfrak{J}\big(\mathfrak{B}^{kT}\big)\big\} + \nonumber\\
&\;\frac{1}{2}\mathrm{tr}\;\big\{\big(\mathfrak{B}-\mathfrak{B}^k\big)\nabla_{\mathbf{B}}^2\mathbf{J}\big(\mathbf{B}^k\big)\big(\mathfrak{B}-\mathfrak{B}^k\big)^T\big\} + \varepsilon_{\mathbf{B}}^k, \label{eq111}
\end{align}
where
\begin{align*}
\varepsilon_{\mathbf{B}}^k = &\frac{1}{6}\mathrm{tr}\;\big\{\big(\mathfrak{B}-\mathfrak{B}^k\big)\big( 6\beta\mathfrak{B}^{kT} \big) \big(\mathfrak{B}-\mathfrak{B}^k\big) \big(\mathfrak{B}-\mathfrak{B}^k\big)^T\big\} + \\
&\frac{1}{24}\mathrm{tr}\;\big\{\big(\mathfrak{B}-\mathfrak{B}^k\big)\big(\mathfrak{B}-\mathfrak{B}^k\big)^T\big( 6\beta\mathbf{I} \big) \big(\mathfrak{B}-\mathfrak{B}^k\big) \big(\mathfrak{B}-\mathfrak{B}^k\big)^T\big\}
\end{align*}
and
\begin{equation*}
\nabla_{\mathbf{B}}^2\mathbf{J}\big(\mathbf{B}^k\big) \equiv
\begin{bmatrix}
\nabla_{\mathbf{B}}^2 J \big(\mathbf{B}^k\big) & & \\
& \ddots & \\
& & \nabla_{\mathbf{B}}^2 J \big(\mathbf{B}^k\big)
\end{bmatrix}\in\mathbb{R}_+^{MP\times MP}
\end{equation*}
with $\nabla_{\mathbf{B}}^2 J \big(\mathbf{B}^k\big)=\mathbf{S}^k\mathbf{C}^k\mathbf{C}^{kT}\mathbf{S}^{kT}+3\beta\mathbf{B}^{kT}\mathbf{B}^k-\beta\mathbf{I}$ components are arranged along its diagonal area (there are $M$ components).

Then, for $\mathfrak{G}$ to be the auxiliary function, we must prove:
\begin{enumerate}
\item $\mathfrak{G}\big(\mathfrak{B}^T,\mathfrak{B}^T\big)=\mathfrak{J}\big(\mathfrak{B}^T\big)$,
\item $\mathfrak{G}\big(\mathfrak{B}^{kT},\mathfrak{B}^{kT}\big)=\mathfrak{J}\big(\mathfrak{B}^{kT}\big)$,
\item $\mathfrak{G}\big(\mathfrak{B}^T,\mathfrak{B}^T\big) \le \mathfrak{G}\big(\mathfrak{B}^T,\mathfrak{B}^{kT}\big)$, and
\item $\mathfrak{G}\big(\mathfrak{B}^T,\mathfrak{B}^{kT}\big) \le \mathfrak{G}\big(\mathfrak{B}^{kT},\mathfrak{B}^{kT}\big)$,
\end{enumerate}
so that $\mathfrak{J}\big(\mathfrak{B}^T\big) \le \mathfrak{J}\big(\mathfrak{B}^{kT}\big)$. This implies $J\big(\mathbf{B}^{(k+1)}\big) \le J\big(\mathbf{B}^k\big)$. The first and second will be proven in theorem \ref{theorem18}, the third in theorem \ref{theorem19}, the fourth in theorem \ref{theorem20}, and the boundedness of $\mathbf{B}^k$, $\mathbf{C}^k$, and $\mathbf{S}^k$ will be proven in theorem \ref{theorem32}.

\begin{theorem} \label{theorem18}
$\mathfrak{G}\big(\mathfrak{B}^T,\mathfrak{B}^T\big)=\mathfrak{J}\big(\mathfrak{B}^T\big)$ and $\mathfrak{G}\big(\mathfrak{B}^{kT},\mathfrak{B}^{kT}\big)=\mathfrak{J}\big(\mathfrak{B}^{kT}\big)$.
\end{theorem}
\begin{proof}
These are obvious from the definition of $\mathfrak{G}$ in eq.~\ref{eq109}.
\end{proof}

\begin{theorem} \label{theorem19}
Given sufficiently large $\delta_{\mathbf{B}}^k$ and the boundedness of $\mathbf{B}^k$, $\mathbf{C}^k$, and $\mathbf{S}^k$, then it can be shown that $\mathfrak{G}\big(\mathfrak{B},\mathfrak{B}\big) \le \mathfrak{G}\big(\mathfrak{B},\mathfrak{B}^k\big)$. Moreover, if and only if $\mathbf{B}^k$ satisfies the KKT conditions, then the equality holds.
\end{theorem}
\begin{proof}
As $\mathfrak{G}\big(\mathfrak{B},\mathfrak{B}\big)=\mathfrak{J}\big(\mathfrak{B}\big)$, we need to show that $\mathfrak{G}\big(\mathfrak{B},\mathfrak{B}^k\big)-\mathfrak{J}\big(\mathfrak{B}\big) \ge 0$ for sufficiently large $\delta_{\mathbf{B}}^k$. By substracting eq.~\ref{eq109} from eq.~\ref{eq111}, we get:
\begin{align}
\mathfrak{G}\big(\mathfrak{B},\mathfrak{B}^k\big)-\mathfrak{J}\big(\mathfrak{B}\big)&=\frac{1}{2}\,\mathrm{tr}\,\big\{\big(\mathfrak{B}-\mathfrak{B}^k\big)\big(\mathbf{D}-\nabla_{\mathbf{B}}^2\mathbf{J}\big(\mathbf{B}^k\big)\big)\big(\mathfrak{B}-\mathfrak{B}^k\big)^T\big\} - \mathbf{\varepsilon}_{\mathbf{B}}^k \nonumber\\
&=\frac{1}{2}\sum_{m=1}^M\left[\big(\mathfrak{b}_m-\mathfrak{b}_m^k\big)\big(\mathbf{D}^m-\nabla_{\mathbf{B}}^2 J \big(\mathbf{B}^k\big)\big)\big(\mathfrak{b}_m-\mathfrak{b}_m^k\big)^T\right] - \mathbf{\varepsilon}_{\mathbf{B}}^k. \label{eq112}
\end{align}

Let $\mathbf{v}_m^T = \mathfrak{b}_m - \mathfrak{b}_m^k$, then:
\begin{align*}
\mathbf{v}_m^T\big(\mathbf{D}^m-\nabla_{\mathbf{B}}^2J\big(\mathbf{B}^k\big)\big)\mathbf{v}_m &= \mathbf{v}_m^T\big(\mathbf{D}^m + \beta\mathbf{I} - \big( \mathbf{S}^k\mathbf{C}^k\mathbf{C}^{kT}\mathbf{S}^{kT} + 3\beta\mathbf{B}^{kT}\mathbf{B}^k\big)\big)\mathbf{v}_m \\
&= \mathbf{v}_m^T\big(\mathbf{\bar{D}}^m + \delta_{\mathbf{B}}^k\mathbf{\hat{D}}^n + \beta\mathbf{I} - \big( \mathbf{S}^k\mathbf{C}^k\mathbf{C}^{kT}\mathbf{S}^{kT} + 3\beta\mathbf{B}^{kT}\mathbf{B}^k \big)\big)\mathbf{v}_m,
\end{align*}
where $\mathbf{\bar{D}}^m$ and $\delta_{\mathbf{B}}^k\mathbf{\hat{D}}^m$ are diagonal matrices that summed up to $\mathbf{D}^m$, with
\begin{align*}
\bar{d}_{pp}^m &\equiv \left\{
 \begin{array}{ll}
   \frac{\big( \mathbf{\bar{B}}^k\mathbf{S}^k\mathbf{C}^k\mathbf{C}^{kT}\mathbf{S}^{kT} + \beta\mathbf{\bar{B}}^k\mathbf{\bar{B}}^{kT}\mathbf{\bar{B}}^k \big)_{mp}}{\bar{b}_{mp}^k} & \mathrm{if}\;\; p\in \mathcal{I}_m \\
   \star & \mathrm{if}\;\; p\notin \mathcal{I}_m,
 \end{array} \right.
\text{and}\;
\hat{d}_{pp}^m &\equiv \left\{
 \begin{array}{ll}
   \frac{1}{\bar{b}_{mp}^k} & \mathrm{if}\;\; p\in \mathcal{I}_m \\
   \star & \mathrm{if}\;\; p\notin \mathcal{I}_m.
 \end{array} \right.
\end{align*}
Accordingly,
\begin{align}
\mathfrak{G}\big(\mathfrak{B},\mathfrak{B}^k\big)-\mathfrak{J}\big(\mathfrak{B}\big) = &\frac{1}{2} \sum_{m=1}^M \left\{\sum_{p=1}^P v_{mp}^2 \bar{d}_{pp}^m + \delta_{\mathbf{B}}^k \sum_{p=1}^P v_{mp}^2 \hat{d}_{pp}^m + \beta \sum_{p=1}^P v_{mp}^2 \right\} \nonumber \\
&- \frac{1}{2} \sum_{m=1}^M \mathbf{v}_m^T \big( \mathbf{S}^k\mathbf{C}^k\mathbf{C}^{kT}\mathbf{S}^{kT} + 3\beta\mathbf{B}^{kT}\mathbf{B}^k \big)\mathbf{v}_m - \varepsilon_{\mathbf{B}}^k. \label{eq113}
\end{align}
As shown, with the boundedness of $\mathbf{B}^k$, $\mathbf{C}^k$, and $\mathbf{S}^k$ and by sufficiently large $\delta_{\mathbf{B}}^k$, $\mathfrak{G}\big(\mathfrak{B},\mathfrak{B}\big) \le \mathfrak{G}\big(\mathfrak{B},\mathfrak{B}^k\big)$ can be guaranteed. Next we prove that if and only if $\mathbf{B}^k$ satisfies the KKT conditions, then the equality holds. 

If $\mathbf{B}^k$ satisfies the KKT conditions, then this is obvious by eq.~\ref{eq112} regardless of $\delta_{\mathbf{B}}^k$. And by eq.~\ref{eq113}, since $\delta_{\mathbf{B}}^k$ is a variable, the equality happens if and only if $\mathbf{B} = \mathbf{B}^k$, which by the update rule in eq.~\ref{eq98} happens if and only if $\mathbf{B}^k$ satisfies the KKT conditions. This completes the proof.
\end{proof}

\begin{theorem} \label{theorem20}
$\mathfrak{G}\big(\mathfrak{B},\mathfrak{B}^k\big) \le \mathfrak{G}\big(\mathfrak{B}^k,\mathfrak{B}^k\big)$. Moreover if and only if $\mathbf{B}^k$ satisfies the KKT conditions in eq.~\ref{eq91}, then $\mathfrak{G}\big(\mathfrak{B},\mathfrak{B}^k\big) = \mathfrak{G}\big(\mathfrak{B}^k,\mathfrak{B}^k\big)$.
\end{theorem}
\begin{proof}
\begin{align*}
\mathfrak{G}\big(\mathfrak{B}^k,\mathfrak{B}^k\big) - \mathfrak{G}\big(\mathfrak{B},\mathfrak{B}^k\big) =\;&-\mathrm{tr}\;\big\{\big(\mathfrak{B}-\mathfrak{B}^k\big)\nabla_{\mathfrak{B}}\mathfrak{J}\big(\mathfrak{B}^{kT}\big)\big\} \\
&- \frac{1}{2}\mathrm{tr}\;\big\{\big(\mathfrak{B}-\mathfrak{B}^k\big)\mathbf{D}\big(\mathfrak{B}-\mathfrak{B}^k\big)^T\big\}.
\end{align*}
By using eq.~\ref{eq110} and the fact that $\mathbf{D}$ is positive semi-definite:
\begin{equation*}
\mathfrak{G}\big(\mathfrak{B}^k,\mathfrak{B}^k\big) - \mathfrak{G}\big(\mathfrak{B},\mathfrak{B}^k\big) = \frac{1}{2}\mathrm{tr}\;\big\{\big(\mathfrak{B}-\mathfrak{B}^k\big)\mathbf{D}\big(\mathfrak{B}-\mathfrak{B}^k\big)^T\big\} \ge 0. 
\end{equation*}
By the update rule eq.~\ref{eq98}, if $\mathbf{B}^k$ satisfies the KKT conditions, then $\mathbf{B}=\mathbf{B}^k$, and therefore the equality holds. Now we need to prove that if the equality holds, then $\mathbf{B}^k$ satisfies the KKT conditions.

To prove this, let consider a contradiction situation where the equality holds but $\mathbf{B}^k$ does not satisfy the KKT conditions. In this case, there exists at least an index $(m,p)$ such that:
\begin{equation*}
b_{mp}\ne b_{mp}^k\;\;\mathrm{and}\;\; d_{pp}^m = \frac{\big( \mathbf{\bar{B}}^k\mathbf{S}^k\mathbf{C}^k\mathbf{C}^{kT}\mathbf{S}^{kT} + \beta\mathbf{\bar{B}}^k\mathbf{\bar{B}}^{kT}\mathbf{\bar{B}}^k \big)_{mp}+\delta_{\mathbf{B}}^k}{\bar{b}_{mp}^k} \ge \frac{\delta_{\mathbf{B}}^k}{\bar{b}_{mp}^k}.
\end{equation*}
Note that by definition in eq.~\ref{eq101}, if $\bar{b}_{mp}^k$ is equal to zero, then $b_{mp}= b_{mp}^k$ which violates the condition for the contradiction, so $\bar{b}_{mp}^k$ cannot be equal to zero. Consequently,
\begin{equation*}
\mathfrak{G}\big(\mathfrak{B}^k,\mathfrak{B}^k\big)-\mathfrak{G}\big(\mathfrak{B},\mathfrak{B}^k\big) \ge \frac{\big(b_{mp}-b_{mp}^k\big)^2\delta_{\mathbf{B}}^k}{\bar{b}_{mp}^k} > 0,
\end{equation*}
which violates the equality. Thus, it is proven that if the equality holds, then $\mathbf{B}^k$ satisfies the KKT conditions.
\end{proof}

\begin{theorem}\label{theorem21}
Given sufficiently large $\delta_{\mathbf{B}}^k$ and the boundedness of $\mathbf{B}^k$, $\mathbf{C}^k$, and $\mathbf{S}^k$, $J\big(\mathbf{B}^{k+1}\big)$ $\le$ $J\big(\mathbf{B}^k\big)\;\,\forall k\ge 0$ under update rule eq.~\ref{eq98} with the equality happens if and only if $\mathbf{B}^k$ satisfies the KKT conditions in eq.~\ref{eq91}. 
\end{theorem}
\begin{proof}
This theorem is the corollary of theorem \ref{theorem18}, \ref{theorem19}, and \ref{theorem20}
\end{proof}

\paragraph{B.~The nonincreasing property of $J\big(\mathbf{C}^k\big)$}

Next we prove the nonincreasing property of $J\big(\mathbf{C}^k\big)$, i.e., $J\big(\mathbf{C}^{(k+1)}\big)\le J\big(\mathbf{C}^k\big)\;\forall k\ge 0$.

By using the auxiliary function approach, the nonincreasing property of $J\big(\mathbf{C}^k\big)$ can be proven by showing that:
\begin{equation*}
J\big(\mathbf{C}^{(k+1)}\big)= G\big(\mathbf{C}^{(k+1)},\mathbf{C}^{(k+1)}\big) \le G\big(\mathbf{C}^{(k+1)},\mathbf{C}^k\big) \le G\big(\mathbf{C}^k,\mathbf{C}^k\big) = J\big(\mathbf{C}^k\big). 
\end{equation*}
To define $G$, $\mathbf{C}$ is rearranged into:
\begin{equation*}
\mathfrak{C} \equiv
\begin{bmatrix}
\mathbf{c}_1 & & & \\
& \mathbf{c}_2 & & \\
& & \ddots & \\
& & & \mathbf{c}_N
\end{bmatrix}\in\mathbb{R}_+^{NQ\times N},
\end{equation*}
where $\mathbf{c}_n$ is the $n$-th column of $\mathbf{C}$. And also let's define:
\begin{equation*}
\nabla_{\mathfrak{C}}\mathfrak{J}\big(\mathfrak{C}^k\big) \equiv
\begin{bmatrix}
\nabla_{\mathbf{C}}\mathfrak{J}\big(\mathbf{C}^k\big)_1 & & & \\
& \nabla_{\mathbf{C}}\mathfrak{J}\big(\mathbf{C}^k\big)_2 & & \\
& & \ddots & \\
& & & \nabla_{\mathbf{C}}\mathfrak{J}\big(\mathbf{C}^k\big)_N
\end{bmatrix}\in\mathbb{R}_+^{NQ\times N},
\end{equation*}
where $\nabla_{\mathbf{C}}\mathfrak{J}\big(\mathbf{C}^k\big)_n$ is the $n$-th column of $\nabla_{\mathbf{C}}J(\mathbf{C}^k)=\mathbf{S}^{kT}\mathbf{B}^{(k+1)T}\mathbf{B}^{(k+1)}\mathbf{S}^k\mathbf{C}^k-\mathbf{S}^{kT}\mathbf{B}^{(k+1)T}\mathbf{A}+\alpha \mathbf{C}^k\mathbf{C}^{kT}\mathbf{C}^k-\alpha \mathbf{C}^k$. And:
\begin{equation*}
\mathbf{D} \equiv \mathrm{diag}\;\big(\mathbf{D}^1,\ldots,\mathbf{D}^N\big)\in\mathbb{R}_+^{NQ\times NQ}, 
\end{equation*}
where $\mathbf{D}^n$ is a diagonal matrix with its diagonal entries defined as:
\begin{equation*}
d_{qq}^n \equiv \left\{
 \begin{array}{ll}
   \frac{\big(\mathbf{S}^{kT}\mathbf{B}^{(k+1)T}\mathbf{B}^{(k+1)}\mathbf{S}^k\mathbf{\bar{C}}^k + \alpha \mathbf{\bar{C}}^k\mathbf{\bar{C}}^{kT}\mathbf{\bar{C}}^k\big)_{qn}+\delta_{\mathbf{C}}^k}{\bar{c}_{qn}^k} & \mathrm{if}\;\; q\in \mathcal{I}_n \\
   \star & \mathrm{if}\;\; q\notin \mathcal{I}_n
 \end{array} \right.
\end{equation*}
with
\begin{align*}
\mathcal{I}_n \equiv \big\{q|&c_{qn}^k>0,\;\nabla_{\mathbf{C}}J\big(\mathbf{C}^k\big)_{qn}\ne 0,\;\mathrm{or} \\
&c_{qn}^k=0,\;\nabla_{\mathbf{C}}J\big(\mathbf{C}^k\big)_{qn} < 0\big\}
\end{align*}
is the set of non-KKT indices in $n$-th column of $\mathbf{C}^k$, and $\star$ is defined as before.

Then, the auxiliary function $\mathfrak{G}$ can be written as:
\begin{equation}
\mathfrak{G}\big(\mathfrak{C},\mathfrak{C}^k\big) \equiv \;\mathfrak{J}\big(\mathfrak{C}^k\big) + \mathrm{tr}\;\big\{\big(\mathfrak{C}-\mathfrak{C}^k\big)^T\nabla_{\mathfrak{C}}\mathfrak{J}\big(\mathfrak{C}^k\big)\big\} + \frac{1}{2}\mathrm{tr}\;\big\{\big(\mathfrak{C}-\mathfrak{C}^k\big)^T\mathbf{D}\big(\mathfrak{C}-\mathfrak{C}^k\big)\big\}. \label{115}
\end{equation}
Also:
\begin{equation*}
\nabla_{\mathfrak{C}}\mathfrak{G}\big(\mathfrak{C},\mathfrak{C}^k\big)=\mathbf{D}\big(\mathfrak{C}-\mathfrak{C}^k\big) + \nabla_{\mathfrak{C}}\mathfrak{J}\big(\mathfrak{C}^k\big).
\end{equation*}
Since $\mathfrak{G}\big(\mathfrak{C},\mathfrak{C}^k\big)$ is a strict convex function, it has a unique minimum.
\begin{align}
\mathbf{D}\big(\mathfrak{C}-\mathfrak{C}^k\big) + \nabla_{\mathfrak{C}}\mathfrak{J}\big(\mathfrak{C}^k\big)=0, \label{116}\\
\mathfrak{C} = \mathfrak{C}^k - \mathbf{D}^{-1}\nabla_{\mathfrak{C}}\mathfrak{J}\big(\mathfrak{C}^k\big), \nonumber
\end{align}
which is exactly the update rule for $\mathbf{C}$ in eq.~\ref{eq99}.

By using the Taylor series, alternative formulation for $\mathfrak{J}\big(\mathfrak{C}\big)$ can be written as:
\begin{align}
\mathfrak{J}\big(\mathfrak{C}\big) = &\;\mathfrak{J}\big(\mathfrak{C}^k\big) + \mathrm{tr}\;\big\{\big(\mathfrak{C}-\mathfrak{C}^k\big)^T\nabla_{\mathfrak{C}}\mathfrak{J}\big(\mathfrak{C}^k\big)\big\} + \nonumber \\
&\;\frac{1}{2}\mathrm{tr}\;\big\{\big(\mathfrak{C}-\mathfrak{C}^k\big)^T\nabla_{\mathbf{C}}^2\mathbf{J}\big(\mathbf{C}^k\big)\big(\mathfrak{C}-\mathfrak{C}^k\big)\big\} + \mathbf{\varepsilon}_{\mathbf{C}}^k \label{117}
\end{align}
where $\mathbf{\varepsilon}_{\mathbf{C}}^k$ is the higher components of the Taylor series:
\begin{align*}
\mathbf{\varepsilon}_{\mathbf{C}}^k = &\;\frac{1}{6}\mathrm{tr}\;\big\{\big(\mathfrak{C}-\mathfrak{C}^k\big)^T(6\alpha\mathfrak{C}^k\big)\big(\mathfrak{C}-\mathfrak{C}^k\big)^T\big(\mathfrak{C}-\mathfrak{C}^k\big)\big\} + \\
&\;\frac{1}{24}\mathrm{tr}\;\big\{\big(\mathfrak{C}-\mathfrak{C}^k\big)^T\big(\mathfrak{C}-\mathfrak{C}^k\big)(6\alpha\mathbf{I})\big(\mathfrak{C}-\mathfrak{C}^k\big)^T\big(\mathfrak{C}-\mathfrak{C}^k\big)\big\},
\end{align*}
and
\begin{equation*}
\nabla_{\mathbf{C}}^2\mathbf{J}\big(\mathbf{C}^k\big) \equiv
\begin{bmatrix}
\nabla_{\mathbf{C}}^2 J \big(\mathbf{C}^k\big) & & \\
& \ddots & \\
& & \nabla_{\mathbf{C}}^2 J \big(\mathbf{C}^k\big)
\end{bmatrix}\in\mathbb{R}_+^{NQ\times NQ}
\end{equation*}
with $\nabla_{\mathbf{C}}^2 J \big(\mathbf{C}^k\big)=\mathbf{S}^{kT}\mathbf{B}^{(k+1)T}\mathbf{B}^{(k+1)}\mathbf{S}^k + 3\alpha\mathbf{C}^k\mathbf{C}^{kT}-\alpha\mathbf{I}$ components are arranged along its diagonal area (there are $N$ components).

As before, for $\mathfrak{G}$ to be the auxiliary function, we must prove:
\begin{enumerate}
\item $\mathfrak{G}\big(\mathfrak{C},\mathfrak{C}\big)=\mathfrak{J}\big(\mathfrak{C}\big)$,
\item $\mathfrak{G}\big(\mathfrak{C}^k,\mathfrak{C}^k\big)=\mathfrak{J}\big(\mathfrak{C}^k\big)$,
\item $\mathfrak{G}\big(\mathfrak{C},\mathfrak{C}\big) \le \mathfrak{G}\big(\mathfrak{C},\mathfrak{C}^k\big)$, and
\item $\mathfrak{G}\big(\mathfrak{C},\mathfrak{C}^k\big) \le \mathfrak{G}\big(\mathfrak{C}^k,\mathfrak{C}^k\big)$,
\end{enumerate}
The first and second will be proven in theorem \ref{theorem22}, the third in theorem \ref{theorem23}, and the fourth in theorem \ref{theorem24}.

\begin{theorem} \label{theorem22}
$\mathfrak{G}\big(\mathfrak{C},\mathfrak{C}\big)=\mathfrak{J}\big(\mathfrak{C}\big)$, and $\mathfrak{G}\big(\mathfrak{C}^k,\mathfrak{C}^k\big)=\mathfrak{J}\big(\mathfrak{C}^k\big)$,
\end{theorem}
\begin{proof}
These are obvious from the definition of $\mathfrak{G}$ in eq.~\ref{115}.
\end{proof}

\begin{theorem} \label{theorem23}
Given sufficiently large $\delta_{\mathbf{C}}^k$ and the boundedness of $\mathbf{B}^k$, $\mathbf{C}^k$, and $\mathbf{S}^k$, then it can be shown that $\mathfrak{G}\big(\mathfrak{C},\mathfrak{C}\big) \le \mathfrak{G}\big(\mathfrak{C},\mathfrak{C}^k\big)$. Moreover, if and only if $\mathbf{C}^k$ satisfies the KKT conditions, then the equality holds.
\end{theorem}
\begin{proof}
As $\mathfrak{G}\big(\mathfrak{C},\mathfrak{C}\big)=\mathfrak{J}\big(\mathfrak{C}\big)$, we need to show that $\mathfrak{G}\big(\mathfrak{C},\mathfrak{C}^k\big)-\mathfrak{J}\big(\mathfrak{C}\big) \ge 0$. By substracting eq.~\ref{115} from eq.~\ref{117}, we get:
\begin{align}
\mathfrak{G}\big(\mathfrak{C},\mathfrak{C}^k\big)-\mathfrak{J}\big(\mathfrak{C}\big)&=\frac{1}{2}\,\mathrm{tr}\,\big\{\big(\mathfrak{C}-\mathfrak{C}^k\big)^T\big(\mathbf{D}-\nabla_{\mathbf{C}}^2\mathbf{J}\big(\mathbf{C}^k\big)\big)\big(\mathfrak{C}-\mathfrak{C}^k\big)\big\} - \mathbf{\varepsilon}_{\mathbf{C}}^k \nonumber\\
&=\frac{1}{2}\sum_{n=1}^N\left[\big(\mathbf{c}_n-\mathbf{c}_n^k\big)^T\big(\mathbf{D}^n-\nabla_{\mathbf{C}}^2 J \big(\mathbf{C}^k\big)\big)\big(\mathbf{c}_n-\mathbf{c}_n^k\big)\right] - \mathbf{\varepsilon}_{\mathbf{C}}^k. \label{118}
\end{align}
Let $\mathbf{v}_n = \mathbf{c}_n - \mathbf{c}_n^k$, then:
\begin{align*}
\mathbf{v}_n^T\big(\mathbf{D}^n-\nabla_{\mathbf{C}}^2J\big(\mathbf{C}^k\big)\big)\mathbf{v}_n &= \mathbf{v}_n^T\big(\mathbf{D}^n + \alpha\mathbf{I} - \big(\mathbf{S}^{kT}\mathbf{B}^{(k+1)T}\mathbf{B}^{(k+1)}\mathbf{S}^k + 3\alpha\mathbf{C}^k\mathbf{C}^{kT}\big)\big)\mathbf{v}_n \\
&= \mathbf{v}_n^T\big(\mathbf{\bar{D}}^n + \delta_{\mathbf{C}}^k\mathbf{\hat{D}}^n + \alpha\mathbf{I} - \big(\mathbf{S}^{kT}\mathbf{B}^{(k+1)T}\mathbf{B}^{(k+1)}\mathbf{S}^k + 3\alpha\mathbf{C}^k\mathbf{C}^{kT}\big)\big)\mathbf{v}_n,
\end{align*}
where $\mathbf{\bar{D}}^n$ and $\delta_{\mathbf{C}}^k\mathbf{\hat{D}}^n$ are diagonal matrices that summed up to $\mathbf{D}^n$, with
\begin{align*}
\bar{d}_{qq}^n &\equiv \left\{
 \begin{array}{ll}
   \frac{\big( \mathbf{S}^{kT}\mathbf{B}^{(k+1)T}\mathbf{B}^{(k+1)}\mathbf{S}^k\mathbf{\bar{C}}^k + \alpha\mathbf{\bar{C}}^k\mathbf{\bar{C}}^{kT}\mathbf{\bar{C}}^k \big)_{qn}}{\bar{c}_{qn}^k} & \mathrm{if}\;\; q\in \mathcal{I}_n \\
   \star & \mathrm{if}\;\; q\notin \mathcal{I}_n,
 \end{array} \right.
\text{and}\;
\hat{d}_{qq}^n &\equiv \left\{
 \begin{array}{ll}
   \frac{1}{\bar{c}_{qn}^k} & \mathrm{if}\;\; q\in \mathcal{I}_n \\
   \star & \mathrm{if}\;\; q\notin \mathcal{I}_n.
 \end{array} \right.
\end{align*}
Accordingly,
\begin{align}
\mathfrak{G}\big(\mathfrak{C},\mathfrak{C}^k\big)-\mathfrak{J}\big(\mathfrak{C}\big) = &\frac{1}{2} \sum_{n=1}^N \left\{\sum_{q=1}^Q v_{qn}^2 \bar{d}_{qq}^n + \delta_{\mathbf{C}}^k \sum_{q=1}^Q v_{qn}^2 \hat{d}_{qq}^n + \alpha \sum_{q=1}^Q v_{qn}^2 \right\} \nonumber \\
&- \frac{1}{2} \sum_{n=1}^N \mathbf{v}_n^T \big( \mathbf{S}^{kT}\mathbf{B}^{(k+1)T}\mathbf{B}^{(k+1)}\mathbf{S}^k + 3\alpha\mathbf{C}^k\mathbf{C}^{kT}\big)\mathbf{v}_n - \varepsilon_{\mathbf{C}}^k. \label{119}
\end{align}
As shown, with the boundedness of $\mathbf{B}^k$, $\mathbf{C}^k$, and $\mathbf{S}^k$, and by sufficiently large $\delta_{\mathbf{C}}^k$, $\mathfrak{G}\big(\mathfrak{C},\mathfrak{C}\big) \le \mathfrak{G}\big(\mathfrak{C},\mathfrak{C}^k\big)$ can be guaranteed. Next we prove that if and only if $\mathbf{C}^k$ satisfies the KKT conditions, then the equality holds. 

If $\mathbf{C}^k$ satisfies the KKT conditions, then this is obvious by eq.~\ref{118} regardless of $\delta_{\mathbf{C}}^k$. And by eq.~\ref{119}, since $\delta_{\mathbf{C}}^k$ is a variable, the equality happens if and only if $\mathbf{C} = \mathbf{C}^k$ which by the update rule in eq.~\ref{eq99} happens if and only if $\mathbf{C}^k$ satisfies the KKT conditions. This completes the proof.
\end{proof}

\begin{theorem} \label{theorem24}
$\mathfrak{G}\big(\mathfrak{C},\mathfrak{C}^k\big) \le \mathfrak{G}\big(\mathfrak{C}^k,\mathfrak{C}^k\big)$. Moreover if and only if $\mathbf{C}^k$ satisfies the KKT conditions in eq.~\ref{eq54}, then $\mathfrak{G}\big(\mathfrak{C},\mathfrak{C}^k\big) = \mathfrak{G}\big(\mathfrak{C}^k,\mathfrak{C}^k\big)$.
\end{theorem}
\begin{proof}
\begin{equation*}
\mathfrak{G}\big(\mathfrak{C}^k,\mathfrak{C}^k\big) - \mathfrak{G}\big(\mathfrak{C},\mathfrak{C}^k\big) = -\mathrm{tr}\;\big\{\big(\mathfrak{C}-\mathfrak{C}^k\big)^T\nabla_{\mathfrak{C}}\mathfrak{J}\big(\mathfrak{C}^{kT}\big)\big\} - \frac{1}{2}\mathrm{tr}\;\big\{\big(\mathfrak{C}-\mathfrak{C}^k\big)^T\mathbf{D}\big(\mathfrak{C}-\mathfrak{C}^k\big)\big\}.
\end{equation*}
By using eq.~\ref{116} and the fact that $\mathbf{D}$ is positive semi-definite:
\begin{equation*}
\mathfrak{G}\big(\mathfrak{C}^k,\mathfrak{C}^k\big) - \mathfrak{G}\big(\mathfrak{C},\mathfrak{C}^k\big) = \frac{1}{2}\mathrm{tr}\;\big\{\big(\mathfrak{C}-\mathfrak{C}^k\big)^T\mathbf{D}\big(\mathfrak{C}-\mathfrak{C}^k\big)\big\} \ge 0, 
\end{equation*}
By the update rule eq.~\ref{eq99}, if $\mathbf{C}^k$ satisfies the KKT conditions, then $\mathbf{C}=\mathbf{C}^k$, and therefore the equality holds. Now we need to prove that if the equality holds, then $\mathbf{C}^k$ satisfies the KKT conditions.

To prove this, let consider a contradiction situation where the equality holds but $\mathbf{C}^k$ does not satisfy the KKT conditions. In this case, there exists at least an index $(q,n)$ such that:
\begin{equation*}
c_{qn}\ne c_{qn}^k\;\;\mathrm{and}\;\; d_{qq}^n = \frac{\big( \mathbf{S}^{kT}\mathbf{B}^{(k+1)T}\mathbf{B}^{(k+1)}\mathbf{S}^k\mathbf{\bar{C}}^k + \alpha \mathbf{\bar{C}}^k\mathbf{\bar{C}}^{kT}\mathbf{\bar{C}}^k\big)_{qn}+\delta_{\mathbf{C}}^k}{\bar{c}_{qn}^k} \ge \frac{\delta_{\mathbf{C}}^k}{\bar{c}_{qn}^k}.
\end{equation*}
Note that by the definition in eq.~\ref{eq102}, if $\bar{c}_{qn}^k$ is equal to zero, then $c_{qn}= c_{qn}^k$ which violates the condition for the contradiction, so $\bar{c}_{qn}^k$ cannot be equal to zero. Consequently,
\begin{equation*}
\mathfrak{G}\big(\mathfrak{C}^k,\mathfrak{C}^k\big)-\mathfrak{G}\big(\mathfrak{C},\mathfrak{C}^k\big) \ge \frac{\big(c_{qn}-c_{qn}^k\big)^2\delta_{\mathbf{C}}^k}{\bar{c}_{qn}^k} > 0,
\end{equation*}
which violates the equality. Thus, it is proven that if the equality holds, then $\mathbf{C}^k$ satisfies the KKT conditions.
\end{proof}

\begin{theorem}\label{theorem25}
Given sufficiently large $\delta_{\mathbf{C}}^k$ and the boundedness of $\mathbf{B}^k$, $\mathbf{C}^k$, and $\mathbf{S}^k$, $J\big(\mathbf{C}^{k+1}\big)$ $\le$ $J\big(\mathbf{C}^k\big)\;\,\forall k\ge 0$ under update rule eq.~\ref{eq99} with the equality happens if and only if $\mathbf{C}^k$ satisfies the KKT conditions in eq.~\ref{eq91}. 
\end{theorem}
\begin{proof}
This theorem is the corollary of theorem \ref{theorem22}, \ref{theorem23}, and \ref{theorem24}
\end{proof}

\paragraph{C.~The nonincreasing property of $J\big(\mathbf{S}^k\big)$}

Next we prove the nonincreasing property of $J\big(\mathbf{S}^k\big)$, i.e., $J\big(\mathbf{S}^{(k+1)}\big)\le J\big(\mathbf{S}^k\big)\;\forall k\ge 0$.

By using the auxiliary function approach, the nonincreasing property of $J\big(\mathbf{S}^k\big)$ can be proven by showing that:
\begin{equation*}
J\big(\mathbf{S}^{(k+1)}\big)= G\big(\mathbf{S}^{(k+1)},\mathbf{S}^{(k+1)}\big) \le G\big(\mathbf{S}^{(k+1)},\mathbf{S}^k\big) \le G\big(\mathbf{S}^k,\mathbf{S}^k\big) = J\big(\mathbf{S}^k\big). 
\end{equation*}
To define $G$, $\mathbf{S}$ is rearranged into:
\begin{equation*}
\mathfrak{S} \equiv
\begin{bmatrix}
\mathbf{s}_1 & & & \\
& \mathbf{s}_2 & & \\
& & \ddots & \\
& & & \mathbf{s}_Q
\end{bmatrix}\in\mathbb{R}_+^{PQ\times Q},
\end{equation*}
where $\mathbf{s}_q$ is the $q$-th column of $\mathbf{S}$. And also let's define:
\begin{equation*}
\nabla_{\mathfrak{S}}\mathfrak{J}\big(\mathfrak{S}^k\big) \equiv
\begin{bmatrix}
\nabla_{\mathbf{S}}\mathfrak{J}\big(\mathbf{S}^k\big)_1 & & & \\
& \nabla_{\mathbf{S}}\mathfrak{J}\big(\mathbf{S}^k\big)_2 & & \\
& & \ddots & \\
& & & \nabla_{\mathbf{S}}\mathfrak{J}\big(\mathbf{S}^k\big)_Q
\end{bmatrix}\in\mathbb{R}_+^{PQ\times Q},
\end{equation*}
where $\nabla_{\mathbf{S}}\mathfrak{J}\big(\mathbf{S}^k\big)_q$ is the $q$-th column of $\nabla_{\mathbf{S}}J(\mathbf{S}^k)=\mathbf{B}^{(k+1)T}\mathbf{B}^{(k+1)}\mathbf{S}^k\mathbf{C}^{(k+1)}\mathbf{C}^{(k+1)T}-\mathbf{B}^{(k+1)T}\mathbf{A}\mathbf{C}^{(k+1)T}$. And:
\begin{equation*}
\mathbf{D} \equiv \mathrm{diag}\;\big(\mathbf{D}^1,\ldots,\mathbf{D}^Q\big)\in\mathbb{R}_+^{PQ\times PQ}, 
\end{equation*}
where $\mathbf{D}^q$ is a diagonal matrix with its diagonal entries defined as:
\begin{equation*}
d_{pp}^q \equiv \left\{
 \begin{array}{ll}
   \frac{\big( \mathbf{B}^{(k+1)T}\mathbf{B}^{(k+1)}\mathbf{\bar{S}}^k\mathbf{C}^{(k+1)}\mathbf{C}^{(k+1)T} \big)_{pq}+\delta_{\mathbf{S}}^k}{\bar{s}_{pq}^k} & \mathrm{if}\;\; p\in \mathcal{I}_q \\
   \star & \mathrm{if}\;\; p\notin \mathcal{I}_q
 \end{array} \right.
\end{equation*}
with
\begin{align*}
\mathcal{I}_q \equiv \big\{p|&s_{pq}^k>0,\;\nabla_{\mathbf{S}}J\big(\mathbf{S}^k\big)_{pq}\ne 0,\;\mathrm{or} \\
&s_{pq}^k=0,\;\nabla_{\mathbf{S}}J\big(\mathbf{S}^k\big)_{pq} < 0\big\}
\end{align*}
is the set of non-KKT indices in $q$-th column of $\mathbf{S}^k$, and $\star$ is defined as before.

Then, the auxiliary function $\mathfrak{G}$ can be written as:
\begin{equation}
\mathfrak{G}\big(\mathfrak{S},\mathfrak{S}^k\big) \equiv \;\mathfrak{J}\big(\mathfrak{S}^k\big) + \mathrm{tr}\;\big\{\big(\mathfrak{S}-\mathfrak{S}^k\big)^T\nabla_{\mathfrak{S}}\mathfrak{J}\big(\mathfrak{S}^k\big)\big\} + \frac{1}{2}\mathrm{tr}\;\big\{\big(\mathfrak{S}-\mathfrak{S}^k\big)^T\mathbf{D}\big(\mathfrak{S}-\mathfrak{S}^k\big)\big\}. \label{123}
\end{equation}
Also:
\begin{equation*}
\nabla_{\mathfrak{S}}\mathfrak{G}\big(\mathfrak{S},\mathfrak{S}^k\big)=\mathbf{D}\big(\mathfrak{S}-\mathfrak{S}^k\big) + \nabla_{\mathfrak{S}}\mathfrak{J}\big(\mathfrak{S}^k\big).
\end{equation*}
Since $\mathfrak{G}\big(\mathfrak{S},\mathfrak{S}^k\big)$ is a strict convex function, it has a unique minimum.
\begin{align}
\mathbf{D}\big(\mathfrak{S}-\mathfrak{S}^k\big) + \nabla_{\mathfrak{S}}\mathfrak{J}\big(\mathfrak{S}^k\big)=0, \label{124}\\
\mathfrak{S} = \mathfrak{S}^k - \mathbf{D}^{-1}\nabla_{\mathfrak{S}}\mathfrak{J}\big(\mathfrak{S}^k\big), \nonumber
\end{align}
which is exactly the update rule for $\mathbf{S}$ in eq.~\ref{eq100}.

By using the Taylor series, alternative formulation for $\mathfrak{J}\big(\mathfrak{S}\big)$ can be written as:
\begin{equation}
\mathfrak{J}\big(\mathfrak{S}\big) = \;\mathfrak{J}\big(\mathfrak{S}^k\big) + \mathrm{tr}\;\big\{\big(\mathfrak{S}-\mathfrak{S}^k\big)^T\nabla_{\mathfrak{S}}\mathfrak{J}\big(\mathfrak{S}^k\big)\big\} + \frac{1}{2}\mathrm{tr}\;\big\{\big(\mathfrak{S}-\mathfrak{S}^k\big)^T\nabla_{\mathbf{S}}^2\mathbf{J}\big(\mathbf{S}^k\big)\big(\mathfrak{S}-\mathfrak{S}^k\big)\big\} \label{125}
\end{equation}
where
\begin{equation*}
\nabla_{\mathbf{S}}^2\mathbf{J}\big(\mathbf{S}^k\big) \equiv
\begin{bmatrix}
\nabla_{\mathbf{S}}^2 J \big(\mathbf{S}^k\big) & & \\
& \ddots & \\
& & \nabla_{\mathbf{S}}^2 J \big(\mathbf{S}^k\big)
\end{bmatrix}\in\mathbb{R}_+^{PQ\times PQ}
\end{equation*}
with $\nabla_{\mathbf{S}}^2 J \big(\mathbf{S}^k\big)=\mathbf{B}^{(k+1)T}\mathbf{B}^{(k+1)}\mathbf{C}^{(k+1)}\mathbf{C}^{(k+1)T}$ components are arranged along its diagonal area (there are $Q$ components).

For $\mathfrak{G}$ to be the auxiliary function, we must prove:
\begin{enumerate}
\item $\mathfrak{G}\big(\mathfrak{S},\mathfrak{S}\big)=\mathfrak{J}\big(\mathfrak{S}\big)$,
\item $\mathfrak{G}\big(\mathfrak{S}^k,\mathfrak{S}^k\big)=\mathfrak{J}\big(\mathfrak{S}^k\big)$,
\item $\mathfrak{G}\big(\mathfrak{S},\mathfrak{S}\big) \le \mathfrak{G}\big(\mathfrak{S},\mathfrak{S}^k\big)$, and
\item $\mathfrak{G}\big(\mathfrak{S},\mathfrak{S}^k\big) \le \mathfrak{G}\big(\mathfrak{S}^k,\mathfrak{S}^k\big)$,
\end{enumerate}
The first and second will be proven in theorem \ref{theorem26}, the third in theorem \ref{theorem27}, and the fourth in theorem \ref{theorem28}.

\begin{theorem} \label{theorem26}
$\mathfrak{G}\big(\mathfrak{S},\mathfrak{S}\big)=\mathfrak{J}\big(\mathfrak{S}\big)$, and $\mathfrak{G}\big(\mathfrak{S}^k,\mathfrak{S}^k\big)=\mathfrak{J}\big(\mathfrak{S}^k\big)$,
\end{theorem}
\begin{proof}
These are obvious from the definition of $\mathfrak{G}$ in eq.~\ref{123}.
\end{proof}

\begin{theorem} \label{theorem27}
Given sufficiently large $\delta_{\mathbf{S}}^k$ and the boundedness of $\mathbf{B}^k$, $\mathbf{C}^k$, and $\mathbf{S}^k$, then it can be shown that $\mathfrak{G}\big(\mathfrak{S},\mathfrak{S}\big) \le \mathfrak{G}\big(\mathfrak{S},\mathfrak{S}^k\big)$. Moreover, if and only if $\mathbf{S}^k$ satisfies the KKT conditions, then the equality holds.
\end{theorem}
\begin{proof}
As $\mathfrak{G}\big(\mathfrak{S},\mathfrak{S}\big)=\mathfrak{J}\big(\mathfrak{S}\big)$, we need to show that $\mathfrak{G}\big(\mathfrak{S},\mathfrak{S}^k\big)-\mathfrak{J}\big(\mathfrak{S}\big) \ge 0$. By substracting eq.~\ref{123} from eq.~\ref{125}, we get:
\begin{align}
\mathfrak{G}\big(\mathfrak{S},\mathfrak{S}^k\big)-\mathfrak{J}\big(\mathfrak{S}\big)&=\frac{1}{2}\,\mathrm{tr}\,\big\{\big(\mathfrak{S}-\mathfrak{S}^k\big)^T\big(\mathbf{D}-\nabla_{\mathbf{S}}^2\mathbf{J}\big(\mathbf{S}^k\big)\big)\big(\mathfrak{S}-\mathfrak{S}^k\big)\big\} \nonumber \\
&=\frac{1}{2}\sum_{q=1}^Q\left[\big(\mathbf{s}_q-\mathbf{s}_q^k\big)^T\big(\mathbf{D}^q-\nabla_{\mathbf{S}}^2 J \big(\mathbf{S}^k\big)\big)\big(\mathbf{s}_q-\mathbf{s}_q^k\big)\right]. \label{126}
\end{align}
Let $\mathbf{v}_q = \mathbf{s}_q - \mathbf{s}_q^k$, then:
\begin{align*}
\mathbf{v}_q^T\big(\mathbf{D}^q-\nabla_{\mathbf{S}}^2J\big(\mathbf{S}^k\big)\big)\mathbf{v}_q &= \mathbf{v}_q^T\big(\mathbf{D}^q - \big( \mathbf{B}^{(k+1)T}\mathbf{B}^{(k+1)}\mathbf{C}^{(k+1)}\mathbf{C}^{(k+1)T} \big)\big)\mathbf{v}_q \\
&= \mathbf{v}_q^T\big(\mathbf{\bar{D}}^q + \delta_{\mathbf{S}}^k\mathbf{\hat{D}}^q - \big( \mathbf{B}^{(k+1)T}\mathbf{B}^{(k+1)}\mathbf{C}^{(k+1)}\mathbf{C}^{(k+1)T} \big)\big)\mathbf{v}_q,
\end{align*}
where $\mathbf{\bar{D}}^q$ and $\delta_{\mathbf{S}}^k\mathbf{\hat{D}}^q$ are diagonal matrices that summed up to $\mathbf{D}^q$, with
\begin{align*}
\bar{d}_{pp}^q &\equiv \left\{
 \begin{array}{ll}
   \frac{\big( \mathbf{B}^{(k+1)T}\mathbf{B}^{(k+1)}\mathbf{\bar{S}}^k\mathbf{C}^{(k+1)}\mathbf{C}^{(k+1)T} \big)_{pq}}{\bar{s}_{pq}^k} & \mathrm{if}\;\; p\in \mathcal{I}_q \\
   \star & \mathrm{if}\;\; p\notin \mathcal{I}_q,
 \end{array} \right.
\text{and}\;
\hat{d}_{pp}^q &\equiv \left\{
 \begin{array}{ll}
   \frac{1}{\bar{s}_{pq}^k} & \mathrm{if}\;\; p\in \mathcal{I}_q \\
   \star & \mathrm{if}\;\; p\notin \mathcal{I}_q.
 \end{array} \right.
\end{align*}
Accordingly,
\begin{align}
\mathfrak{G}\big(\mathfrak{S},\mathfrak{S}^k\big)-\mathfrak{J}\big(\mathfrak{S}\big) = &\frac{1}{2} \sum_{q=1}^Q \left\{\sum_{p=1}^P v_{pq}^2 \bar{d}_{pp}^q + \delta_{\mathbf{S}}^k \sum_{p=1}^P v_{pq}^2 \hat{d}_{pp}^q \right\} \nonumber \\
&- \frac{1}{2} \sum_{q=1}^Q \mathbf{v}_q^T \big( \mathbf{B}^{(k+1)T}\mathbf{B}^{(k+1)}\mathbf{C}^{(k+1)}\mathbf{C}^{(k+1)T} \big)\mathbf{v}_q. \label{127}
\end{align}
As shown, with the boundedness of $\mathbf{B}^k$, $\mathbf{C}^k$, and $\mathbf{S}^k$, and by sufficiently large $\delta_{\mathbf{S}}^k$, $\mathfrak{G}\big(\mathfrak{S},\mathfrak{S}\big) \le \mathfrak{G}\big(\mathfrak{S},\mathfrak{S}^k\big)$ can be guaranteed. Next we prove that if and only if $\mathbf{S}^k$ satisfies the KKT conditions, then the equality holds. 

If $\mathbf{S}^k$ satisfies the KKT conditions, then this is obvious by eq.~\ref{126} regardless of $\delta_{\mathbf{S}}^k$. And by eq.~\ref{127}, since $\delta_{\mathbf{S}}^k$ is a variable, the equality happens if and only if $\mathbf{S} = \mathbf{S}^k$ which by the update rule in eq.~\ref{eq100} happens if and only if $\mathbf{S}^k$ satisfies the KKT conditions. This completes the proof.
\end{proof}

\begin{theorem} \label{theorem28}
$\mathfrak{G}\big(\mathfrak{S},\mathfrak{S}^k\big) \le \mathfrak{G}\big(\mathfrak{S}^k,\mathfrak{S}^k\big)$. Moreover if and only if $\mathbf{S}^k$ satisfies the KKT conditions in eq.~\ref{eq91}, then $\mathfrak{G}\big(\mathfrak{S},\mathfrak{S}^k\big) = \mathfrak{G}\big(\mathfrak{S}^k,\mathfrak{S}^k\big)$.
\end{theorem}
\begin{proof}
\begin{equation*}
\mathfrak{G}\big(\mathfrak{S}^k,\mathfrak{S}^k\big) - \mathfrak{G}\big(\mathfrak{S},\mathfrak{S}^k\big) = -\mathrm{tr}\;\big\{\big(\mathfrak{S}-\mathfrak{S}^k\big)^T\nabla_{\mathfrak{S}}\mathfrak{J}\big(\mathfrak{S}^{kT}\big)\big\} - \frac{1}{2}\mathrm{tr}\;\big\{\big(\mathfrak{S}-\mathfrak{S}^k\big)^T\mathbf{D}\big(\mathfrak{S}-\mathfrak{S}^k\big)\big\}.
\end{equation*}
By using eq.~\ref{124} and the fact that $\mathbf{D}$ is positive semi-definite:
\begin{equation*}
\mathfrak{G}\big(\mathfrak{S}^k,\mathfrak{S}^k\big) - \mathfrak{G}\big(\mathfrak{S},\mathfrak{S}^k\big) = \frac{1}{2}\mathrm{tr}\;\big\{\big(\mathfrak{S}-\mathfrak{S}^k\big)^T\mathbf{D}\big(\mathfrak{S}-\mathfrak{S}^k\big)\big\} \ge 0, 
\end{equation*}
By the update rule eq.~\ref{eq100}, if $\mathbf{S}^k$ satisfies the KKT conditions, then $\mathbf{S}=\mathbf{S}^k$, and therefore the equality holds. Now we need to prove that if the equality holds, then $\mathbf{S}^k$ satisfies the KKT conditions.

To prove this, let consider a contradiction situation where the equality holds but $\mathbf{S}^k$ does not satisfy the KKT conditions. In this case, there exists at least an index $(p,q)$ such that:
\begin{equation*}
s_{pq}\ne s_{pq}^k\;\;\mathrm{and}\;\; d_{pp}^q = \frac{ \big( \mathbf{B}^{(k+1)T}\mathbf{B}^{(k+1)}\mathbf{\bar{S}}^k\mathbf{C}^{(k+1)}\mathbf{C}^{(k+1)T} \big)_{pq}+\delta_{\mathbf{S}}^k}{\bar{s}_{pq}^k} \ge \frac{\delta_{\mathbf{S}}^k}{\bar{s}_{pq}^k}.
\end{equation*}
Note that by the definition in eq.~\ref{eq103}, if $\bar{s}_{pq}^k$ is equal to zero, then $s_{pq}= s_{pq}^k$ which violates the condition for the contradiction, so $\bar{s}_{pq}^k$ cannot be equal to zero. Consequently,
\begin{equation*}
\mathfrak{G}\big(\mathfrak{S}^k,\mathfrak{S}^k\big)-\mathfrak{G}\big(\mathfrak{S},\mathfrak{S}^k\big) \ge \frac{\big(s_{pq}-s_{pq}^k\big)^2\delta_{\mathbf{S}}^k}{\bar{s}_{pq}^k} > 0,
\end{equation*}
which violates the equality. Thus, it is proven that if the equality holds, then $\mathbf{S}^k$ satisfies the KKT conditions.
\end{proof}

\begin{theorem}\label{theorem29}
Given sufficiently large $\delta_{\mathbf{S}}^k$ and the boundedness of $\mathbf{B}^k$, $\mathbf{C}^k$, and $\mathbf{S}^k$, $J\big(\mathbf{S}^{k+1}\big)$ $\le$ $J\big(\mathbf{S}^k\big)\;\,\forall k\ge 0$ under update rule eq.~\ref{eq100} with the equality happens if and only if $\mathbf{S}^k$ satisfies the KKT conditions in eq.~\ref{eq91}. 
\end{theorem}
\begin{proof}
This theorem is the corollary of theorem \ref{theorem26}, \ref{theorem27}, and \ref{theorem28}
\end{proof}

\paragraph{D.~The convergence guarantee of algorithm \ref{algorithm8}}

To shown the convergence of algorithm \ref{algorithm8}, the following statements must be proven:
\begin{enumerate}
\item the nonincreasing property of sequence $J\big(\mathbf{B}^k$, $\mathbf{S}^k$, $\mathbf{C}^k\big)$, i.e., $J\big(\mathbf{B}^{(k+1)}$, $\mathbf{S}^{(k+1)}$, $\mathbf{C}^{(k+1)}\big)$ $\le$ $J\big(\mathbf{B}^{(k+1)}$, $\mathbf{S}^k$, $\mathbf{C}^{(k+1)}\big)$ $\le$ $J\big(\mathbf{B}^{(k+1)}$, $\mathbf{S}^k$, $\mathbf{C}^k\big)$ $\le$ $J\big(\mathbf{B}^k$, $\mathbf{S}^k$, $\mathbf{C}^k\big)$,
\item any limit point of the sequence $\big\{\mathbf{B}^k$, $\mathbf{S}^k$, $\mathbf{C}^k\big\}$ generated by algorithm \ref{algorithm8} is a stationary point, and
\item the sequence $\big\{\mathbf{B}^k$, $\mathbf{S}^k$, $\mathbf{C}^k\big\}$ has at least one limit point.
\end{enumerate}
The first will be proven in theorem \ref{theorem30}, the second in theorem \ref{theorem31}, and the third in theorem \ref{theorem32}.

\begin{theorem} \label{theorem30}
Given sufficiently large $\delta_{\mathbf{B}}^k$, $\delta_{\mathbf{C}}^k$, and $\delta_{\mathbf{S}}^k$, and the boundedness of $\mathbf{B}^k$, $\mathbf{C}^k$, and $\mathbf{S}^k$, $J\big(\mathbf{B}^{(k+1)}$, $\mathbf{S}^{(k+1)}$, $\mathbf{C}^{(k+1)}\big)$ $\le$ $J\big(\mathbf{B}^{(k+1)}$, $\mathbf{S}^k$, $\mathbf{C}^{(k+1)}\big)$ $\le$ $J\big(\mathbf{B}^{(k+1)}$, $\mathbf{S}^k$, $\mathbf{C}^k\big)$ $\le$ $J\big(\mathbf{B}^k$, $\mathbf{S}^k$, $\mathbf{C}^k\big)$ under update rules in algorithm \ref{algorithm8} with the equalities happen if and only if $\big(\mathbf{B}^k$, $\mathbf{S}^k$, $\mathbf{C}^k\big)$ is a stationary point.
\end{theorem}
\begin{proof}
$J\big(\mathbf{B}^{(k+1)}$, $\mathbf{S}^k$, $\mathbf{C}^k\big)$ $\le$ $J\big(\mathbf{B}^k$, $\mathbf{S}^k$, $\mathbf{C}^k\big)$ is due to theorem \ref{theorem21} with the equality happens if and only if $\mathbf{B}^k$ satisfies the KKT conditions. $J\big(\mathbf{B}^{(k+1)}$, $\mathbf{S}^k$, $\mathbf{C}^{(k+1)}\big)$ $\le$ $J\big(\mathbf{B}^{(k+1)}$, $\mathbf{S}^k$, $\mathbf{C}^k\big)$ is due to theorem \ref{theorem25} with the equality happens if and only if $\mathbf{C}^k$ satisfies the KKT conditions. And $J\big(\mathbf{B}^{(k+1)}$, $\mathbf{S}^{(k+1)}$, $\mathbf{C}^{(k+1)}\big)$ $\le$ $J\big(\mathbf{B}^{(k+1)}$, $\mathbf{S}^k$, $\mathbf{C}^{(k+1)}\big)$ is due to theorem \ref{theorem29} with the equality happens if and only if $\mathbf{S}^k$ satisfies the KKT conditions. And by combining theorem \ref{theorem21}, \ref{theorem25}, and \ref{theorem29}, algorithm \ref{algorithm8} will stop updating sequence $J\big(\mathbf{B}^{k}$, $\mathbf{S}^{k}$, $\mathbf{C}^{k}\big)$ if and only if $\mathbf{B}^k$, $\mathbf{C}^k$, and $\mathbf{S}^k$ satisfy the KKT conditions, i.e., $\big(\mathbf{B}^k$, $\mathbf{S}^k$, $\mathbf{C}^k\big)$ is a stationary point.
\end{proof}

\begin{theorem} \label{theorem31}
Given sufficiently large $\delta_{\mathbf{B}}^k$, $\delta_{\mathbf{C}}^k$, and $\delta_{\mathbf{S}}^k$, and the boundedness of $\mathbf{B}^k$, $\mathbf{C}^k$, and $\mathbf{S}^k$, it can be shown that any limit point of sequence $\big\{\mathbf{B}^k,\mathbf{S}^k,\mathbf{C}^k\big\}$ generated by algorithm \ref{algorithm8} is a stationary point.
\end{theorem}
\begin{proof}
By theorem \ref{theorem30}, algorithm \ref{algorithm8} produces strictly decreasing sequence $J\big(\mathbf{B}^k$, $\mathbf{S}^k$, $\mathbf{C}^k\big)$ until reaching a point that satisfies the KKT conditions. Because $J\big(\mathbf{B}^k$, $\mathbf{S}^k$, $\mathbf{C}^k\big)$ $\ge$ $0$, this sequence is bounded and thus converges. And by combining the results of the theorem \ref{theorem21}, \ref{theorem25}, \ref{theorem29}, algorithm \ref{algorithm8} will stop updating $J\big(\mathbf{B}^k$, $\mathbf{S}^k$, $\mathbf{C}^k\big)$ if and only if $\big(\mathbf{B}^k$, $\mathbf{S}^k$, $\mathbf{C}^k\big)$ satisfies the KKT conditions. And by the update rules in algorithm \ref{algorithm8}, after a point satisfies the KKT conditions, the algorithm will stop updating $\big(\mathbf{B}^k$, $\mathbf{S}^k$, $\mathbf{C}^k\big)$, i.e., $\mathbf{B}^{(k+1)}$ $=$ $\mathbf{B}^k$, $\mathbf{C}^{(k+1)}$ $=$ $\mathbf{C}^k$, and $\mathbf{S}^{(k+1)}$ $=$ $\mathbf{S}^k\;\,\forall k\ge *$. This completes the proof.
\end{proof}

\begin{theorem} \label{theorem32}
The sequence $\big\{\mathbf{B}^k,\mathbf{S}^k,\mathbf{C}^k\big\}$ has at least one limit point.
\end{theorem}
\begin{proof}
It suffices to prove that sequence $\big\{\mathbf{B}^k,\mathbf{S}^k,\mathbf{C}^k\big\}$ is in a closed and bounded set. The boundedness of $\big\{\mathbf{B}^k\big\}$ and $\big\{\mathbf{C}^k\big\}$ are clear by the objective in eq.~\ref{eq90}; if there exists $l$ such that $\lim b_{mp}^l\to \infty$ or $\lim c_{qn}^l\to \infty$, then $\lim J \to \infty > J(\mathbf{B}^0,\mathbf{S}^0,\mathbf{C}^0)$ which violates theorem \ref{theorem30}. And if $\big\{\mathbf{S}^k\big\}$ is not bounded, then there exists $l$ such that $\lim s_{pq}^l\to \infty$, $s_{pq}^l < s_{pq}^{(l+1)}$. Because due to theorem \ref{theorem30}, $J(\mathbf{B}^k,\mathbf{S}^k,\mathbf{C}^k)$ is bounded, then either $b_{mp}^l$ $\forall m$ or $c_{qn}^l$ $\forall n$ must be equal to zero. If $b_{mp}^l=0\;\,\forall m$, then $\nabla_{\mathbf{S}}J_{pq}=0\;\,\forall q$, so that $s_{pq}^{(l+1)} = s_{pq}^l$. And if $c_{qn}^l=0\;\,\forall n$, then $\nabla_{\mathbf{S}}J_{pq}=0\;\,\forall p$, so that $s_{pq}^{(l+1)} = s_{pq}^l$. Both cases contradict the condition for unboundedness of $\mathbf{S}^l$. Thus, $\mathbf{S}^l$ is also bounded.

With the nonnegativity guarantee from theorem \ref{theorem17}, it is proven that $\big\{\mathbf{B}^k$, $\mathbf{S}^k$, $\mathbf{C}^k\big\}$ is in a closed and bounded set.
\end{proof}

Algorithm \ref{algorithm9} shows modifications to algorithm \ref{algorithm8} in order to guarantee the convergence as suggested by theorem \ref{theorem30}, \ref{theorem31}, and \ref{theorem32} with step is a constant that determine how fast $\delta_{\mathbf{B}}^k$, $\delta_{\mathbf{C}}^k$, and $\delta_{\mathbf{S}}^k$ grow in order to satisfies the nonincreasing property. Note that we set the same step value for all sequences, but setting different values can also be employed.

\begin{algorithm}
\caption{Converged algorithm for BNMF problem}
\label{algorithm9}
\begin{algorithmic}
\STATE Initialization, $\mathbf{B}^0\ge\mathbf{0}$, $\mathbf{C}^0\ge\mathbf{0}$, and $\mathbf{S}^0\ge\mathbf{0}$.
\FOR {$k=0,\ldots,K$}
\STATE $ $
\STATE $\delta_{\mathbf{B}}^k \longleftarrow \delta$
\REPEAT
\STATE \begin{align*} b_{mp}^{(k+1)} \longleftarrow & \;b_{mp}^{k} - \frac{\bar{b}_{mp}^{k}\times\nabla_{\mathbf{B}}J(\mathbf{B}^k,\mathbf{S}^k,\mathbf{C}^k)_{mp}}{\big(\mathbf{\bar{B}}^{k}\mathbf{S}^{k}\mathbf{C}^k\mathbf{C}^{kT}\mathbf{S}^{kT}+\beta\mathbf{\bar{B}}^k\mathbf{\bar{B}}^{kT}\mathbf{\bar{B}}^k\big)_{mp}+\delta_{\mathbf{B}}^k}\;\;\forall m,p 
\\ 
\delta_{\mathbf{B}}^k \longleftarrow & \;\delta_{\mathbf{B}}^k\times \mathrm{step} \end{align*}
\UNTIL {$J\big( \mathbf{B}^{(k+1)},\mathbf{S}^k, \mathbf{C}^k \big) \le J\big(\mathbf{B}^k,\mathbf{S}^k, \mathbf{C}^k \big)$}
\STATE $ $
\STATE $\delta_{\mathbf{C}}^k \longleftarrow \delta$
\REPEAT
\STATE \begin{align*} c_{qn}^{(k+1)} \longleftarrow & \;c_{qn}^{k} - \frac{\bar{c}_{qn}^{k}\times\nabla_{\mathbf{C}}J(\mathbf{B}^{k+1},\mathbf{S}^k,\mathbf{C}^k)_{qn}}{\big(\mathbf{S}^{kT}\mathbf{B}^{(k+1)T}\mathbf{B}^{(k+1)}\mathbf{S}^k\mathbf{\bar{C}}^k+\alpha\mathbf{\bar{C}}^k\mathbf{\bar{C}}^{kT}\mathbf{\bar{C}}^k\big)_{qn}+\delta_{\mathbf{C}}^k} \;\;\forall q,n 
\\ 
\delta_{\mathbf{C}}^k \longleftarrow & \;\delta_{\mathbf{C}}^k\times \mathrm{step} \end{align*}
\UNTIL {$J\big( \mathbf{B}^{(k+1)},\mathbf{S}^k, \mathbf{C}^{(k+1)} \big) \le J\big(\mathbf{B}^{(k+1)},\mathbf{S}^k, \mathbf{C}^k \big)$}
\STATE $ $
\STATE $\delta_{\mathbf{S}}^k \longleftarrow \delta$
\REPEAT
\STATE \begin{align*} s_{pq}^{(k+1)} \longleftarrow & \;s_{pq}^{k} - \frac{\bar{s}_{pq}^{k}\times\nabla_{\mathbf{S}}J(\mathbf{B}^{k+1},\mathbf{S}^k,\mathbf{C}^{(k+1)})_{pq}}{\big(\mathbf{B}^{(k+1)T}\mathbf{B}^{(k+1)}\mathbf{\bar{S}}^{k}\mathbf{C}^{(k+1)}\mathbf{C}^{(k+1)T}\big)_{pq}+\delta_{\mathbf{S}}^k} \;\;\forall p,q \\ \delta_{\mathbf{S}}^k \longleftarrow & \;\delta_{\mathbf{S}}^k\times \mathrm{step} \end{align*}
\UNTIL {$J\big( \mathbf{B}^{(k+1)},\mathbf{S}^{(k+1)}, \mathbf{C}^{(k+1)} \big) \le J\big(\mathbf{B}^{(k+1)},\mathbf{S}^k, \mathbf{C}^{(k+1)} \big)$}
\ENDFOR
\end{algorithmic}
\end{algorithm}

\section{Experimental Results} \label{results}

Experiments are conducted to analyze and compare properties and performances of algorithm \ref{algorithm1} (LS), algorithm \ref{algorithm2} (D-U), algorithm \ref{algorithm3} (D-B), algorithm \ref{algorithm4} (MU-U), algorithm \ref{algorithm6} (AU-U), algorithm \ref{algorithm7} (MU-B), and algorithm \ref{algorithm9} (AU-B). Here, LS is used as the benchmark. All algorithms are developed in Octave under linux platform, and the experiments are conducted by using a notebook with 1.86 GHz Intel processor and 2 GB RAM. 

\subsection{The datasets}\label{datasets}

To evaluate the algorithms, we use the Reuters-21578 data corpus\footnote{http://kdd.ics.uci.edu/databases/reuters21578/reuters21578.html}, a standard dataset for testing learning algorithms and other text-based processing methods. The dataset is especially interesting because many NMF-based clustering methods are tested using it, e.g., \cite{Shahnaz,Ding1,Xu}. The Reuters-21578 contains 21578 documents with 135 topics class created manually with each document is assigned to one or more topics based on its content. The Reuters-21578 are available in two formats: SGML and XML version. The dataset is divided into 22 files with each file contains 1000 documents and the last file contains 578 documents. 

In this experiments, we use the XML version. We use all but the 18$^{\text{th}}$ file because this file is invalid both in its SGML and XML version. We use only documents that belong to exclusively one class (we use ``classes'' for refeering the original grouping, and ``clusters'' for referring groups resulted from the clustering algorithms). Further, we remove the common English stop words\protect\footnote{http://snowball.tartarus.org/algorithms/english/stop.txt}, and then stem the remaining words by using Porter stemmer \cite{Rijsbergen} and remove words that belong to only one documents. And also, we normalize the term-by document matrix $\mathbf{A}$ by: $\mathbf{A} \leftarrow \mathbf{AD}^{-1/2}$ where $\mathbf{D}=\text{diag}\big( \mathbf{A}^T\mathbf{A}\mathbf{e}\big)$ as suggested by Xu et al.~\cite{Xu}. We form test datasets by combining top 2, 4, 6, 8, 10, and 12 classes from the corpus. Table \ref{ch2:table3} summarizes the statistics of these test datasets, where \#doc, \#word, \%nnz, max, and min refer to number of document, number of word, percentage of nonzero entry, maximum cluster size, and minimum cluster size respectively. And table \ref{ch2:table4} gives sizes (\#doc) of these top 12 classes.

\begin{table}
  \begin{center}
    \caption{Statistics of the test datasets}
    \centering
    \begin{tabular}{|l|r|r|r|r|r|}
    \hline
    The data &\#doc & \#word & \%nnz & max & min \\
    \hline
    Reuters2    & 6090 & 8547  & 0.363 & 3874 & 2216 \\
    Reuters4    & 6797 & 9900  & 0.353 & 3874 & 333 \\
    Reuters6    & 7354 & 10319 & 0.347 & 3874 & 269 \\
    Reuters8    & 7644 & 10596 & 0.340 & 3874 & 144 \\
    Reuters10   & 7887 & 10930 & 0.336 & 3874 & 114 \\
    Reuters12   & 8052 & 11172 & 0.333 & 3874 &  75 \\
    \hline
    \end{tabular}
    \label{ch2:table3}
  \end{center}
\end{table}

\begin{table}
  \begin{center}
    \caption{Sizes of the top 12 topics}
    \centering
    \begin{tabular}{|r|r|r|r|r|r|r|r|r|r|r|r|}
    \hline
    class &    1 &    2 &   3 &   4 &   5 &   6 \\ 
    \#doc & 3874 & 2216 & 374 & 333 & 288 & 269 \\ \hline
    class &   7 &   8 &   9 &  10 & 11 & 12 \\ 
    \#doc & 146 & 144 & 129 & 114 & 90 & 75 \\ \hline
    \end{tabular}
    \label{ch2:table4}
  \end{center}
\end{table}

\subsection{The nonincreasing property} 
\label{nonincreasing}

The nonincreasing property, even though does not guarantee the convergence, is still a very important property since usually good results can be achieved by having this property. Moreover, unlike the stationarity, this property is easy to evaluate. Here we will show that while MU-U and MU-B---which do not have convergence guarantee---fail to show this property for large $\alpha$ and/or $\beta$, AU-U and AU-B---which have convergence guarantee---can consistently achieve the desired results even for large $\alpha$ and/or $\beta$. Note that, even though LS \cite{Lee2} has this property, it doesn't imply that other MU based algorithms will inherit it. As shown in figure \ref{fig1} and \ref{fig2}, the original orthogonal NMF algorithms (D-U and D-B) which based on the MU rules do not have this property.

\begin{figure}
 \begin{center}
  \subfigure[Small $\alpha$]{
   \includegraphics[width=0.45\textwidth]{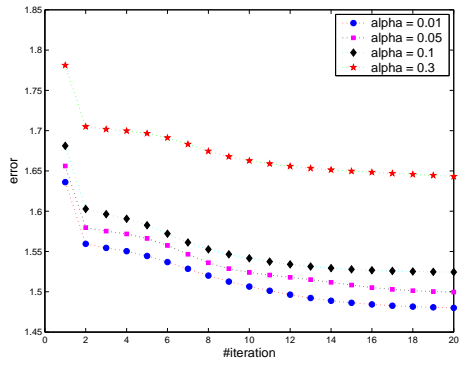}
   \label{fig3a}
  }
  \subfigure[Medium to large $\alpha$]{
   \includegraphics[width=0.45\textwidth]{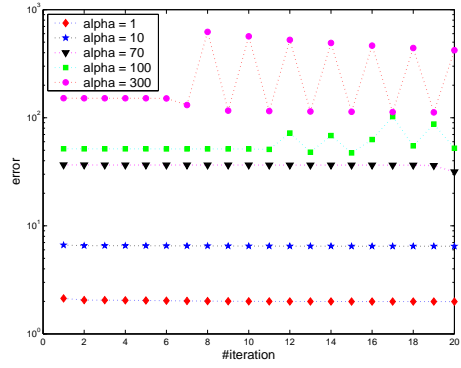}
   \label{fig3b}
  }
\\
  \subfigure[Some values of $\alpha$]{
   \includegraphics[width=0.7\textwidth]{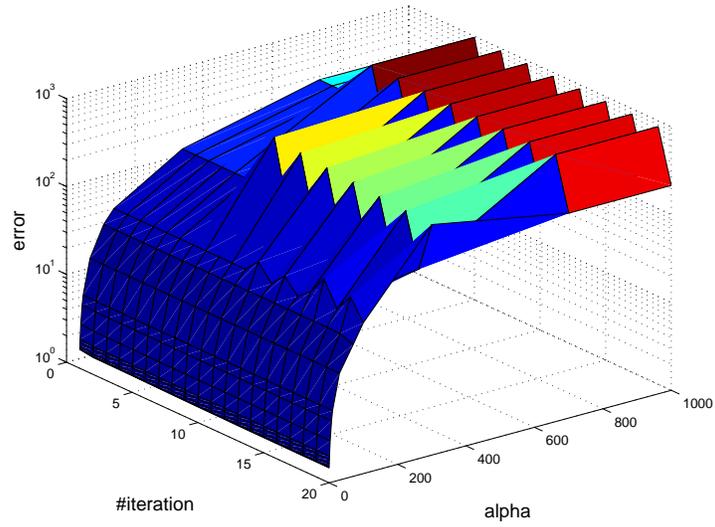}
   \label{fig3c}
  }
  \caption{MU-U error per iteration for Reuters4 dataset}
  \label{fig3}
 \end{center}
\end{figure}

\begin{figure}
 \begin{center}
  \subfigure[Small $\alpha$]{
   \includegraphics[width=0.45\textwidth]{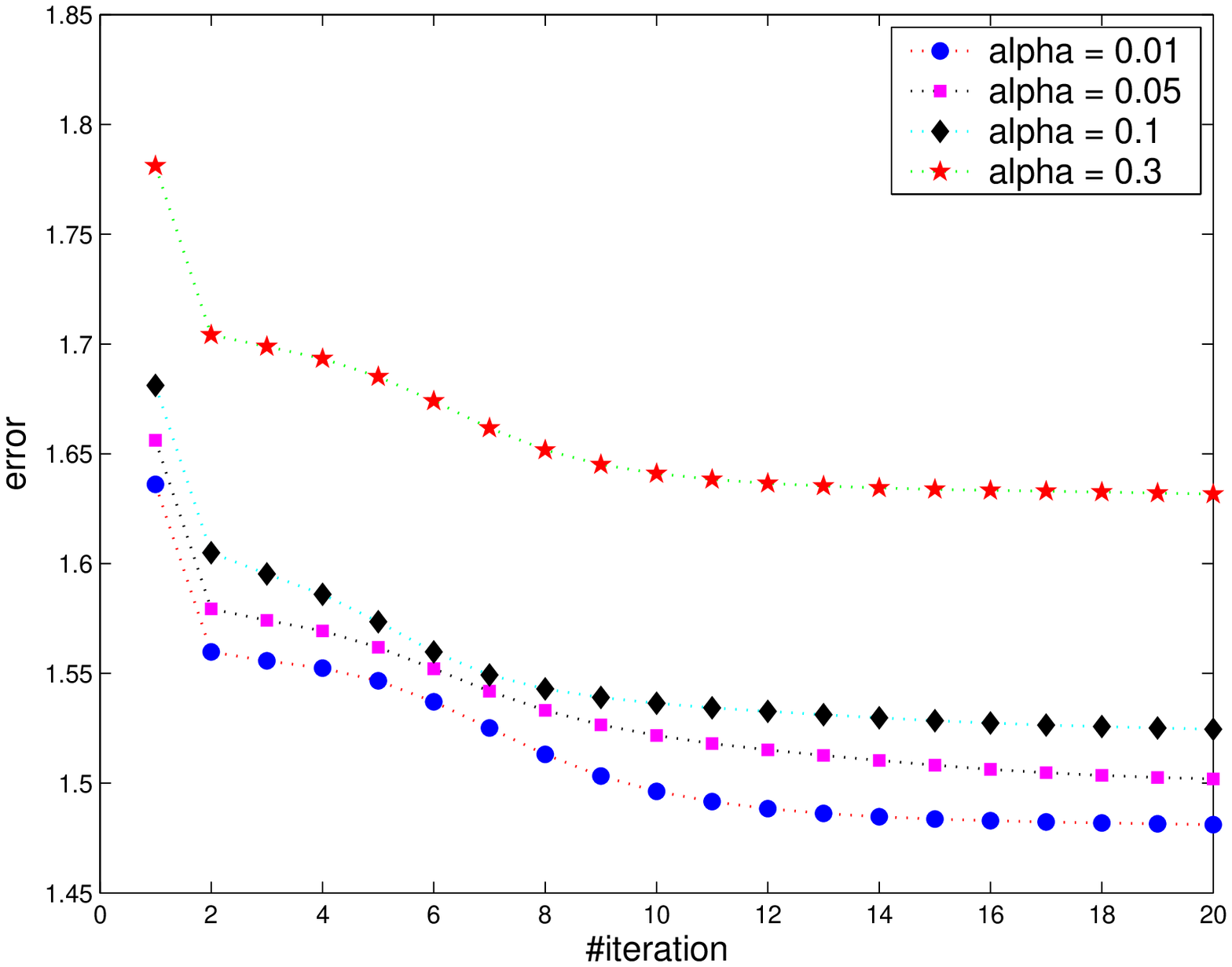}
   \label{fig4a}
  }
  \subfigure[Medium to large $\alpha$]{
   \includegraphics[width=0.45\textwidth]{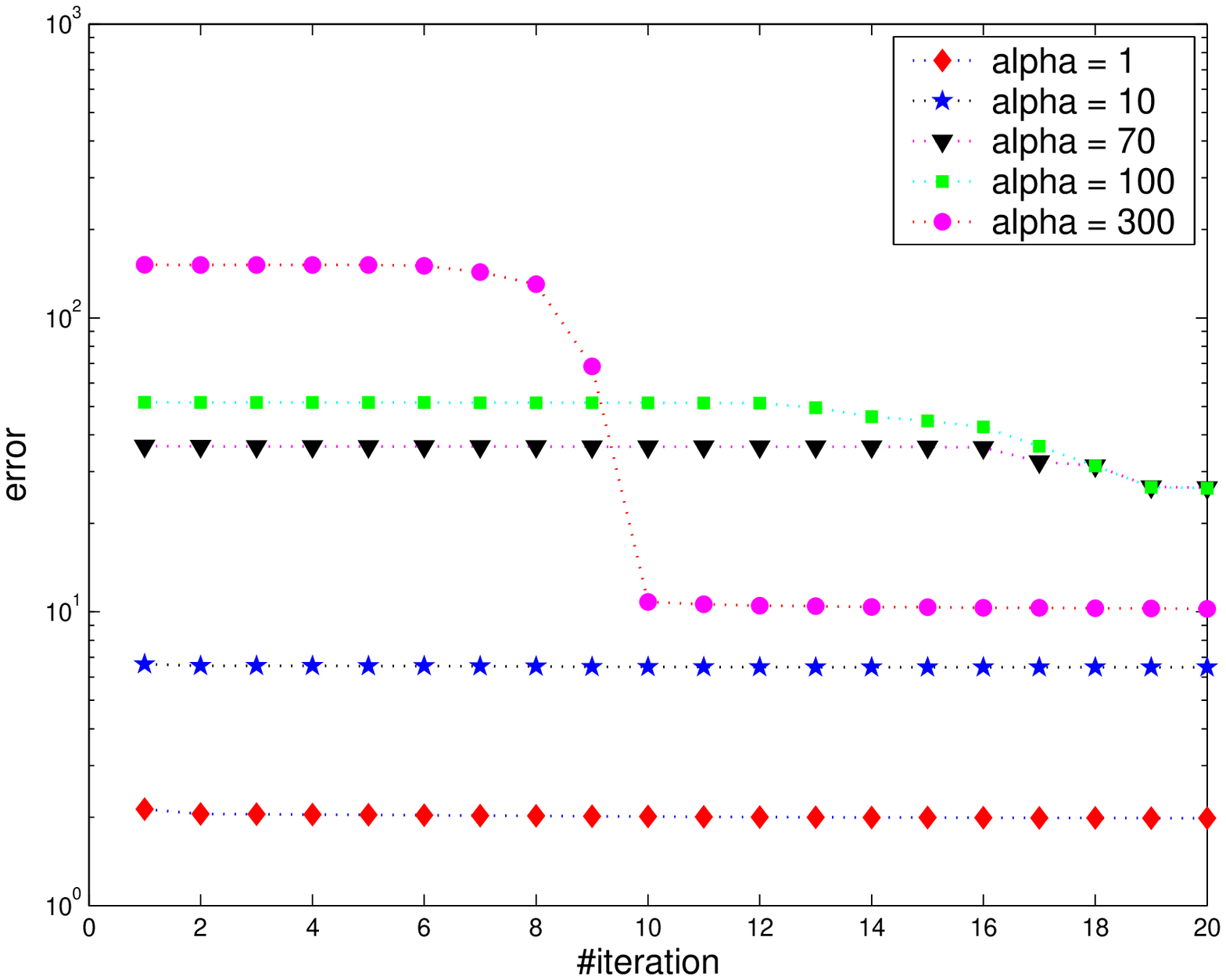}
   \label{fig4b}
  }
\\
  \subfigure[Some values of $\alpha$]{
   \includegraphics[width=0.7\textwidth]{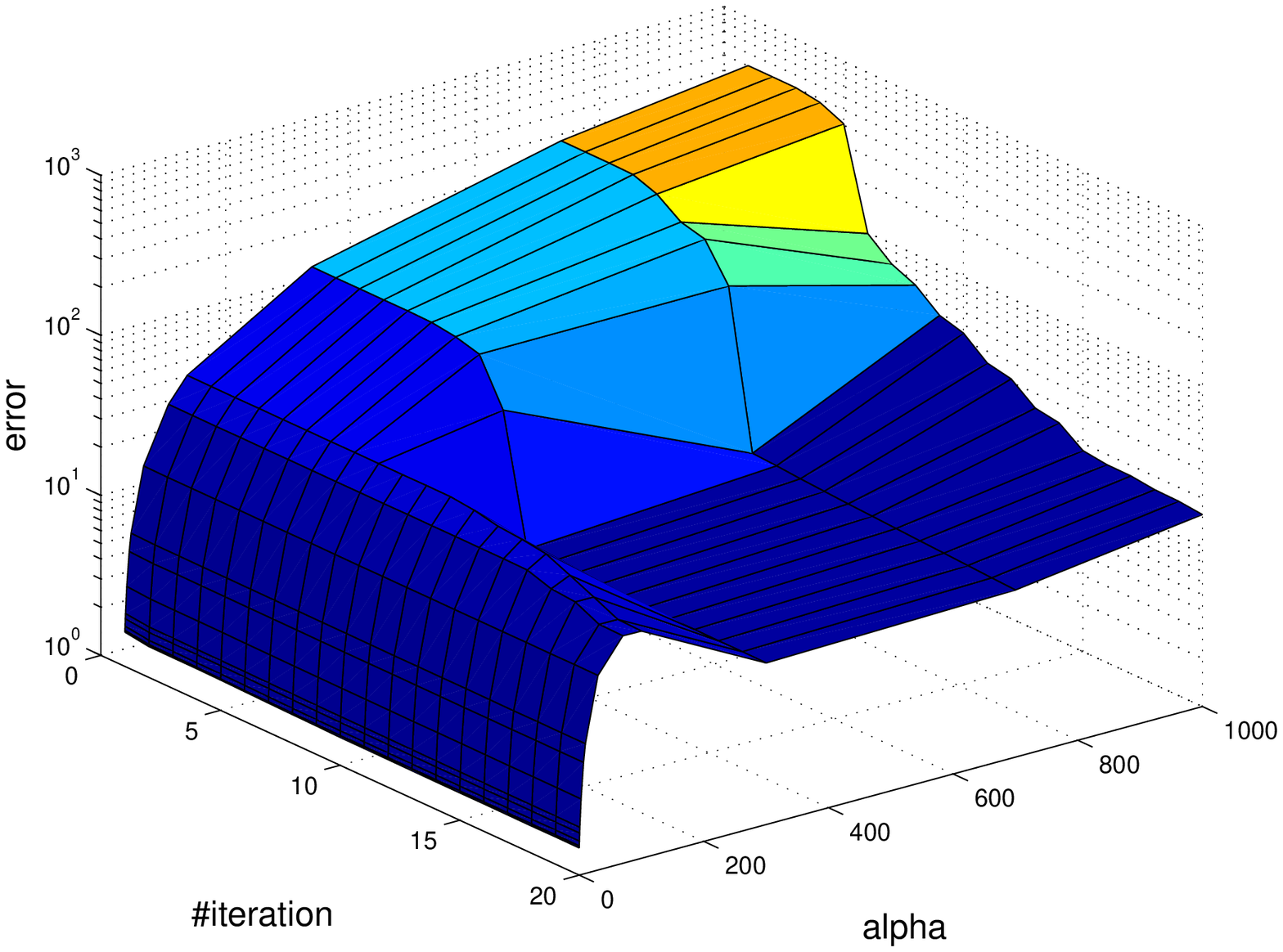}
   \label{fig4c}
  }
  \caption{AU-U error per iteration for Reuters4 dataset}
  \label{fig4}
 \end{center}
\end{figure}

Figure \ref{fig3} show error per iteration produced by MU-U as a function of $\alpha$. As the error, we use the UNMF objective in eq.~\ref{eq52}. As shown, the nonincreasing property vanishes as $\alpha$ grows. And not only the errors are rather large, but also the algorithm seems to fail to settle for large $\alpha$ values. On the other hand, as shown in figure \ref{fig4}, AU-U preserves the nonincreasing property even for large $\alpha$ values (AU-U uses the same error as MU-U). Interestingly, as shown in figure \ref{fig4b}, the errors for $\alpha=300$ are even smaller than the errors for $\alpha=100$ and $\alpha=70$. And since $\frac{\alpha}{2}\big\|\mathbf{CC}^T-\mathbf{I}\big\|_F^2$ is part of the objective in eq.~\ref{eq52}, the small errors for large $\alpha$ values in AU-U indicate that $\mathbf{C}$s produced by AU-U are much more row-orthogonal than those produced by MU-U.

\begin{figure}
 \begin{center}
  \subfigure[Small $\alpha$]{
   \includegraphics[width=0.45\textwidth]{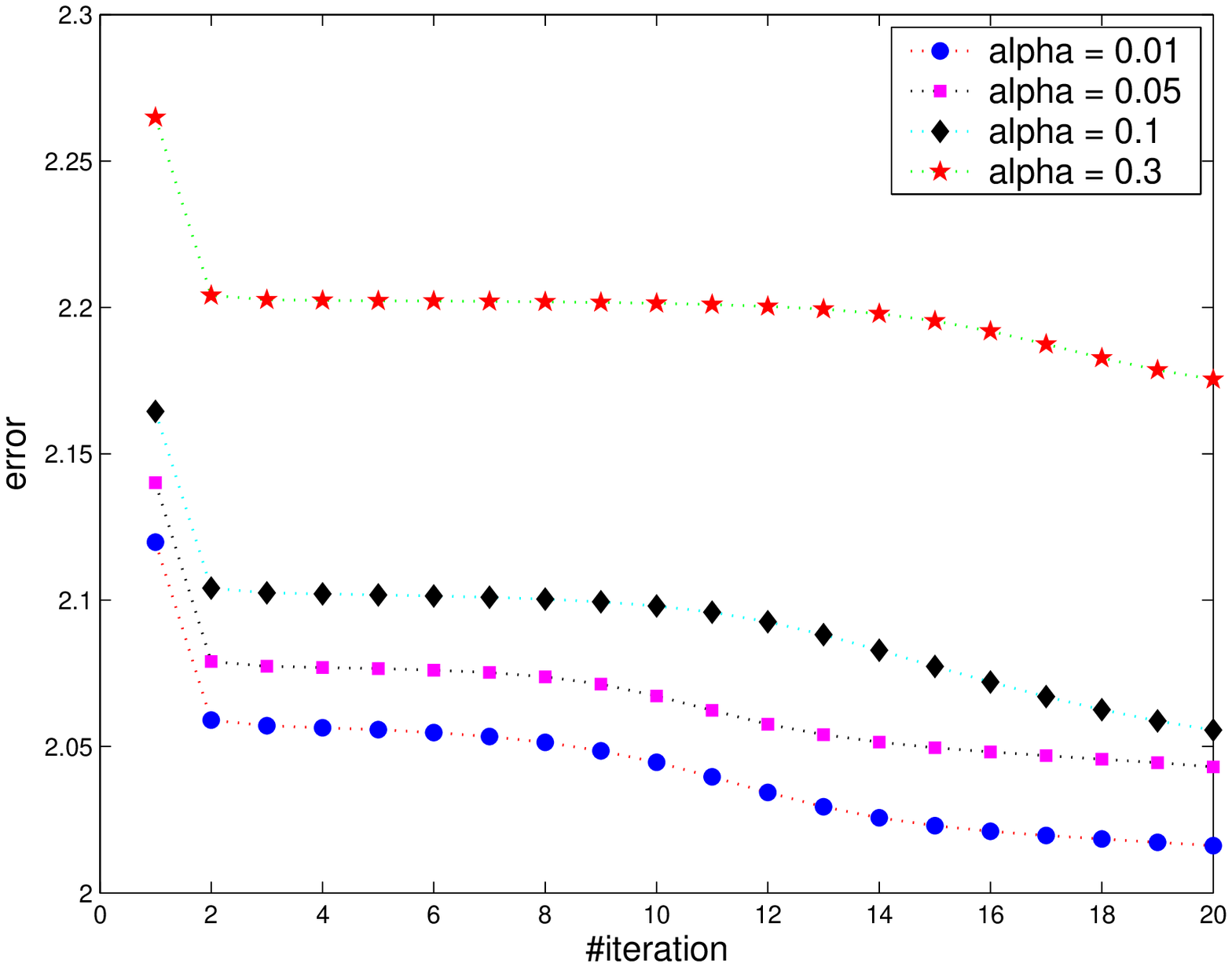}
   \label{fig5a}
  }
  \subfigure[Medium to large $\alpha$]{
   \includegraphics[width=0.45\textwidth]{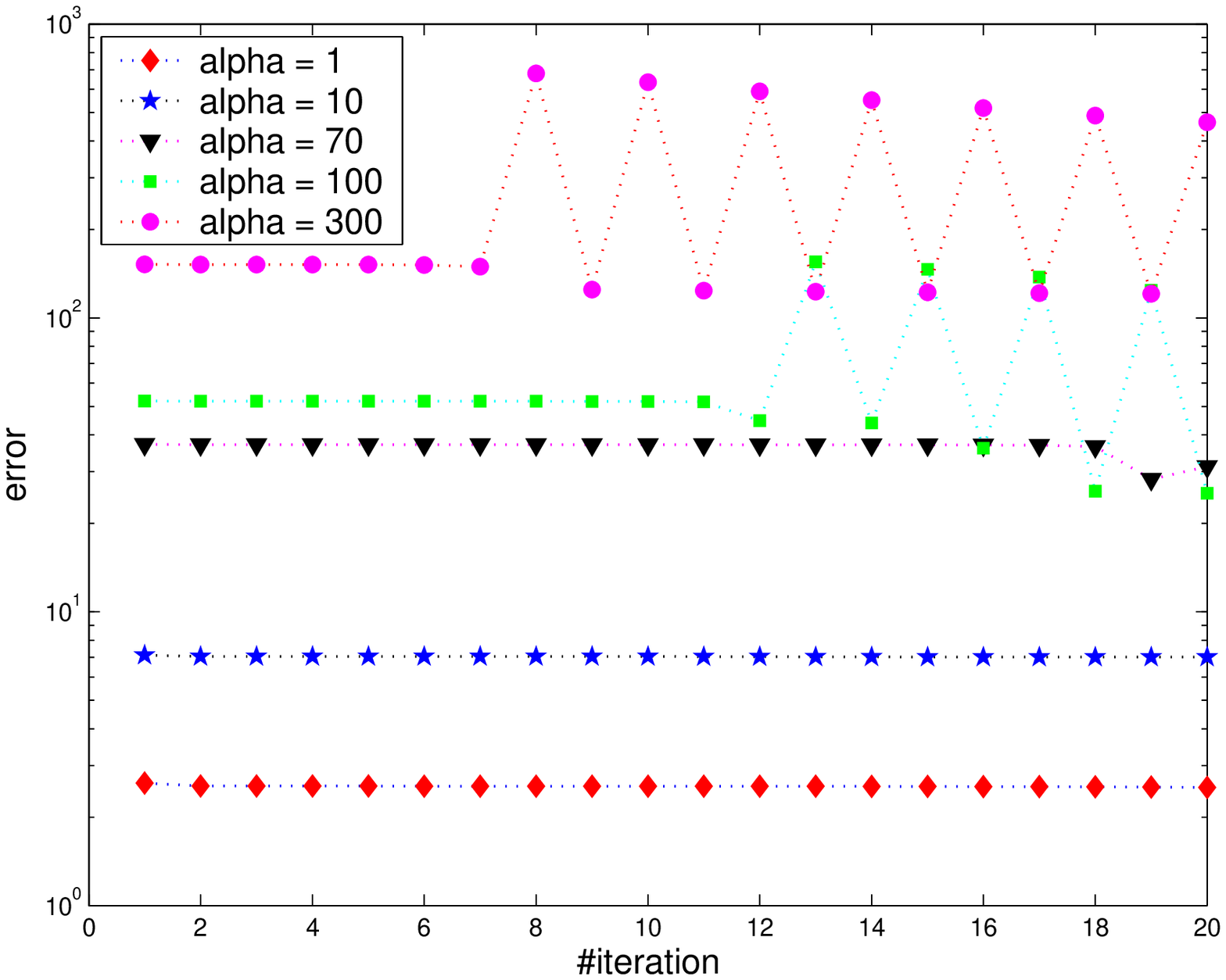}
   \label{fig5b}
  }
\\
  \subfigure[Some values of $\alpha$]{
   \includegraphics[width=0.7\textwidth]{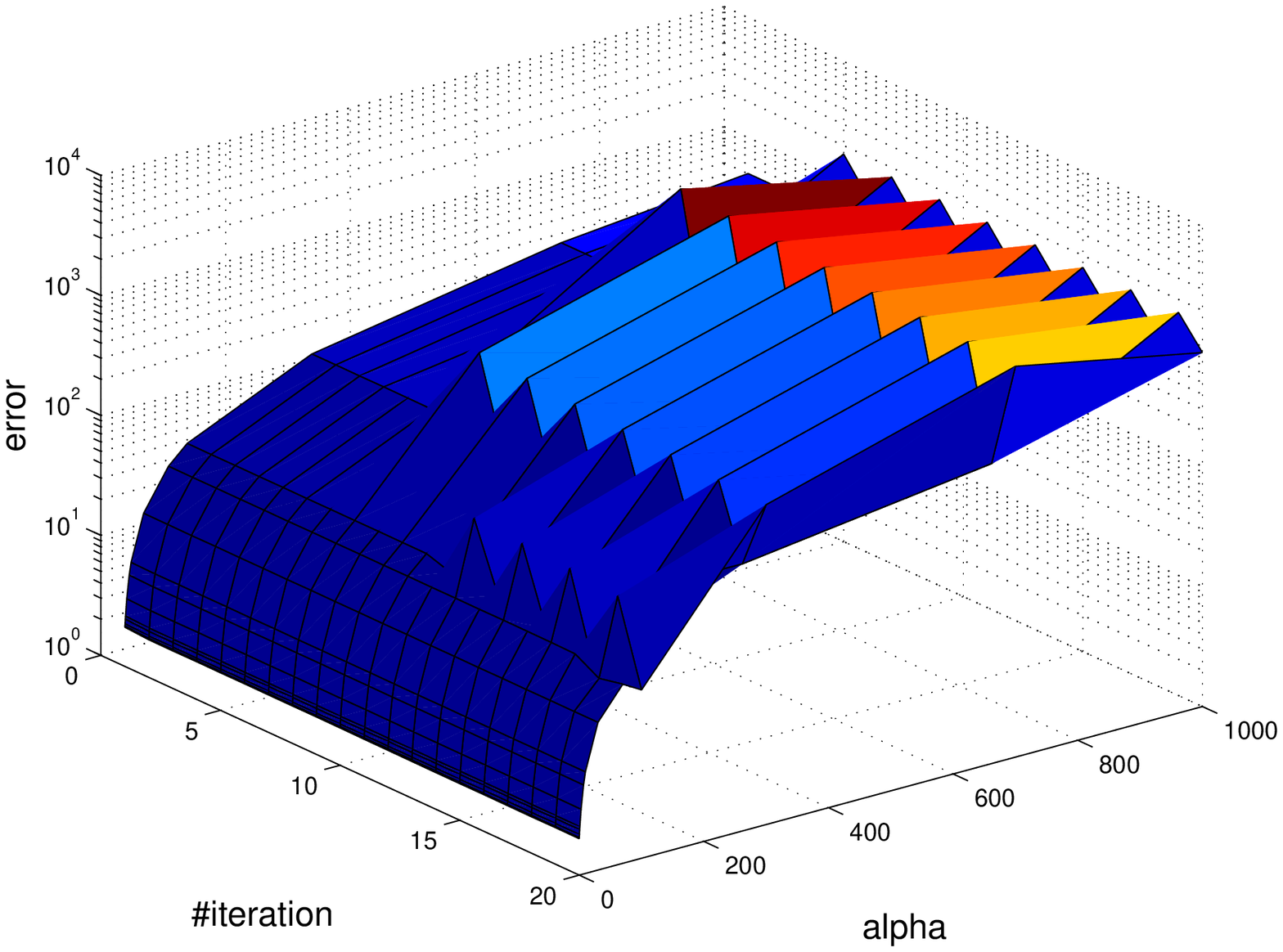}
   \label{fig5c}
  }
  \caption{MU-B($\alpha$) error per iteration for Reuters4 dataset ($\beta=1$).}
  \label{fig5}
 \end{center}
\end{figure}

\begin{figure}
 \begin{center}
  \subfigure[Small $\alpha$]{
   \includegraphics[width=0.45\textwidth]{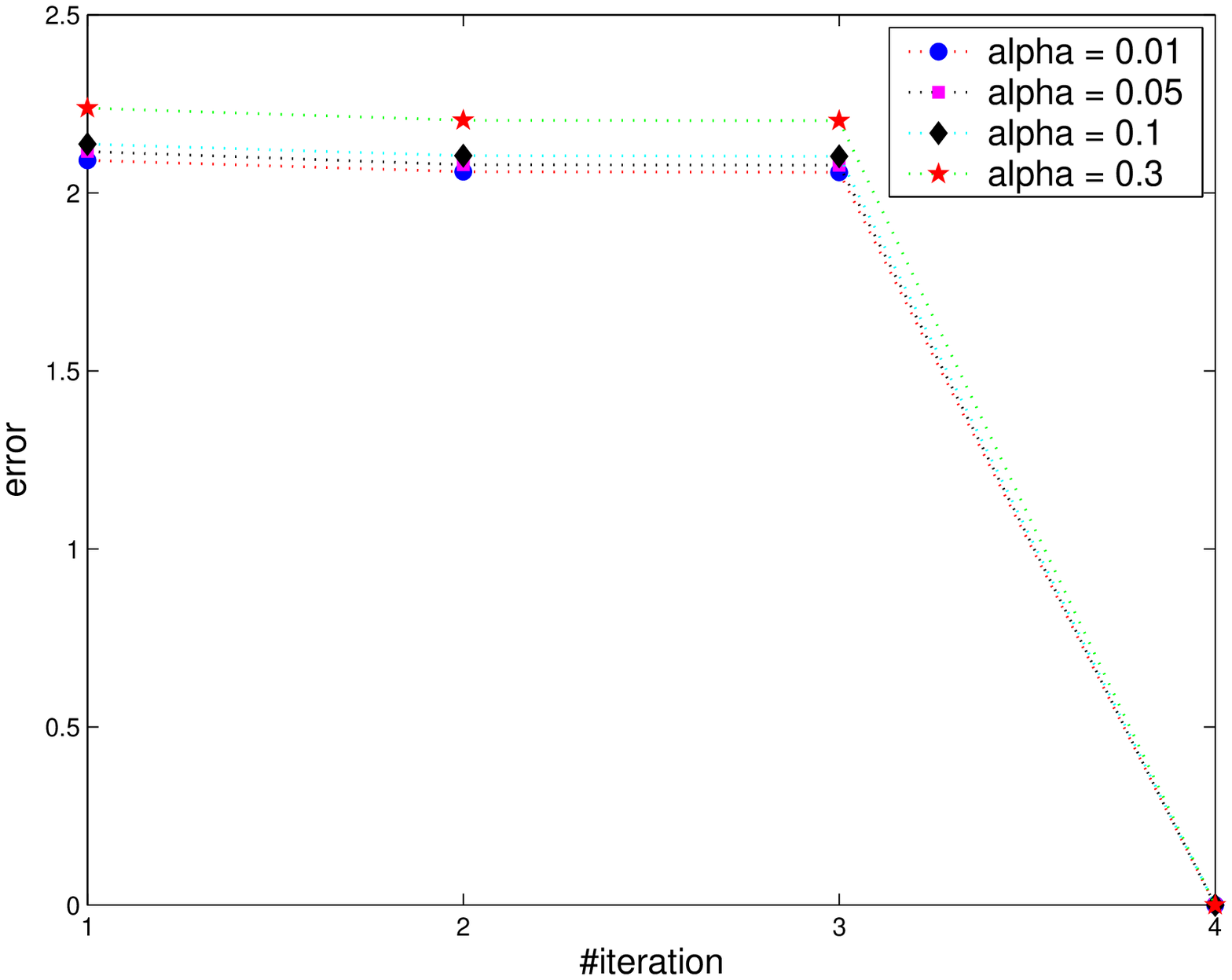}
   \label{fig6a}
  }
  \subfigure[Medium to large $\alpha$]{
   \includegraphics[width=0.45\textwidth]{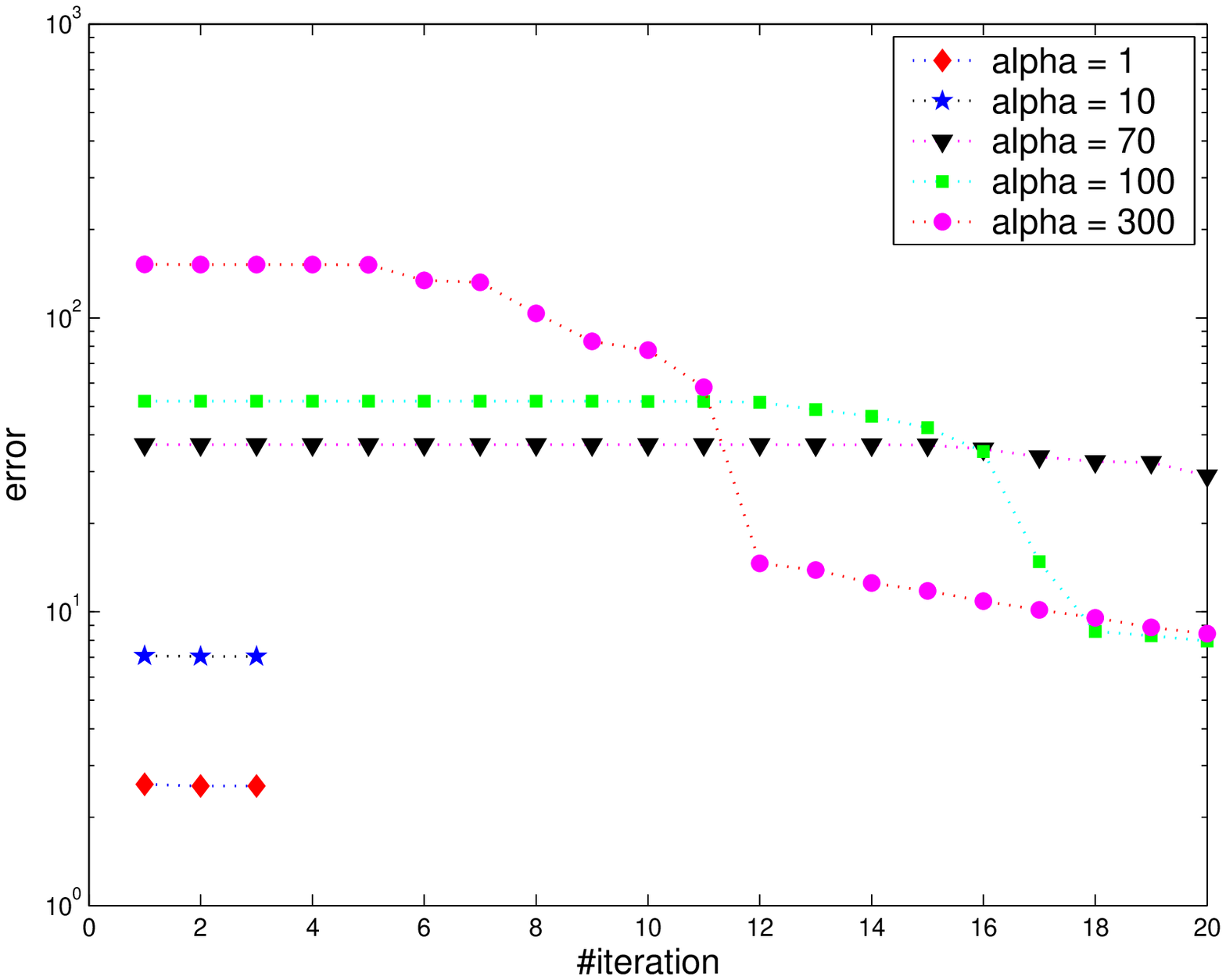}
   \label{fig6b}
  }
\\
  \subfigure[Some values of $\alpha$]{
   \includegraphics[width=0.7\textwidth]{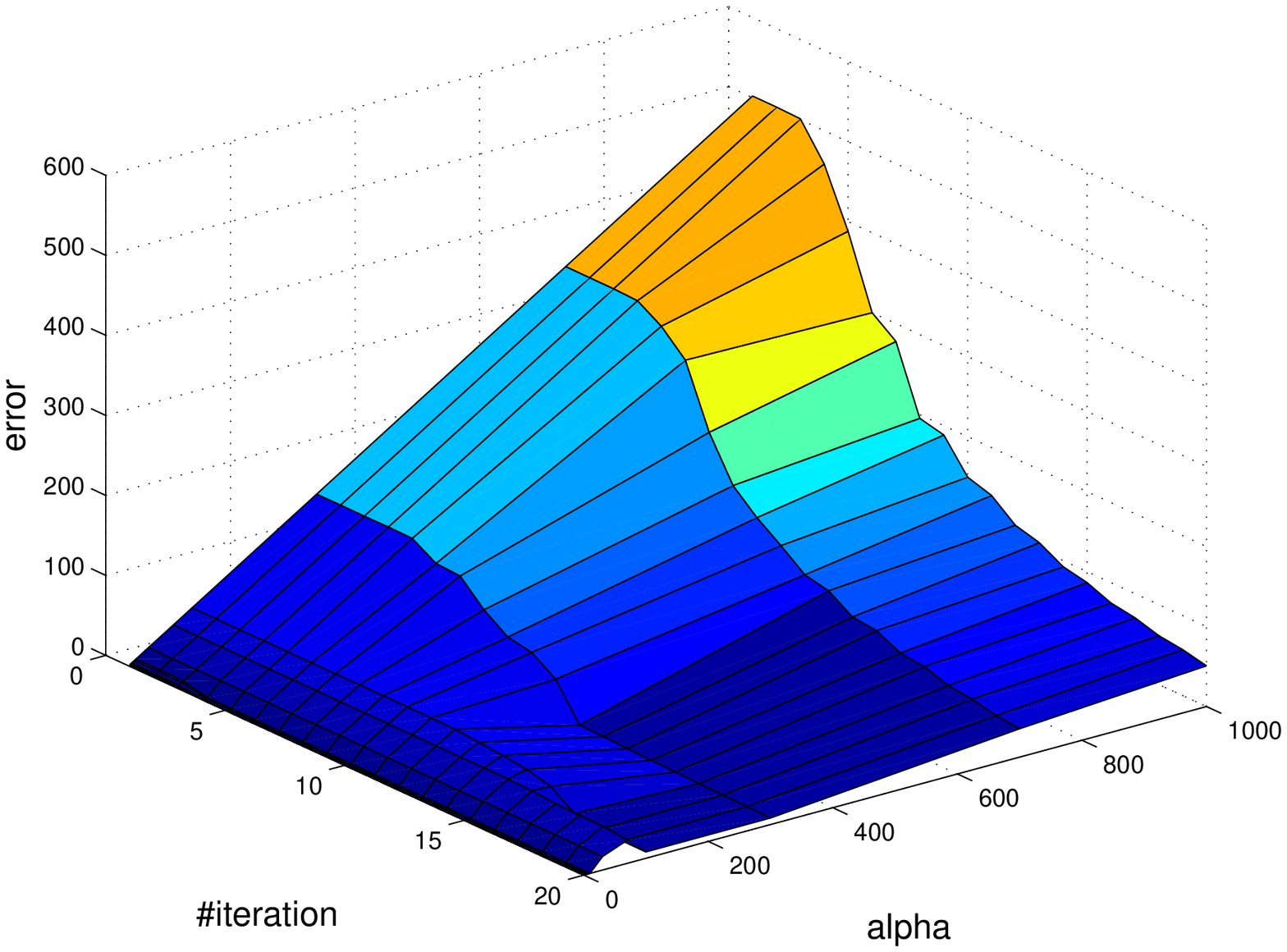}
   \label{fig6c}
  }
  \caption{AU-B($\alpha$) error per iteration for Reuters4 dataset ($\beta=1$).}
  \label{fig6}
 \end{center}
\end{figure}

\begin{figure}
 \begin{center}
  \subfigure[Small $\beta$]{
   \includegraphics[width=0.45\textwidth]{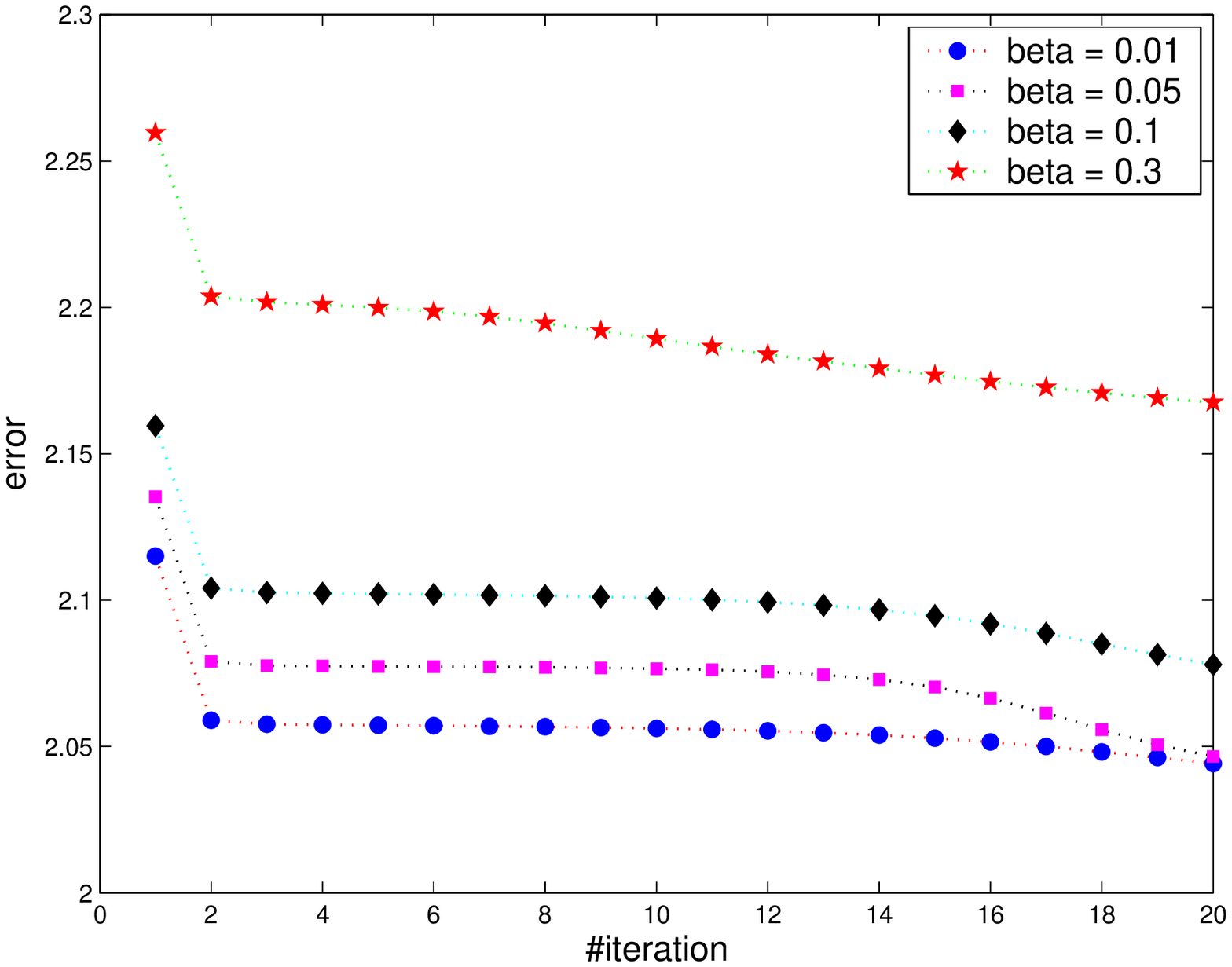}
   \label{fig7a}
  }
  \subfigure[Medium to large $\beta$]{
   \includegraphics[width=0.45\textwidth]{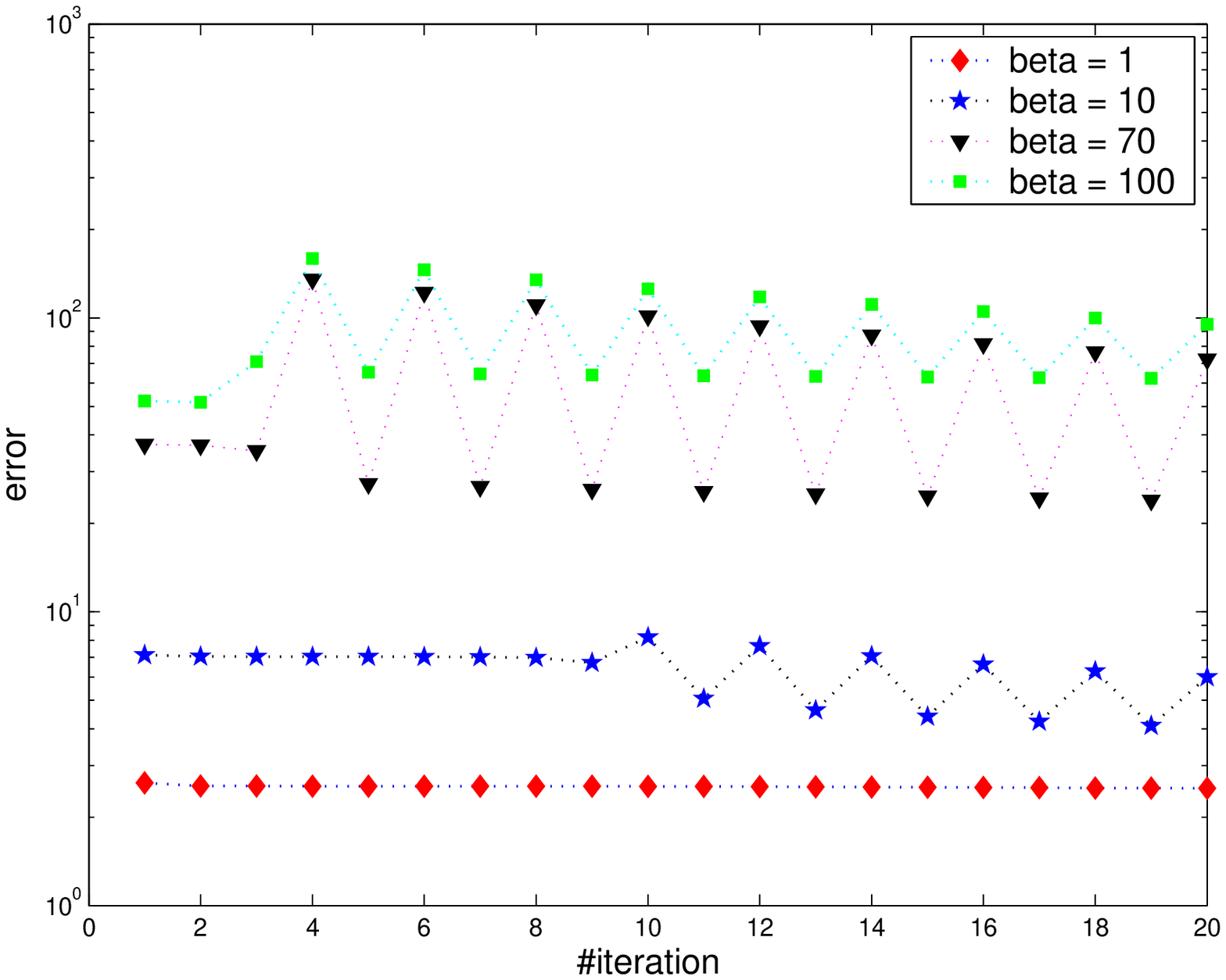}
   \label{fig7b}
  }
\\
  \subfigure[Some values of $\beta$]{
   \includegraphics[width=0.7\textwidth]{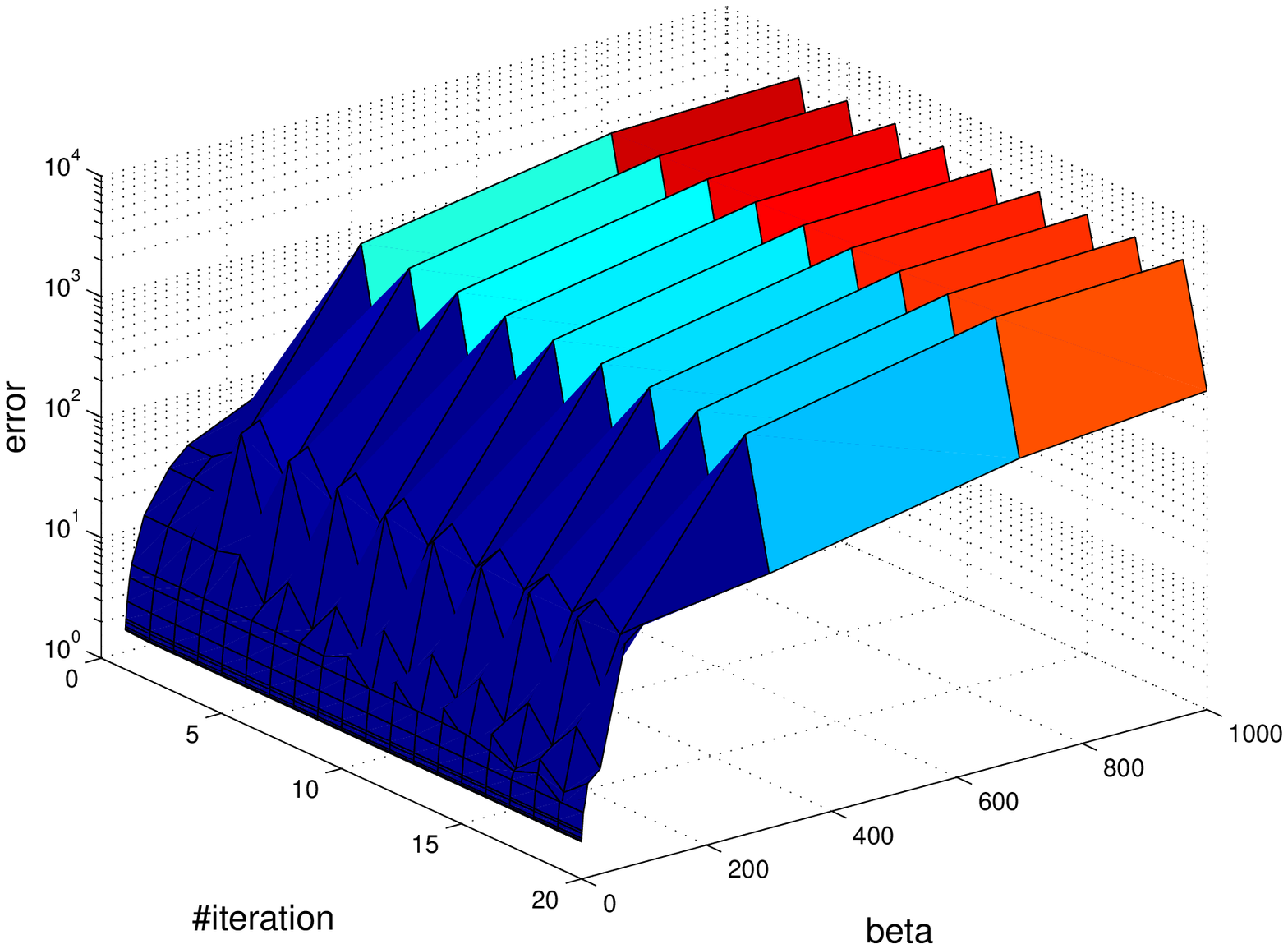}
   \label{fig7c}
  }
  \caption{MU-B($\beta$) error per iteration for Reuters4 dataset ($\alpha=1$).}
  \label{fig7}
 \end{center}
\end{figure}

\begin{figure}
 \begin{center}
  \subfigure[Small $\beta$]{
   \includegraphics[width=0.45\textwidth]{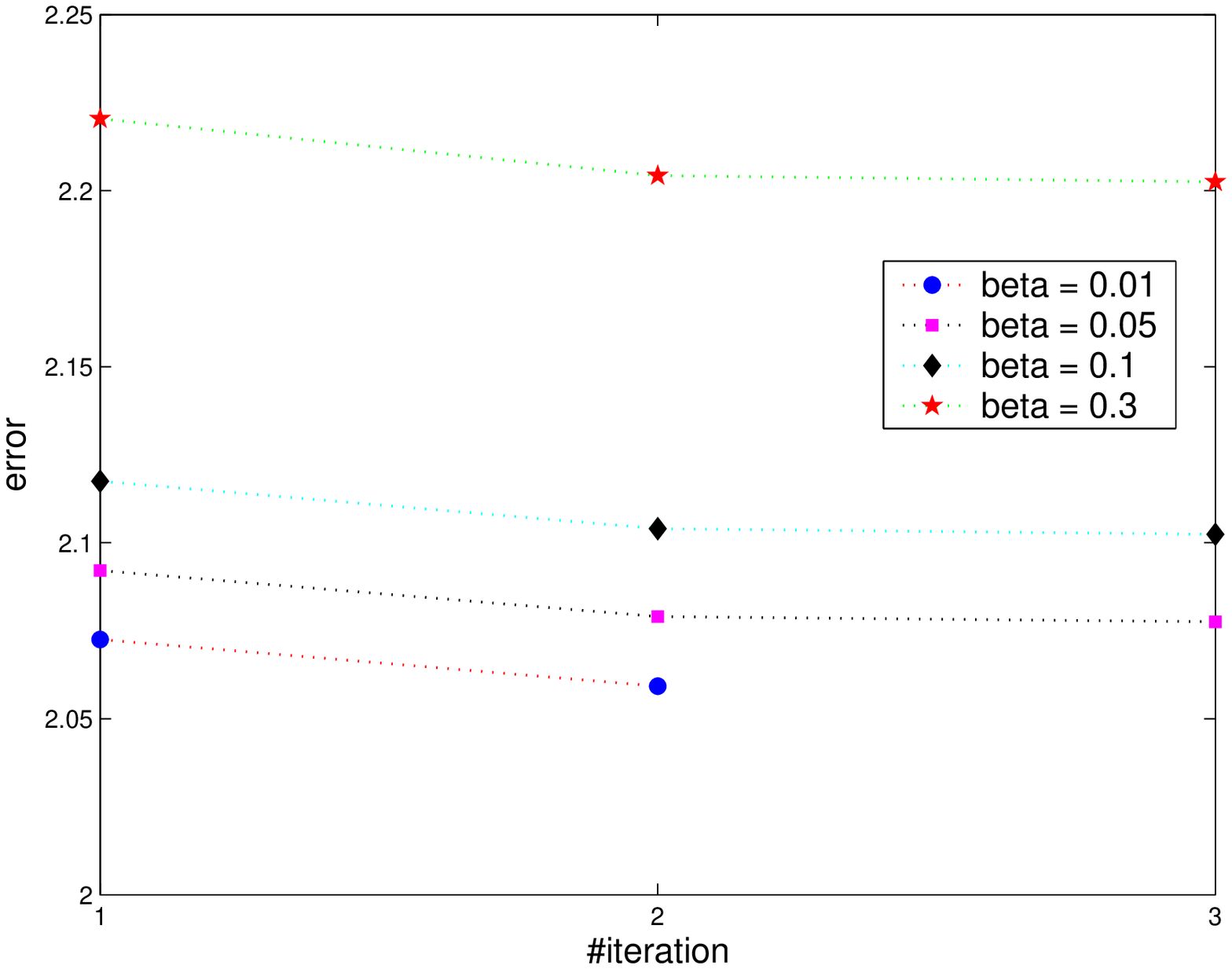}
   \label{fig8a}
  }
  \subfigure[Medium to large $\beta$]{
   \includegraphics[width=0.45\textwidth]{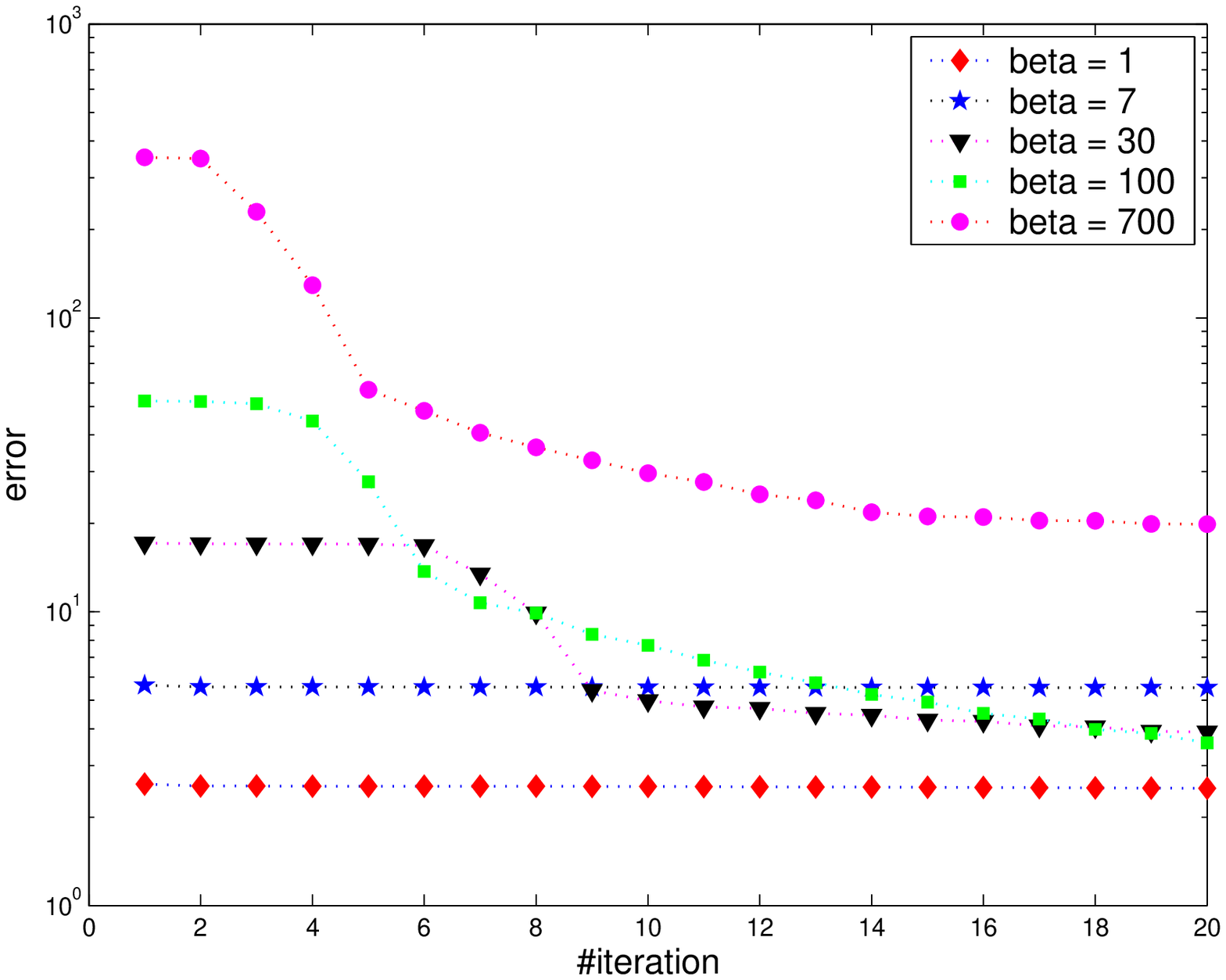}
   \label{fig8b}
  }
\\
  \subfigure[Some values of $\beta$]{
   \includegraphics[width=0.7\textwidth]{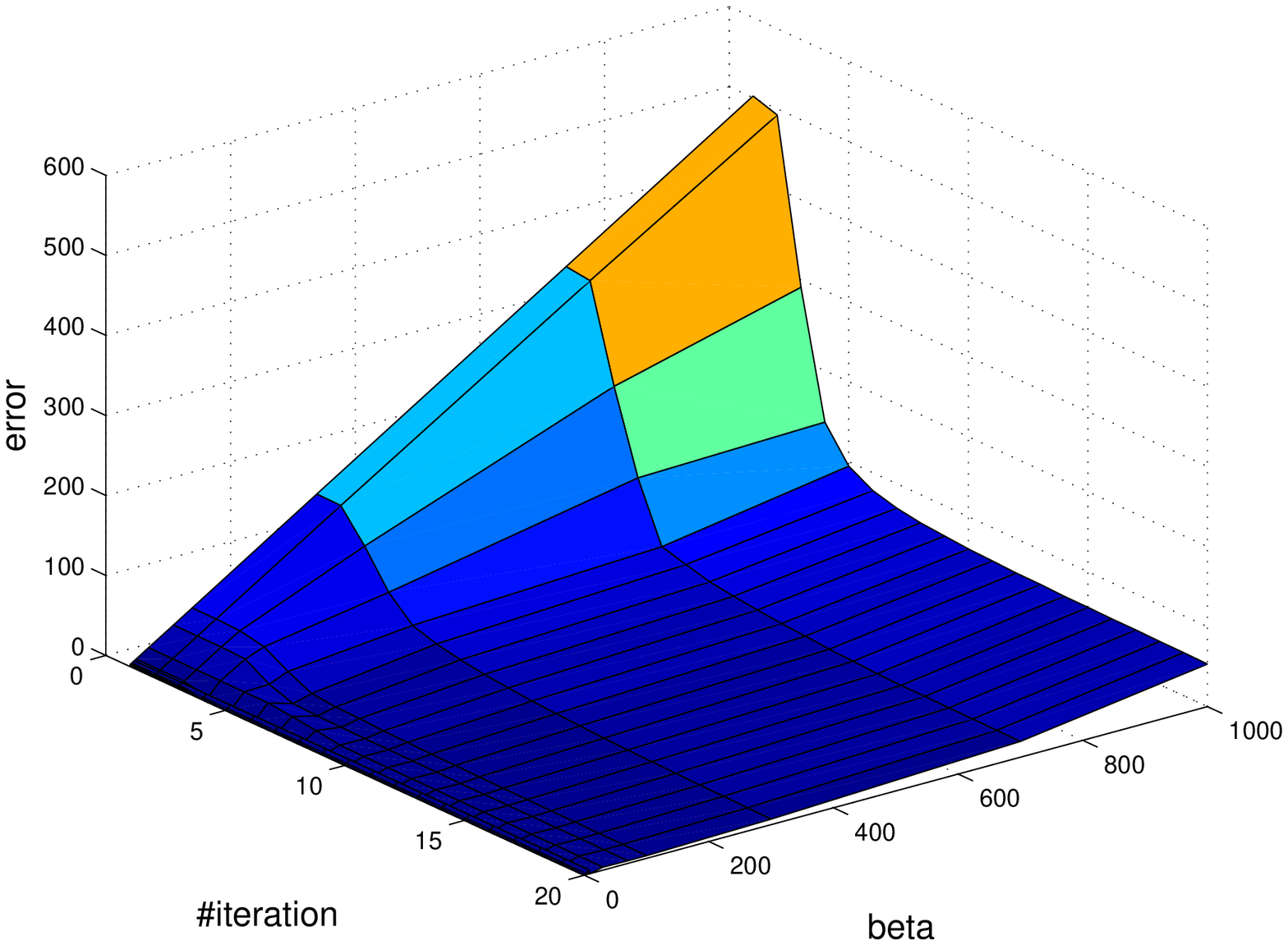}
   \label{fig8c}
  }
  \caption{AU-B($\beta$) error per iteration for Reuters4 dataset ($\alpha=1$).}
  \label{fig8}
 \end{center}
\end{figure}

Figure \ref{fig5}--\ref{fig8} show the equivalent results for BNMF cases. Because there are two adjustable parameters, $\alpha$ and $\beta$, we fix one parameter while studying the other. Figure \ref{fig5} and \ref{fig6} show the results for fixed $\beta=1$, and figure \ref{fig7} and \ref{fig8} for fixed $\alpha=1$. As in UNMF cases, while MU-B fails to show the nonincreasing property for large $\alpha$ and $\beta$ values, AU-B successfully preserves this property regardless of $\alpha$ and $\beta$ values. Note that we set $\delta=\sigma=10^{-8}$, and $\mathrm{step}=10$ for MU-U, AU-U, MU-B, and AU-B in all experiments.

However, there are computational tradeoff for these accuracies as for large $\alpha$ and/or $\beta$, AU rules based algorithms are slower than their MU counterparts. Table \ref{table3} shows time comparisons between these algorithms for Reuters4 dataset. Note that, $\alpha$ or $\beta$ appended to the algorithm's acronyms to tell which parameter is being varied. For example AU-B($\alpha$) means AU-B with fixed $\beta$ and varied $\alpha$.

\begin{table}
\renewcommand{\arraystretch}{1}
  \begin{center}
    \caption{Time comparison (seconds) for Reuters4 dataset.}
    \centering
    \begin{tabular}{|r|r|r|r|r|r|r|}
    \hline
    $\alpha$/$\beta$ & MU-U & AU-U & MU-B($\alpha$) & AU-B($\alpha$) & MU-B($\beta$) & AU-B($\beta$)\\
    \hline

    0.01   & 110 & 110 & 121 & 41.1 & 122 & 27.2\\
    0.05   & 110 & 110 & 121 & 40.9 & 121 & 40.7\\
     0.1   & 109 & 109 & 121 & 40.8 & 121 & 41.2\\
     0.3   & 110 & 109 & 121 & 40.4 & 121 & 41.1\\
     0.7   & 110 & 110 & 121 & 272  & 121 & 41.2\\
       1   & 110 & 110 & 121 & 40.8 & 121 & 273\\
       3   & 110 & 110 & 121 & 40.4 & 121 & 40.7\\
       7   & 110 & 110 & 121 & 40.4 & 121 & 273\\
      10   & 110 & 110 & 121 & 40.8 & 121 & 41.1\\
      30   & 109 & 110 & 121 & 272  & 121 & 442\\
      70   & 109 & 137 & 121 & 332  & 121 & 525\\
     100   & 110 & 232 & 121 & 382  & 121 & 605\\
     300   & 110 & 232 & 121 & 514  & 121 & 579\\
     700   & 110 & 461 & 121 & 607  & 121 & 606\\
    1000   & 110 & 411 & 121 & 606  & 121 & 365\\
    \hline
    \end{tabular}
    \label{table3}
  \end{center}
\end{table}

As shown in table \ref{table3}, the computational times of MU algorithms practically are independent from $\alpha$ and $\beta$ values. And AU algorithms seem to become slower for some large $\alpha$ or $\beta$. This probably because for large $\alpha$ or $\beta$ values, the AU algorithms execute the inner iterations (shown as $\mathbf{repeat}$ $\mathbf{until}$ loops in algorithm \ref{algorithm6} and \ref{algorithm9}). Also, there are some anomalies in the AU-B($\alpha$) and AU-B($\beta$) cases, where for some $\alpha$ or $\beta$ values, execution times are unexpectedly very fast. To investigate these, we display number of iteration (\#iter) and inner iteration (\#initer) for AU algorithms in table \ref{table4}. Note that MU algorithms reach maximum predefined number of iteration for all cases: 20 iterations.

As shown in table \ref{table4}, when AU algorithms perform worse than their MU counterparts, then they execute the inner iteration which happened for large $\alpha$/$\beta$. And when AU algorithms perform better, then their \#iter are smaller than \#iter of MU algorithms (and the inner iteration is not executed). These explain the differences in computational times in table \ref{table3}.

\begin{table}
\renewcommand{\arraystretch}{1}
  \begin{center}
    \caption{\#iter and \#initer of AU algorithms (Reuters4).}
    \centering
    \begin{tabular}{|r|r|r|r|r|r|r|}
    \hline
    $\alpha$/$\beta$ & AU-U & AU-B($\alpha$) & AU-B($\beta$)\\
    \hline
    & \#iter / \#initer & \#iter / \#initer & \#iter / \#initer \\
    \hline
    0.01   & 20 / 0 &  3 / 0 &  2 / 0 \\
    0.05   & 20 / 0 &  3 / 0 &  3 / 0 \\
     0.1   & 20 / 0 &  3 / 0 &  3 / 0 \\
     0.3   & 20 / 0 &  3 / 0 &  3 / 0 \\
     0.7   & 20 / 0 & 20 / 0 &  3 / 0 \\
       1   & 20 / 0 &  3 / 0 & 20 / 0 \\
       3   & 20 / 0 &  3 / 0 &  3 / 0 \\
       7   & 20 / 0 &  3 / 0 & 20 / 0 \\
      10   & 20 / 0 &  3 / 0 &  3 / 0 \\
      30   & 20 / 0 & 20 / 0 & 20 / 44 \\
      70   & 20 / 7 & 20 / 23 & 20 / 66 \\
     100   & 20 / 32 & 20 / 22 & 20 / 88 \\
     300   & 20 / 32 & 20 / 65 & 20 / 81 \\
     700   & 20 / 92 & 20 / 75 & 20 / 88 \\
    1000   & 20 / 79 & 20 / 90 & 20 / 24 \\
    \hline
    \end{tabular}
    \label{table4}
  \end{center}
\end{table}

\subsection{Maximum number of iteration}

Maximum number of iteration is very crucial in MU and AU algorithms since these algorithms are known to be very slow \cite{CJLin2, Hoyer, HKim2, JKim, Shahnaz, Berry, HKim, DKim, DKim2, JKim2, CJLin}. As shown by Lin \cite{CJLin}, LS is very fast to minimize the objective for some first iterations, but then tends to become slower. In table \ref{table5}, we display errors for some first iterations for LS, MU-U, AU-U, MU-B, and AU-B. We choose the cases where $\alpha=0.1$ and $\beta=1$ since for these values, our algorithms are settled. Note that error0 refers to the initial error before the algorithms start running, and error$n$ is the error at $n$-th iteration. 

As shown in table \ref{table5} all algorithms are exceptionally very good at reducing errors in the first iterations. But then, the improvements are rather negligible with respect to the first improvements and the sizes of the datasets. Accordingly, we set maximum number of iteration to 20.

\begin{table}
 \begin{center}
   \caption{Errors for some first iterations (Reuters4).}
   \centering
   \begin{tabular}{|l|r|r|r|r|r|r|}
   \hline
   & error0 & error1 & error2 & error3 & error4 & error5 \\
   \hline

   LS   & 1373  & 0.476 & 0.474 & 0.472 & 0.469 & 0.466 \\
   MU-U & 4652  & 1.681 & 1.603 & 1.596 & 1.591 & 1.583 \\
   AU-U & 4657  & 1.681 & 1.605 & 1.595 & 1.586 & 1.573 \\
   MU-B & 12474 & 2.164 & 2.104 & 2.103 & 2.102 & 2.102 \\
   AU-B & 12680 & 2.137 & 2.104 & 2.103 & -     & -     \\
   \hline
  \end{tabular}
  \label{table5}
 \end{center}
\end{table}

\subsection{Determining $\alpha$ and $\beta$}

In our proposed algorithms, there are two dataset-dependent parameters, $\alpha$ and $\beta$, that have to be learned first. Because orthogonal NMFs are introduced to improve clustering capability of the standard NMF \cite{Ding1}, these parameters will be learned based on clustering results on test dataset. We will used Reuters4 for this purpose. These parameters do not exist in the original orthogonal NMFs \cite{Ding1} nor in other orthogonal NMF algorithms \cite{Yoo1, Yoo2, Choi}. However, we notice that our formulations resemble sparse NMF formulation \cite{HKim2,JKim,HKim}, or in general case also known as constrained NMF \cite{Pauca}. As shown in ref.~\cite{HKim2,JKim,HKim}, sparse NMF usually can give good results if $\alpha$ and/or $\beta$ are rather small positive numbers.

To determine $\alpha$ and $\beta$, we evaluate clustering qualities produced by our algorithms as $\alpha$ or $\beta$ values grow measured by the standard metrics: \emph{mutual information} (MI), \emph{entropy} (E), \emph{purity} (P), and \emph{Fmeasure} (F) (see section \ref{metrics} for discussions on these metrics).

\begin{figure}
 \begin{center}

  \subfigure[MU-U]{
   \includegraphics[width=0.45\textwidth]{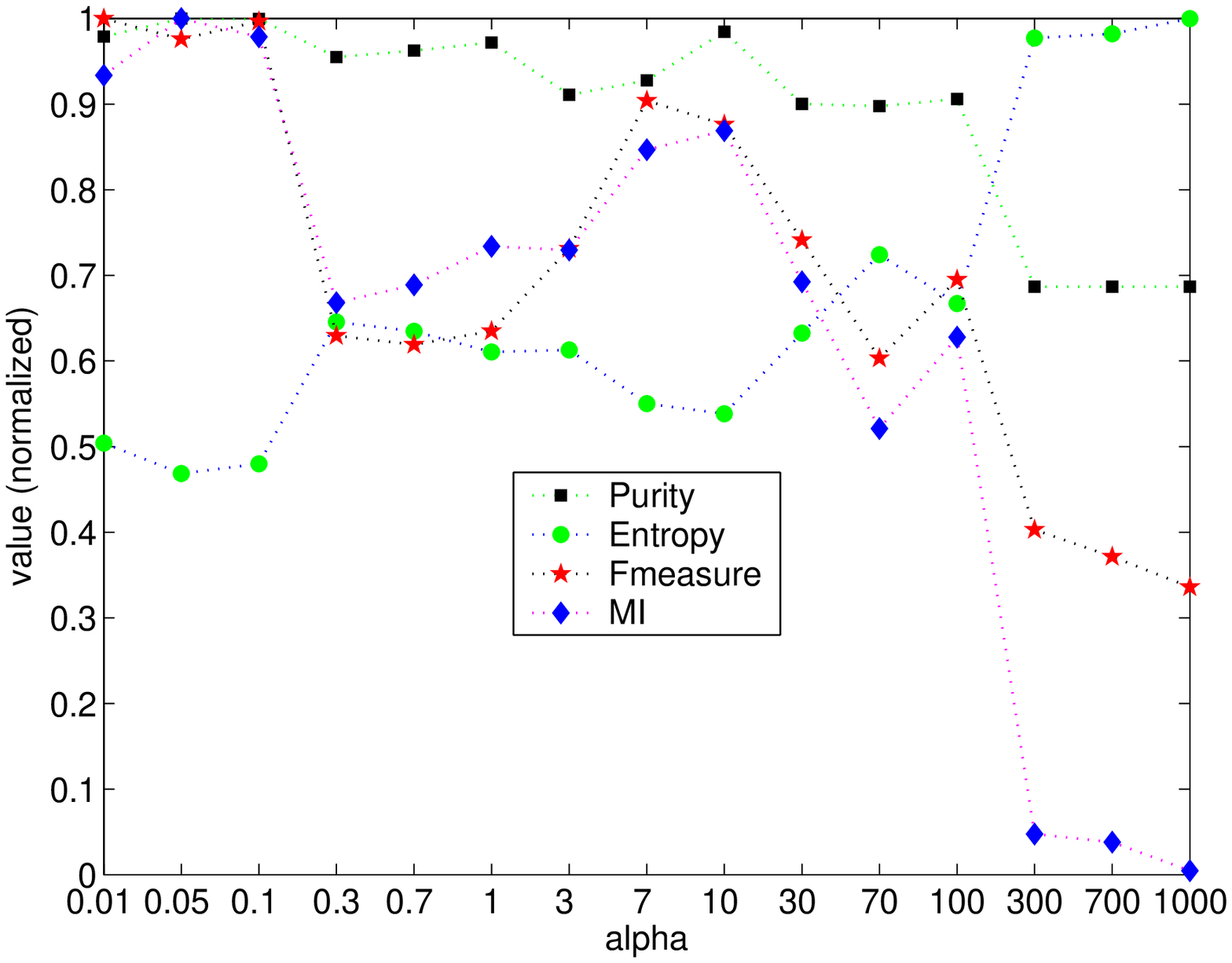}
   \label{fig9a}
  }
  \subfigure[AU-U]{
   \includegraphics[width=0.45\textwidth]{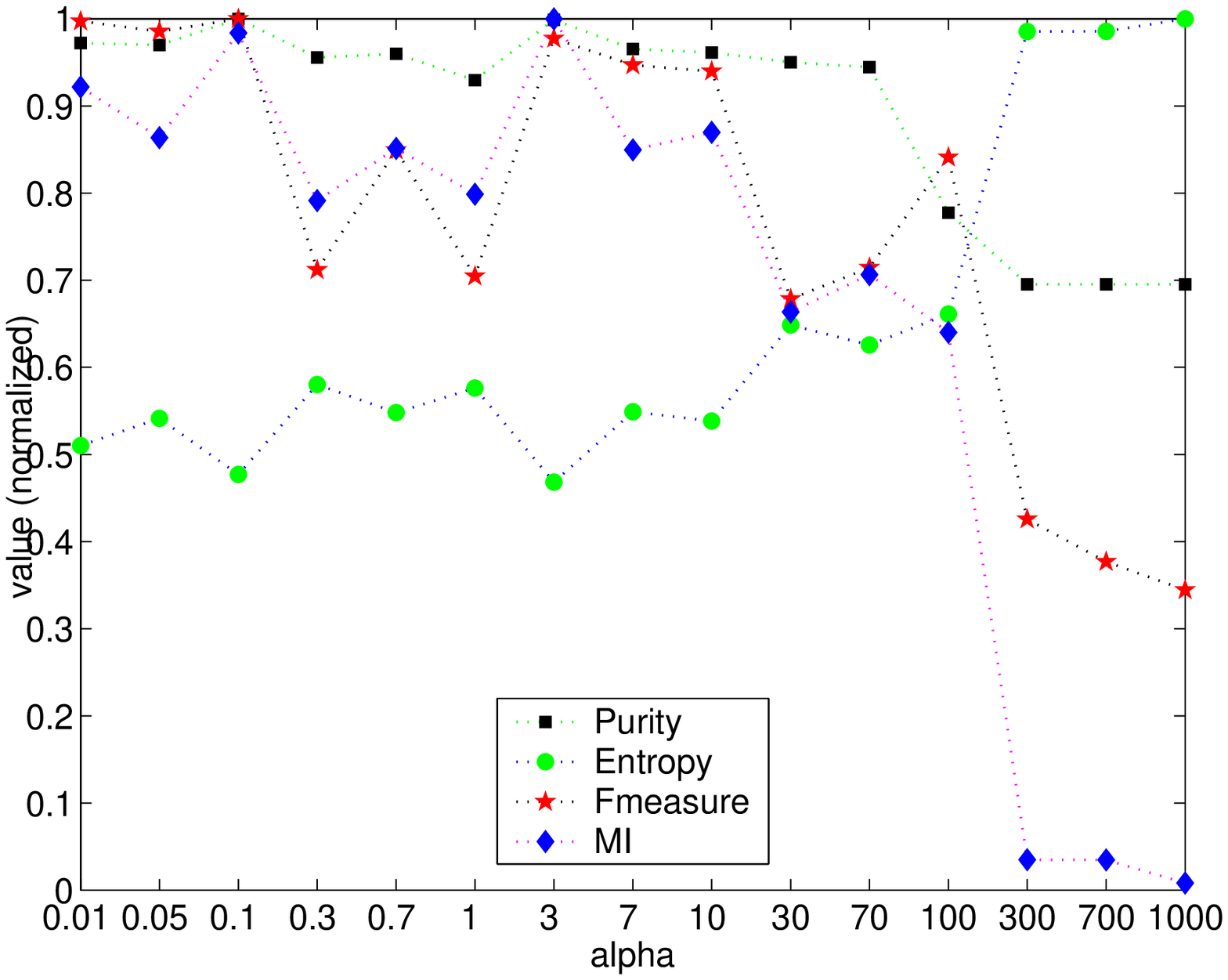}
   \label{fig9b}
  } 
\\
  \subfigure[MU-B($\alpha$), $\beta=1$]{
   \includegraphics[width=0.45\textwidth]{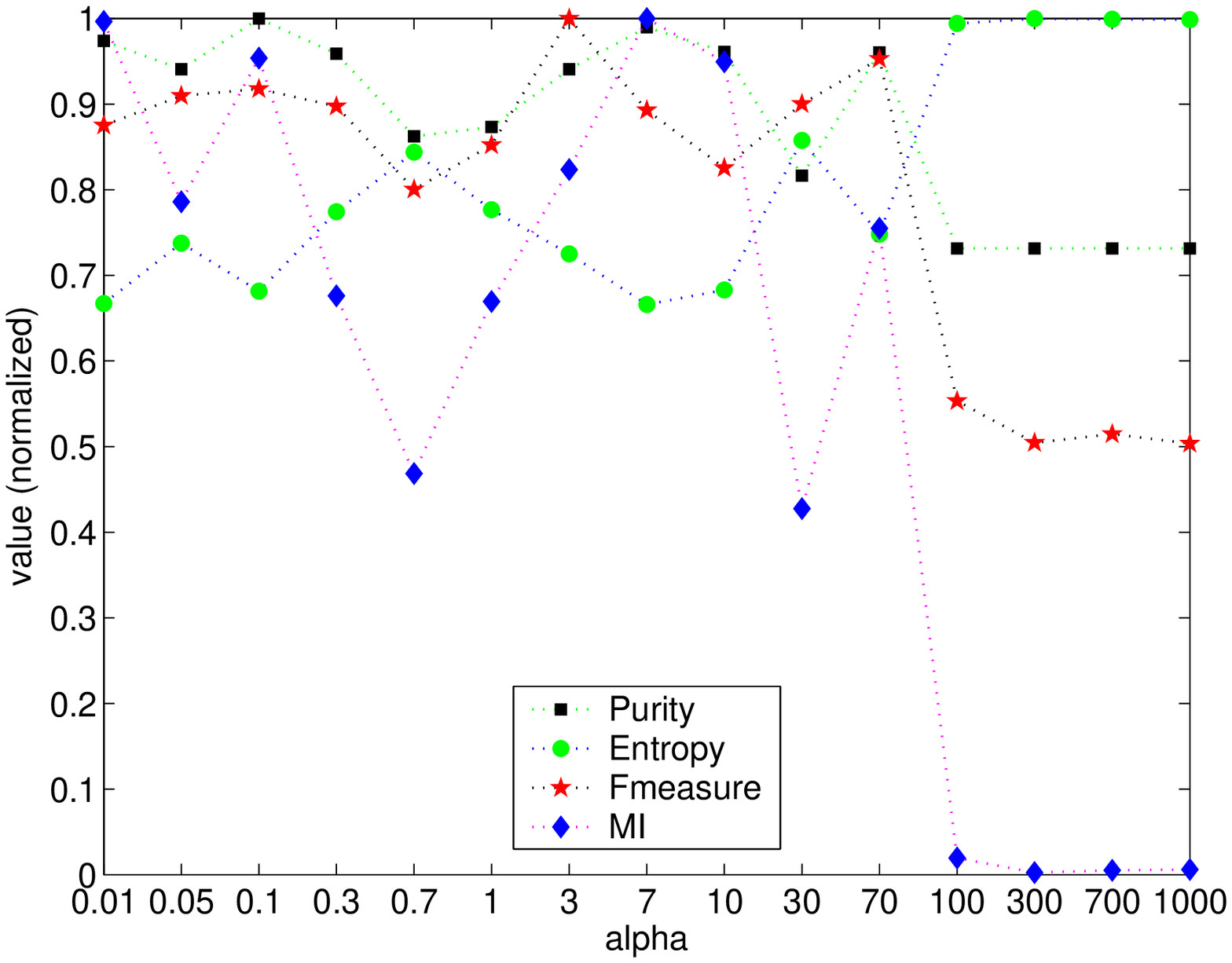}
   \label{fig9c}
  }
  \subfigure[AU-B($\alpha$), $\beta=1$]{
   \includegraphics[width=0.45\textwidth]{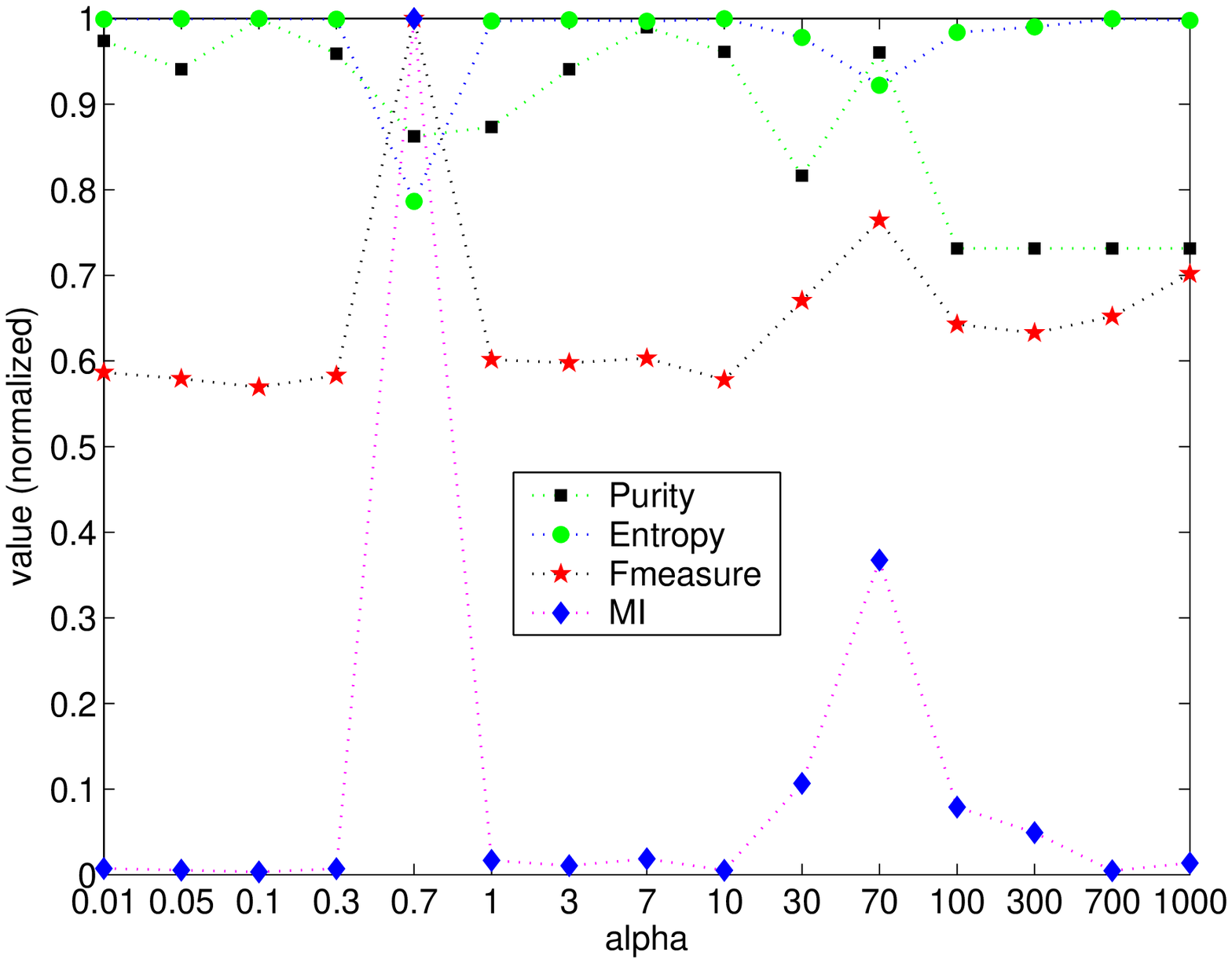}
   \label{fig9d}
  }
\\
  \subfigure[MU-B($\beta$), $\alpha=1$]{
   \includegraphics[width=0.45\textwidth]{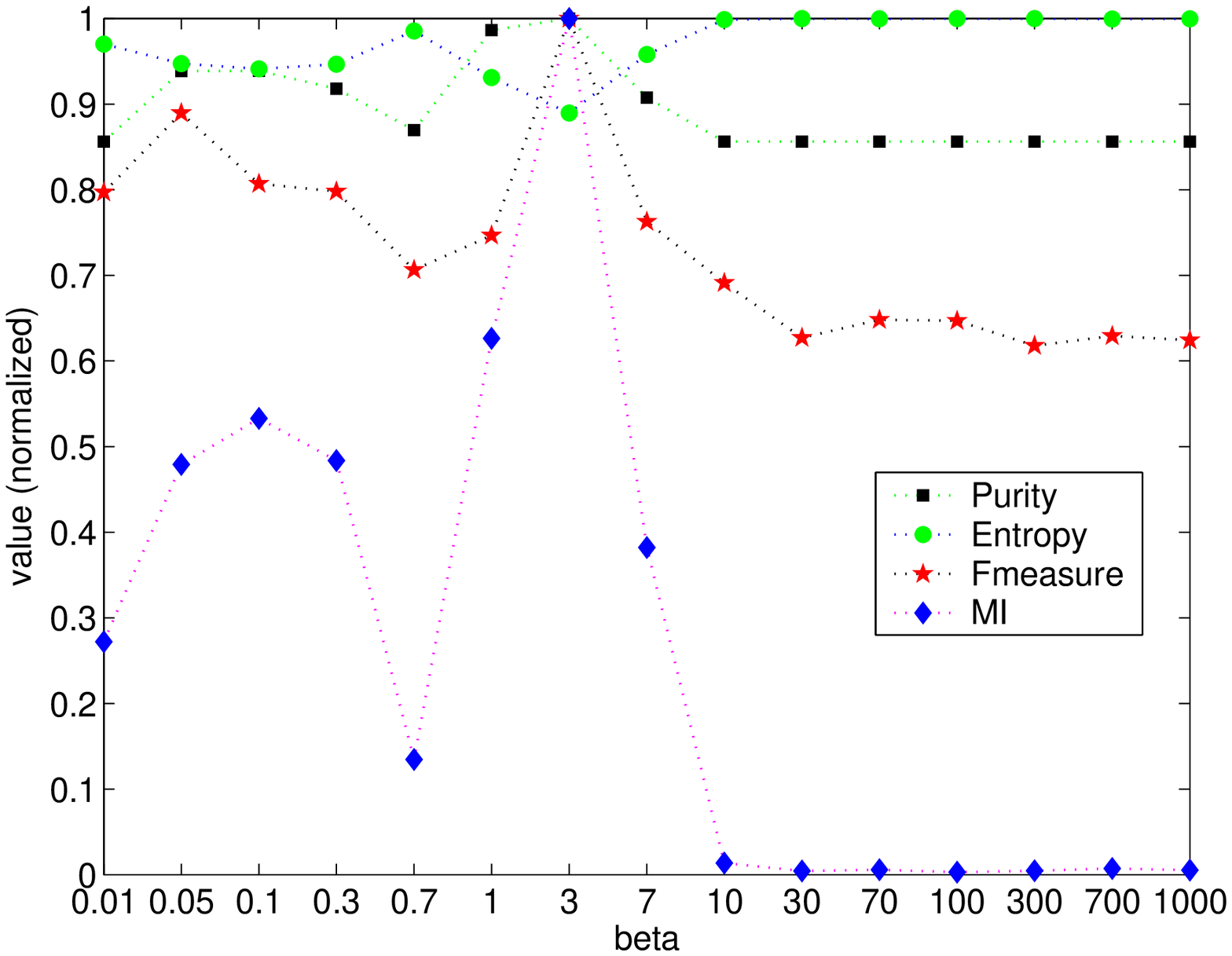}
   \label{fig9e}
  }
  \subfigure[AU-B($\beta$), $\alpha=1$]{
   \includegraphics[width=0.45\textwidth]{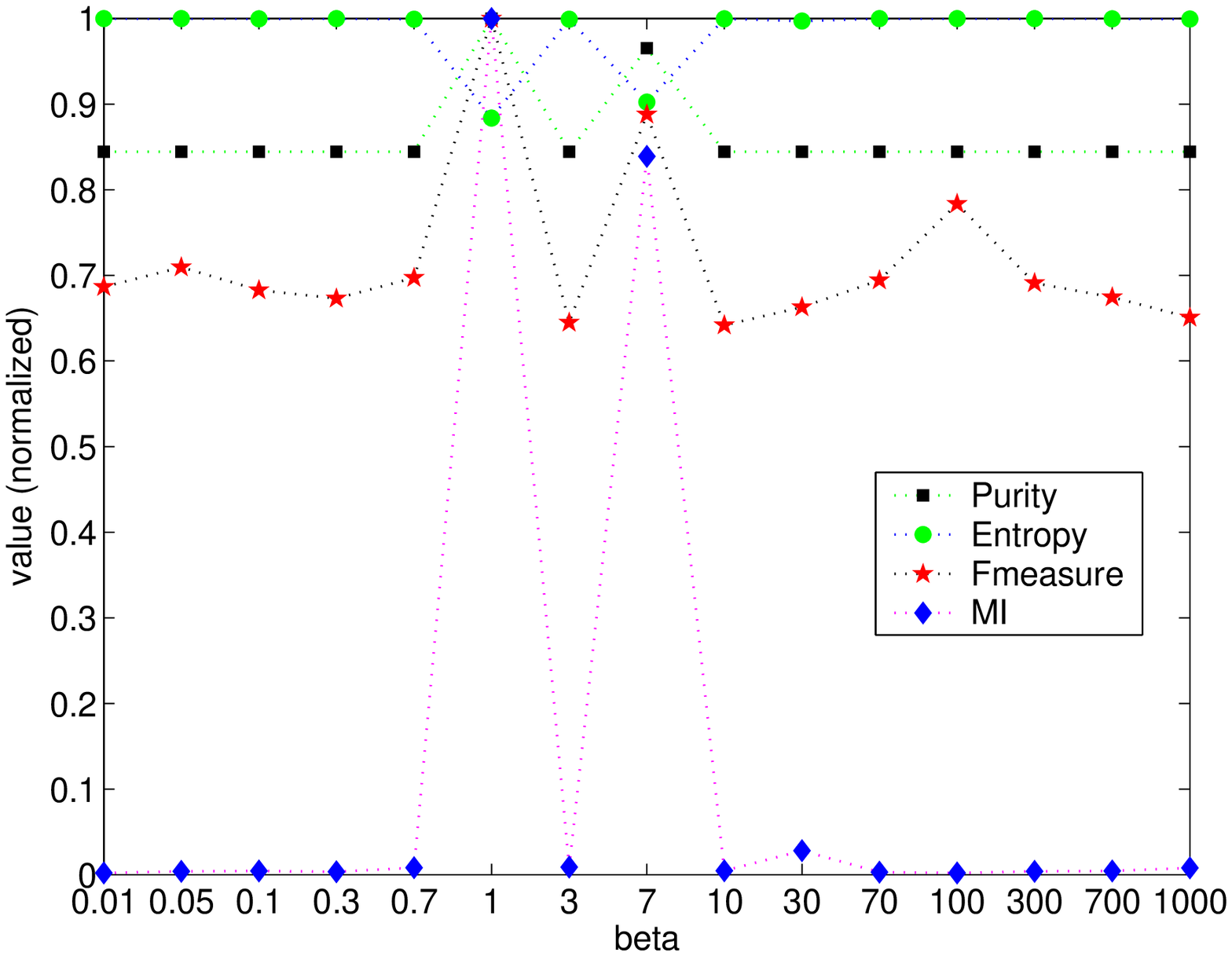}
   \label{fig9f}
  }
\caption{Clustering qualities as functions of $\alpha$ or $\beta$ for Reuters4.}
  \label{fig9}
 \end{center}
\end{figure}

As shown in figure \ref{fig9}, for UNMF algorithms (MU-U and AU-U) $\alpha=0.1$ seems to be a good choice. For MU-B it seems that $\alpha=0.1$ and $\beta=3$ are acceptable settings. And for AU-B, $\alpha=0.7$ and $\beta=1$ seem to be good settings. Based on this results, we decide to set $\alpha=0.1$ and $\beta=1$ for all datasets and algorithms. Note that, other mechanisms like using some small samples for deriving optimal $\alpha$s and $\beta$s for each dataset and algorithm may be a better choice since every dataset can have different characteristics.

\subsection{Times, \#iterations, and errors}

To evaluate computational performances of the algorithms, we measure their average and maximum running times, average and maximum \#iterations, and average and maximum errors produced at the last iterations for 10 trials. Table \ref{table6}-\ref{table8} show the results.

\begin{figure}
\begin{center}
\includegraphics[width=0.6\textwidth]{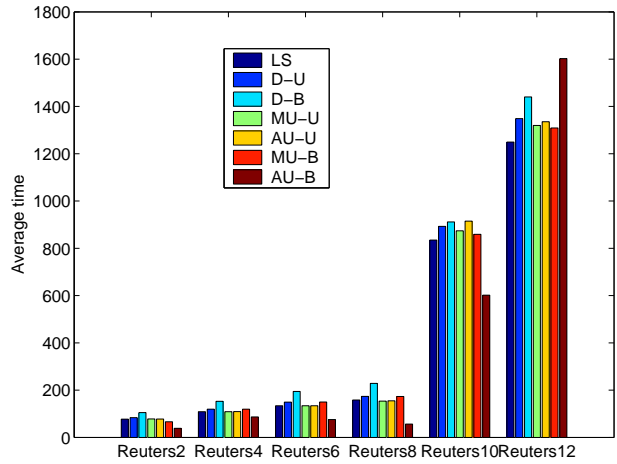}
\caption{Average running time comparison as the datasets grow.}
\label{fig10}
\end{center}
\end{figure}

As shown in the table \ref{table6}, LS generally is the fastest with exception when MU-B or AU-B converge before reaching the maximum iteration (20 iterations), then these algorithms will outperform LS. Our uni-orthogonal algorithms (MU-U and AU-U) seem to have comparable running times with LS. MU-B seems to be slower for smaller datasets and then performs better than MU-U and AU-U for bigger datasets: Reuters10 and Reuters12. Since AU-B usually converges before reaching the maximum iteration, comparison can be done by using maximum running times for Reuters4, Reuters6, Reuters10, and Reuters12 in which the data is available (see table \ref{table7}). As shown, AU-B is the slowest to perform calculation per iteration. There are also abrupt changes in the running times for Reuters10 and Reuters12 for all algorithms which are unfortunate since as shown in table \ref{ch2:table3}, the sizes of the datasets only slightly change. Figure \ref{fig10} shows the bar chart of average running times as the sizes of the datasets grow.

\begin{figure}
 \begin{center}
  \subfigure[MU-U and AU-U.]{
   \includegraphics[width=0.45\textwidth]{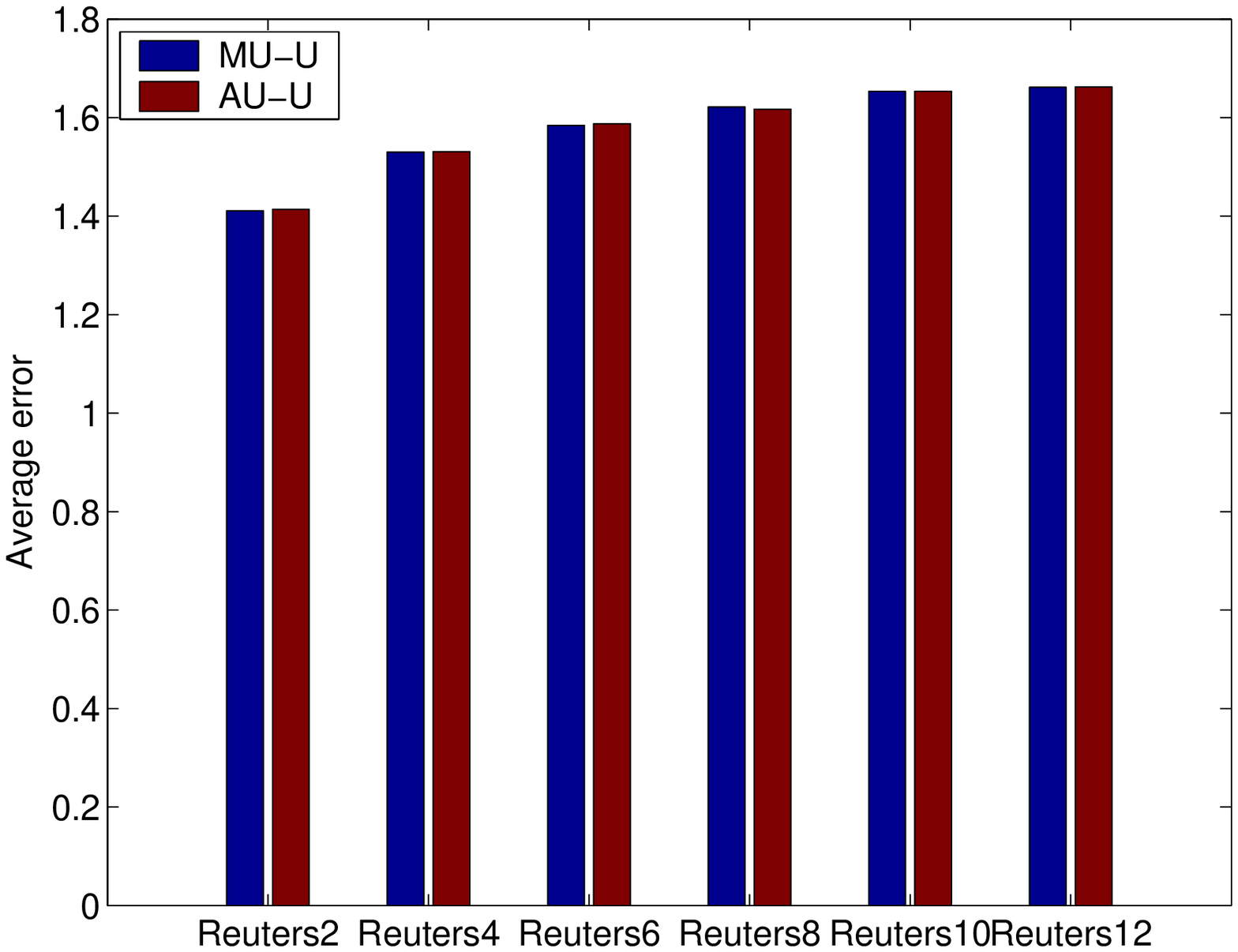}
   \label{fig11a}
  }
  \subfigure[MU-B and AU-B.]{
   \includegraphics[width=0.45\textwidth]{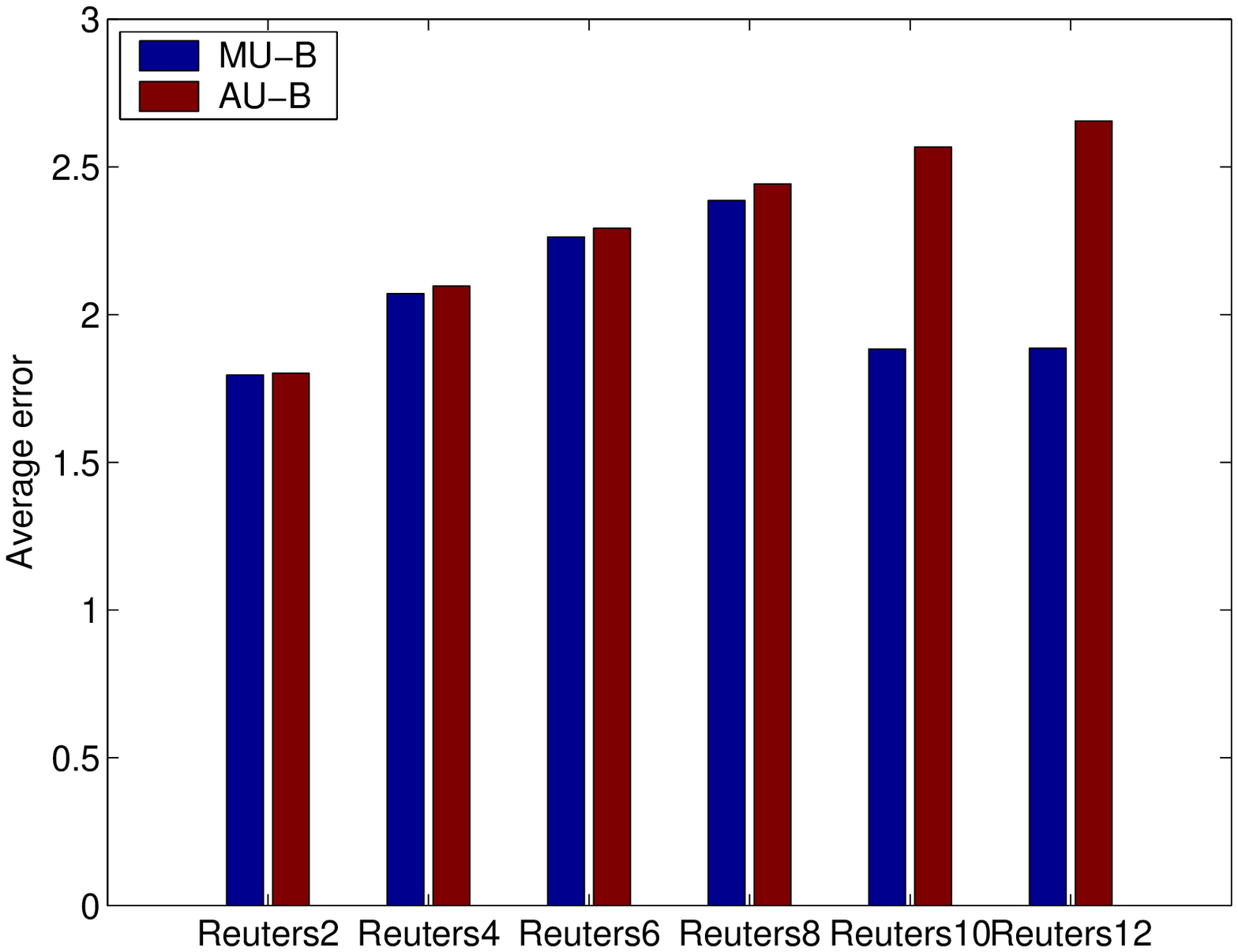}
   \label{fig11b}
  }
  \caption{Average errors comparison as the datasets grow.}
  \label{fig11}
 \end{center}
\end{figure}

Average and maximum errors at the last iterations are shown in table \ref{table8}. Results for D-U and D-B support the previous results: algorithm \ref{algorithm2} and \ref{algorithm3} do not minimize the objectives that are supposed to be minimized, i.e., eq.~\ref{eq23} and \ref{eq39}. Because only MU-U \& AU-U and MU-B \& AU-B pairs have the same objective each, we compare average errors for these pairs in figure \ref{fig11}. There is no significant difference between MU-U \& AU-U in the average errors, but as shown in figure \ref{fig10}, MU-U has better average running times especially for larger datasets. And for MU-B \& AU-B, the differences in the average errors grow as the size and classes of the datasets grow with significant differences happened at Reuters10 and Reuters12. However, as shown in table \ref{table7}, AU-B is more likely to converge, so generally its running times are shorter. 

\begin{table}
\renewcommand{\arraystretch}{1}
 \begin{center}
   \caption{Average and maximum running time.}
   \centering
   \small{
   \begin{tabular}{|l|l|r|r|r|r|r|r|r|}
   \hline
Data & Time & LS & D-U & D-B & MU-U & AU-U & MU-B & AU-B \\
\hline
Reuters2 & Av.  & 77.266 & 83.655 & 104.98 & 78.068 & 77.825 & 66.318 & 38.367 \\
         & Max. & 79.031 & 84.743 & 106.25 & 79.075 & 79.176 & 83.960 & 49.477 \\ \hline 
Reuters4 & Av.  & 108.84 & 119.42 & 152.77 & 109.04 & 109.12 & 119.46 & 86.745 \\
         & Max. & 109.39 & 119.55 & 153.17 & 109.20 & 109.28 & 119.72 & 271.40 \\ \hline 
Reuters6 & Av.  & 134.02 & 149.32 & 194.43 & 133.91 & 134.19 & 149.63 & 75.432 \\
         & Max. & 134.50 & 149.62 & 194.75 & 134.27 & 134.51 & 149.95 & 327.70 \\ \hline 
Reuters8 & Av.  & 158.37 & 173.43 & 228.59 & 153.53 & 155.03 & 173.00 & 56.464 \\
         & Max. & 181.58 & 175.71 & 235.54 & 155.15 & 159.19 & 174.05 & 59.021 \\ \hline 
Reuters10 & Av. & 834.69 & 892.91 & 911.34 & 874.18 & 914.93 & 859.31 & 601.57 \\
         & Max. & 1004.5 & 1141.2 & 1127.3 & 1137.5 & 1162.0 & 1059.0 & 2794.1 \\ \hline 
Reuters12 & Av. & 1249.2 & 1348.4 & 1440.1 & 1319.7 & 1335.6 & 1309.0 & 1602.4 \\
         & Max. & 1389.0 & 1590.4 & 1746.1 & 1565.7 & 1529.4 & 1506.7 & 4172.2 \\ 
\hline
  \end{tabular}}
  \label{table6}
 \end{center}
\end{table}

\begin{table}
\renewcommand{\arraystretch}{1}
 \begin{center}
   \caption{Average and maximum \#iteration.}
   \centering
   \small{
   \begin{tabular}{|l|l|r|r|r|r|r|r|r|}
   \hline
Data &\#iter. & LS & D-U & D-B & MU-U & AU-U & MU-B & AU-B \\
\hline
Reuters2 & Av.  & 20 & 20 & 20 & 20 & 20 & 16.2 & 4.9 \\
         & Max. & 20 & 20 & 20 & 20 & 20 & 20 & 6 \\ \hline 
Reuters4 & Av.  & 20 & 20 & 20 & 20 & 20 & 20 & 7.2 \\
         & Max. & 20 & 20 & 20 & 20 & 20 & 20 & 20 \\ \hline 
Reuters6 & Av.  & 20 & 20 & 20 & 20 & 20 & 20 & 5.5 \\
         & Max. & 20 & 20 & 20 & 20 & 20 & 20 & 20 \\ \hline 
Reuters8 & Av.  & 20 & 20 & 20 & 20 & 20 & 20 & 4 \\
         & Max. & 20 & 20 & 20 & 20 & 20 & 20 & 4 \\ \hline 
Reuters10 & Av. & 20 & 20 & 20 & 20 & 20 & 20 & 5.6 \\
         & Max. & 20 & 20 & 20 & 20 & 20 & 20 & 20 \\ \hline 
Reuters12 & Av. & 20 & 20 & 20 & 20 & 20 & 20 & 8.8 \\
         & Max. & 20 & 20 & 20 & 20 & 20 & 20 & 20 \\ 
\hline
  \end{tabular}}
  \label{table7}
 \end{center}
\end{table}

\begin{table}
\renewcommand{\arraystretch}{1}
 \begin{center}
   \caption{Average and maximum errors at the last iteration.}
   \centering
   \small{
   \begin{tabular}{|l|l|r|r|r|r|r|r|r|}
   \hline
Data &\#iter. & LS & D-U & D-B & MU-U & AU-U & MU-B & AU-B \\
\hline
Reuters2 & Av.  & 1.3763 & 3435.6 & 3626.5 & 1.4106 & 1.4138 & 1.7955 & 1.8021  \\
         & Max. & 1.3854 & 3587.2 & 3867.4 & 1.4201 & 1.4230 & 1.8022 & 1.8025  \\ \hline 
Reuters4 & Av.  & 1.4791 & 9152.8 & 8689.0 & 1.5299 & 1.5310 & 2.0708 & 2.0962 \\
         & Max. & 1.4855 & 9474.9 & 9297.9 & 1.5408 & 1.5402 & 2.0880 & 2.1028 \\ \hline 
Reuters6 & Av.  & 1.5229 & 17135 & 15823 & 1.5844 & 1.5878 & 2.2627 & 2.2921 \\
         & Max. & 1.5301 & 17971 & 16955 & 1.5884 & 1.5952 & 2.2758 & 2.2998 \\ \hline 
Reuters8 & Av.  & 1.5434 & 25913 & 22893 & 1.6215 & 1.6171 & 2.3863 & 2.4421 \\
         & Max. & 1.5473 & 27462 & 25553 & 1.6342 & 1.6262 & 2.3993 & 2.4422 \\ \hline 
Reuters10 & Av. & 1.5696 & 34154 & 30518 & 1.6533 & 1.6533 & 1.8836 & 2.5673 \\
         & Max. & 1.5801 & 35236 & 35152 & 1.6662 & 1.6618 & 1.9529 & 2.5718 \\ \hline 
Reuters12 & Av. & 1.5727 & 42739 & 37038 & 1.6620 & 1.6621 & 1.8860 & 2.6551  \\
         & Max. & 1.5815 & 44325 & 41940 & 1.6705 & 1.6713 & 1.9193 & 2.6697 \\ 
\hline
  \end{tabular}}
  \label{table8}
 \end{center}
\end{table}

\subsection{Clustering capability}

One of the prominent application of NMF is in clustering, which is reported to be better than the spectral clustering \cite{Xu}. Especially, the orthogonal NMFs are designed to improve the clustering capability of the standard NMF \cite{Ding1}. Thus, the real assessment of the orthogonal NMFs qualities is in their clustering capability. 

\subsubsection{The metrics} \label{metrics}

There are some standard metrics in evaluating clustering quality. The most commonly used metrics are \emph{mutual information}, \emph{entropy}, and \emph{purity}. We will use these metrics together with an additional metric, \emph{Fmeasure}. In the following, the definitions of these metrics are outlined.

\emph{Mutual information} (MI) measures dependency between the clusters produced by the algorithms and the reference classes. The higher the MI, the most related the clusters with the classes, and therefore the better the clustering will be. It is shown that MI is a superior measure than \emph{purity} and \emph{entropy} \cite{Strehl} because it is tolerant to the dif\mbox{}ference between \#cluster and \#class. MI is defined with the following formula:
\begin{equation*}
MI \equiv \sum_{r=1}^{R}\sum_{s=1}^{S}p(r,s)\log_2\left(\frac{p(r,s)}{p(r)p(s)}\right),
\end{equation*}
where $r$ and $s$ denote the $r$-th cluster and $s$-th class respectively, $p(r,s)$ denotes the joint probability distribution function of the clusters and the classes, $p(r)$ and $p(s)$ denote the marginal probability distribution functions of the clusters and the classes respectively, and binary logarithm is used here (other bases are also possible). Note that because of inconsistency in the formulation of normalized MI (a more commonly used metric) in the literatures, we use MI instead. Accordingly, MI's values are comparable only for the same dataset.

\emph{Entropy} addresses the composition of classes in a cluster. It measures uncertainty in the cluster, thus the lower the \emph{entropy}, the better the clustering will be. Unlike MI, if there is discrepancy between \#cluster and \#class, \emph{entropy} won't be very indicative about the the clustering quality. \emph{Entropy} is defined with the following:
\begin{equation*}
entropy \equiv \frac{1}{N\log_2S}\sum_{r=1}^{R}\sum_{s=1}^{S}c_{rs}\log_2\frac{c_{rs}}{c_r},
\end{equation*}
where $N$ is the number of samples (\#doc for document clustering), $c_{rs}$ denotes the number of samples in $r$-th cluster that belong to $s$-th class, and $c_r$ denotes the size of $r$-th cluster.

\emph{Purity} is the most commonly used metric. It measures the percentage of the dominant class in a cluster, so the high the better. As in \emph{entropy}, \emph{purity} is also sensitive to the discrepancy between \#cluster and \#class. \emph{Purity} is defined with:
\begin{equation*}
purity = \frac{1}{N}\sum_{r=1}^R\max_s c_{rs}.
\end{equation*}

And \emph{Fmeasure} combines two concept in IR: \emph{recall} and \emph{precision}. \emph{Recall} measures the proportion of the retrieved relevant documents to all relevant documents, and \emph{precision} measures the proportion of the retrieved relevant documents to all retrieved documents. In the context of assessing clustering quality, \emph{Fmeasure} is defined with \cite{Andrews}:
\begin{align*}
Fmeasure \equiv \frac{1}{R}\sum_{r=1}^R F_r,\;\;F_r = 2\,\frac{precision_r \times recall_r}{precision_r + recall_r},
\end{align*}
where $precision_r$ and $recall_r$ denote the \emph{precision} and \emph{recall} of $r$-th cluster.

\subsubsection{Document clustering}

The results of document clustering are shown in table \ref{table9}--\ref{table12}. In average, MU-U gives the best performances in all metrics especially for datasets with small \#clusters. Then followed by LS, AU-U, and D-U with small margins. LS seems to be better for datasets with large \#clusters. Generally, MU-U, LS, AU-U and D-U can give consistent results for variety \#clusters, but unfortunately this is not the case for D-B, MU-B and AU-B which are all bi-orthogonal NMF algorithms. AU-B especially seems to offer only slightly better clustering than random results. Note that even though there are adjustable parameters in MU-B and AU-B, it is unlikely that the poor results are due to these parameters.

\begin{table}
 \begin{center}
   \caption{Average mutual information over 10 trials (document clustering).}
   \centering
   \footnotesize{
   \begin{tabular}{|l|r|r|r|r|r|r|r|}
   \hline
Data & LS & D-U & D-B & MU-U & AU-U & MU-B & AU-B \\
\hline
Reuters2 & 0.40392 & 0.42487 & 0.36560 & $\mathbf{0.47507}$ & 0.42150 & 0.057799 & 0.00087646 \\ 
Reuters4 & 0.62879 & 0.61723 & 0.48007 & $\mathbf{0.65080}$ & 0.63640 & 0.32142 & 0.072621 \\ 
Reuters6 & 0.79459 & 0.81831 & 0.52498 & 0.81811 & $\mathbf{0.82425}$ & 0.37924 & 0.078201 \\ 
Reuters8 & 0.92285 & 0.90260 & 0.54534 & $\mathbf{0.94165}$ & 0.92720 & 0.48435 & 0.013518 \\ 
Reuters10 & $\mathbf{1.0415}$ & 1.0275 & 0.62125 & 1.0063 & 1.0138 & 0.50980 & 0.072014 \\ 
Reuters12 & $\mathbf{1.1326}$ & 1.0865 & 0.58469 & 1.1195 & 1.0821 & 0.47697 & 0.16389 \\ \hline
Average & 0.82071 & 0.81283 & 0.52032 & $\mathbf{0.83523}$ & 0.81754 & 0.37160 & 0.066853 \\
\hline
  \end{tabular}}
  \label{table9}
 \end{center}
\end{table}

\begin{table}
 \begin{center}
   \caption{Average entropy over 10 trials (document clustering).}
   \centering
   \footnotesize{
   \begin{tabular}{|l|r|r|r|r|r|r|r|}
   \hline
Data & LS & D-U & D-B & MU-U & AU-U & MU-B & AU-B \\
\hline
Reuters2 & 0.54193 & 0.52098 & 0.58025 & $\mathbf{0.47078}$ & 0.52435 & 0.88805 & 0.94498 \\ 
Reuters4 &  0.40202 & 0.40780 & 0.47638 & $\mathbf{0.39102}$ & 0.39822 & 0.55571 & 0.68011 \\ 
Reuters6 &  0.38391 & 0.37473 & 0.48821 & 0.37481 & $\mathbf{0.37243}$ & 0.54459 & 0.66105 \\ 
Reuters8 & 0.35568 & 0.36242 & 0.48151 & $\mathbf{0.34941}$ & 0.35423 & 0.50184 & 0.65879 \\ 
Reuters10 & $\mathbf{0.33601}$ & 0.34023 & 0.46253 & 0.34661 & 0.34434 & 0.49608 & 0.62786 \\ 
Reuters12 & $\mathbf{0.31953}$ & 0.33239 & 0.47236 & 0.32319 & 0.33362 & 0.50241 & 0.58974 \\ 
\hline
Average & 0.38985 & 0.389760 & 0.49354 & $\mathbf{0.37597}$ & 0.38787 & 0.58145 & 0.69375 \\ 
\hline
  \end{tabular}}
  \label{table10}
 \end{center}
\end{table}

\begin{table}
 \begin{center}
   \caption{Average purity over 10 trials (document clustering).}
   \centering
   \footnotesize{
   \begin{tabular}{|l|r|r|r|r|r|r|r|}
   \hline
Data & LS & D-U & D-B & MU-U & AU-U & MU-B & AU-B \\
\hline
Reuters2 & 0.82154 & 0.83599 & 0.80452 & $\mathbf{0.85089}$ & 0.82507 & 0.66102 & 0.63612 \\ 
Reuters4 &  0.79417 & 0.78023 & 0.73778 & $\mathbf{0.80400}$ & 0.79704 & 0.70119 & 0.59657 \\ 
Reuters6 &  0.74510 & $\mathbf{0.75158}$ & 0.68844 & 0.74868 & 0.75069 & 0.66433 & 0.54569 \\ 
Reuters8 &  $\mathbf{0.74906}$ & 0.73982 & 0.66536 & 0.74869 & 0.73987 & 0.65033 & 0.50680 \\ 
Reuters10 & 0.73120 & $\mathbf{0.73762}$ & 0.64845 & 0.72813 & 0.73330 & 0.63194 & 0.50639 \\ 
Reuters12 & 0.73877 & 0.72719 & 0.62223 & $\mathbf{0.74127}$ & 0.72340 & 0.60118 & 0.52019 \\ 
\hline
Average & 0.76331 & 0.76207 & 0.69446 & $\mathbf{0.77028}$ & 0.76156 & 0.65166 & 0.55196 \\ 
\hline
  \end{tabular}}
  \label{table11}
 \end{center}
\end{table}

\begin{table}
 \begin{center}
   \caption{Average Fmeasure over 10 trials (document clustering).}
   \centering
   \footnotesize{
   \begin{tabular}{|l|r|r|r|r|r|r|r|}
   \hline
Data & LS & D-U & D-B & MU-U & AU-U & MU-B & AU-B \\
\hline
Reuters2 & 0.81904 & 0.83234 & 0.79163 & $\mathbf{0.84823}$ & 0.82241 & 0.58237 & 0.50399 \\ 
Reuters4 & 0.56154 & 0.53754 & 0.44352 & $\mathbf{0.57989}$ & 0.54267 & 0.36917 & 0.24585 \\ 
Reuters6 & 0.46225 & 0.47714 & 0.33910 & $\mathbf{0.48444}$ & 0.47270 & 0.26372 & 0.17171 \\ 
Reuters8 & 0.40408 & 0.40554 & 0.25052 & 0.41822 & $\mathbf{0.42996}$ & 0.23904 & 0.10869 \\ 
Reuters10 & 0.38001 & $\mathbf{0.38041}$ & 0.23309 & 0.36923 & 0.35947 & 0.19552 & 0.094912 \\ 
Reuters12 & 0.35671 & $\mathbf{0.35811}$ & 0.17387 & 0.35214 & 0.34435 & 0.16401 & 0.099949 \\ 
\hline
Average & 0.49727 & 0.49851 & 0.37196 & $\mathbf{0.50869}$ & 0.49526 & 0.30231 & 0.20418 \\ 
\hline
  \end{tabular}}
  \label{table12}
 \end{center}
\end{table}

\subsubsection{Word clustering}

In some cases, the ability of clustering methods to simultaneously group similar documents with related words (co-clustering) is a concern. And because the original bi-orthogonal NMF is designed to have this ability \cite{Ding1}, we will also investigate the quality of word clustering (in the context of co-clustering) produced by all algorithms. Since word clustering has no reference class, we adopt idea from ref.~\cite{Ding1} in which reference classes are created by using word frequencies: each word is assigned to class with the highest frequency. Table \ref{table13}--\ref{table16} show the results.

As shown in table \ref{table13}--\ref{table16}, D-U has the best overall results followed by LS, MU-U and AU-U by small margins. MU-U is especially good for small \#clusters and LS is good for large \#clusters. But unfortunately, all bi-orthogonal NMF algorithms, D-B, MU-B, and AU-B, which designed to accomodate co-clustering task, seem to have poor results. These results are in accord with document clustering cases where bi-orthogonal NMFs also perform poorly.

\begin{table}
 \begin{center}
   \caption{Average mutual information over 10 trials (word clustering).}
   \centering
   \footnotesize{
   \begin{tabular}{|l|r|r|r|r|r|r|r|}
   \hline
Data & LS & D-U & D-B & MU-U & AU-U & MU-B & AU-B \\
\hline
Reuters2 & 0.15715 & 0.16609 & 0.12966 & $\mathbf{0.17351}$ & 0.14978 & 0.013995 & 0.00029807 \\ 
Reuters4 &  0.42558 & 0.39193 & 0.21495 & $\mathbf{0.42619}$ & 0.41663 & 0.11812 & 0.026943 \\ 
Reuters6 &  0.54112 & $\mathbf{0.57472}$ &	0.26971 & 0.54239 & 0.54828 & 0.12460 & 0.035309 \\ 
Reuters8 &  0.63022 & 0.63368 & 0.29277 & 0.64699 & $\mathbf{0.65774}$ & 0.15692 & 0.0037071 \\ 
Reuters10 &  0.70386 & $\mathbf{0.73345}$ & 0.33046 & 0.66262 & 0.68367 & 0.025320 & 0.029618 \\ 
Reuters12 & 0.80111 & 0.77959 & 0.28412 & 0.76128 & 0.73517 & 0.013483 & 0.073478 \\ \hline
Average &  0.54317 & $\mathbf{0.54658}$ & 0.25361 & 0.53549 & 0.53188 & 0.075407 & 0.028226 \\
\hline
  \end{tabular}}
  \label{table13}
 \end{center}
\end{table}

\begin{table}
 \begin{center}
   \caption{Average entropy over 10 trials (word clustering).}
   \centering
   \footnotesize{
   \begin{tabular}{|l|r|r|r|r|r|r|r|}
   \hline
Data & LS & D-U & D-B & MU-U & AU-U & MU-B & AU-B \\
\hline
Reuters2 & 0.76778 & 0.75884 & 0.79527 & $\mathbf{0.75142}$ & 0.77515 & 0.91094 & 0.92463 \\ 
Reuters4 & 0.62965 & 0.64647 & 0.73496 & $\mathbf{0.62934}$ & 0.63412 & 0.78338 & 0.82897 \\ 
Reuters6 & 0.56184 & $\mathbf{0.54884}$ & 0.66683 & 0.56134 & 0.55906 & 0.72297 & 0.75751 \\ 
Reuters8 & 0.52006 & 0.51891 & 0.63255 & 0.51447 & $\mathbf{0.51089}$ & 0.67783 & 0.72890 \\ 
Reuters10 & 0.50612 & $\mathbf{0.49721}$ & 0.61852 & 0.51853 & 0.51220 & 0.71038 & 0.70909 \\ 
Reuters12 & $\mathbf{0.48211}$ & 0.48811 &	0.62632 & 0.49322 & 0.50050 & 0.70181 & 0.68507 \\ \hline
Average & 0.57792 & $\mathbf{0.57640}$ & 0.67908 & 0.57806 & 0.58199 & 0.75122 & 0.77236 \\
\hline
  \end{tabular}}
  \label{table14}
 \end{center}
\end{table}

\begin{table}
 \begin{center}
   \caption{Average purity over 10 trials (word clustering).}
   \centering
   \footnotesize{
   \begin{tabular}{|l|r|r|r|r|r|r|r|}
   \hline
Data & LS & D-U & D-B & MU-U & AU-U & MU-B & AU-B \\
\hline
Reuters2 & 0.76987 & 0.77082 & 0.75378 & $\mathbf{0.77730}$ & 0.76021 & 0.67006 & 0.65988 \\ 
Reuters4 & 0.64400 & 0.62881 & 0.60566 & $\mathbf{0.64676}$ & 0.64184 & 0.55808 & 0.53116 \\ 
Reuters6 & 0.59830 & $\mathbf{0.61733}$ & 0.55949 & 0.59763 & 0.59103 & 0.52966 & 0.49661 \\ 
Reuters8 & $\mathbf{0.59560}$ & 0.58935 & 0.54296 & 0.59179 & 0.58770 & 0.50933 & 0.46499 \\ 
Reuters10 & 0.58123 & $\mathbf{0.60236}$ & 0.51576 & 0.57045 & 0.58724 & 0.44765 & 0.45395 \\ 
Reuters12 &  $\mathbf{0.60208}$ & 0.59563 & 0.49555 & 0.58628 & 0.56846 & 0.43611 & 0.44882 \\ \hline
Average & 0.63185 & $\mathbf{0.63405}$ & 0.57887 & 0.62837 & 0.62274 & 0.52515 & 0.50923 \\
\hline
  \end{tabular}}
  \label{table15}
 \end{center}
\end{table}

\begin{table}
 \begin{center}
   \caption{Average Fmeasure over 10 trials (word clustering).}
   \centering
   \footnotesize{
   \begin{tabular}{|l|r|r|r|r|r|r|r|}
   \hline
Data & LS & D-U & D-B & MU-U & AU-U & MU-B & AU-B \\
\hline
Reuters2 & 0.59287 & 0.59471 & 0.58733 & $\mathbf{0.59696}$ & 0.59427 & 0.52628 & 0.49976 \\ 
Reuters4 & 0.46891 & 0.43469 & 0.36397 & $\mathbf{0.48118}$ & 0.46180 & 0.32520 & 0.27101 \\ 
Reuters6 & 0.37490 & 0.38365 & 0.27356 & $\mathbf{0.38648}$ & 0.38026 & 0.21620 & 0.17572 \\ 
Reuters8 & 0.32488 & 0.32674 & 0.20820 & 0.33527 & $\mathbf{0.34251}$ & 0.17127 & 0.12565 \\ 
Reuters10 & 0.29864 & $\mathbf{0.30768}$ & 0.18626 & 0.28930 & 0.28573 & 0.10700 & 0.10545 \\ 
Reuters12 & $\mathbf{0.29116}$ & 0.29072 & 0.14255 & 0.27525 & 0.27380 & 0.088517 & 0.095880 \\ 
\hline
Average & 0.39189 & 0.38970 & 0.29365 & $\mathbf{0.39407}$ & 0.38973 & 0.23908 & 0.21224 \\
\hline
  \end{tabular}}
  \label{table16}
 \end{center}
\end{table}

\section{Conclusions}

We have presented orthogonal NMF algorithms based on the additive update rules with rigorous convergence proofs. There are two versions of the converged algorithms: AU-U for uni-orthogonal NMF, and AU-B for bi-orthogonal NMF with their respective multiplicative update rules versions: MU-U and MU-B.

The only way to numerically evaluate whether the algorithm has converged to a stationary point is to check whether it has satisfied the KKT conditions on that point. While the nonnegativity conditions are easy to check, the complementary slackness conditions are hard since we must check $\nabla_{\mathbf{X}}J\big(\mathbf{X}^k\big)\odot\mathbf{X}^k=\mathbf{0}\;\,\forall k \ge *$. Not only there are some large matrix multiplications which can be inaccurate numerically, but also we must make sure that the stationary point is reachable in a reasonable amount of time. Accordingly, only the nonincreasing properties were evaluated which as shown in section \ref{nonincreasing}, the converged version of our algorithms kept these properties even for large $\alpha$ or $\beta$.

The maximum allowed \#iterations is an important issue in the multiplicative and additive update rules based NMF algorithms since these algorithms are known to be slow. As shown in table \ref{table5}, the multiplicative and additive update rules based algorithms were exceptionally very good at reducing the errors even in the first iterations, but then the errors were only slightly reduced for the remaining iterations. This inspired us to use 20 iterations as the maximum \#iterations. Because this is a rather small number, it is very likely that the algorithms stop before reaching a stationary point.

There are adjustable parameters in our proposed algorithms. These parameters are dataset-dependent, and thus should be learned based on the datasets. Because the objectives of our algorithms resemble the objectives of sparse NMFs, better clustering results probably can be achieved by using the same strategy: setting these parameters to small numbers.

There were differences in the running times of the algorithms, but were not significant since all algorithms have the same computation complexity: \#iterations$\times M\times N\times R$, where $M\times N$ is the size of the data matrix, and $R$ is the number of decomposition factors.

The document clustering results favoured our MU-U algorithm in which it showed the best average performances for all used metrics followed closely by LS, AU-U, and D-U. MU-U was especially good for small \#cluster and LS for large \#clusters. There is possibility that because we learned $\alpha$ from Reuters4 dataset, then MU-U performed best at the small datasets. But, because adjusting $\alpha$ for each different dataset is rather unfair, we believe that these are the best results can be offered by MU-U. All bi-orthogonal NMF algorithms, D-B, MU-B, and AU-B, performed rather poorly in these datasets, which was unfortunate since there are some works that show D-B is a better clustering method compared to LS and D-U \cite{Ding1,TLi}. 

The word clustering results were not as conclusive as the document clustering results since there is no a prior label to compare with. Here we used strategy from ref.~\cite{Ding1} to assign the words to the classes. In this task, D-U offered the best overall performances followed closely by LS, MU-U and AU-U. As in the document clustering, all bi-orthogonal NMF algorithms also performed poorly in this task.

\end{document}